\documentclass{./include/latex-class-tlp/new_tlp}

\usepackage[utf8]{inputenc}
\usepackage[T1]{fontenc}
\usepackage{lmodern}
\usepackage{amsmath}
\usepackage{amssymb}
\usepackage{microtype}
\usepackage{url}
\usepackage{relsize}
\usepackage{tikz}
\usepackage{listings}
\usepackage{hyperref}
\usepackage{float}

\newcommand{\sysfont}{\textit}

\newcommand{\Clingo}{\sysfont{Clingo}}

\newcommand{\adsolver}{\sysfont{adsolver}}

\newcommand{\aspmt}{\sysfont{aspmt}}

\newcommand{\asprin}{\sysfont{asprin}}

\newcommand{\clasp}{\sysfont{clasp}}
\newcommand{\clingcon}{\sysfont{clingcon}}
\newcommand{\clingo}{\sysfont{clingo}}

\newcommand{\dingo}{\sysfont{dingo}}
\newcommand{\dlvhex}{\sysfont{dlvhex}}
\newcommand{\dlv}{\sysfont{dlv}}
\newcommand{\eclingo}{\sysfont{eclingo}}

\newcommand{\ezsmt}{\sysfont{ezsmt}}

\newcommand{\gringo}{\sysfont{gringo}}

\newcommand{\iclingo}{\sysfont{iclingo}}
\newcommand{\idlv}{\sysfont{idlv}}

\newcommand{\lctocasp}{\sysfont{lc2casp}}
\newcommand{\lparse}{\sysfont{lparse}}
\newcommand{\lpconvert}{\sysfont{lpconvert}}

\newcommand{\mingo}{\sysfont{mingo}}

\newcommand{\minisat}{\sysfont{minisat}}

\newcommand{\plasp}{\sysfont{plasp}}

\newcommand{\smodels}{\sysfont{smodels}}

\newcommand{\telingo}{\sysfont{telingo}}

\newcommand{\wasp}{\sysfont{wasp}}

\newcommand{\clingoM}[1]{\clingo{\small\textnormal{[}\textsc{#1}\textnormal{]}}}

\newcommand{\aspif}{\sysfont{aspif}}

\newcommand{\C}{C}
\newcommand{\cpp}{C++}

\newcommand{\java}{Java}
\newcommand{\lua}{Lua}

\newcommand{\python}{Python}
\newcommand{\rust}{Rust}

\providecommand{\Underscore}{\textunderscore}

\lstdefinelanguage{clingo}{basicstyle=\ttfamily,keywordstyle=[1]\bfseries,keywordstyle=[2]\bfseries,keywordstyle=[3]\bfseries,showstringspaces=false,literate={_}{\Underscore}1 {\%\%}{}0,escapeinside={\#(}{\#)},alsoletter={\#,\&},keywords=[1]{not,from,import,def,if,else,elif,return,while,break,and,or,for,in,del,and,class,with,as,is,yield,async},keywords=[2]{\#const,\#show,\#minimize,\#base,\#theory,\#count,\#external,\#program,\#script,\#end,\#heuristic,\#edge,\#project,\#show,\#sum},keywords=[3]{&,&dom,&sum,&diff,&show},morecomment=[l]{\#\ },morecomment=[l]{\%\ },morestring=[b]",stringstyle={\itshape},commentstyle={\color{darkgray}}}

\lstdefinelanguage{python}{basicstyle=\ttfamily,keywordstyle=[1]\bfseries,showstringspaces=false,literate={_}{\Underscore}{1},escapeinside={\#(}{\#)},alsoletter={\#,\&},keywords=[1]{not,from,import,def,if,else,elif,return,while,break,and,or,for,in,del,and,class,with,as,is,yield,async},morecomment=[l]{\#\ },morestring=[b]",stringstyle={\itshape},commentstyle={\color{darkgray}}}

 \newcommand{\potassco}{\textsc{Potassco}}

\newcommand{\HT}{\ensuremath{\mathrm{HT}}}

\newtheorem{remark}{Remark}

\lstdefinelanguage{shell}{frame=single,basicstyle=\footnotesize\ttfamily,numbers=none,escapeinside={|}{|},xrightmargin=2\parindent }

\lstdefinelanguage{shells}{frame=single,basicstyle=\footnotesize\ttfamily,numbers=none,xrightmargin=2\parindent }

\lstdefinelanguage{others}{numbers=none,
  basicstyle=\small\ttfamily }

\lstdefinelanguage{pythons}{language=python,basicstyle=\small\ttfamily }

\lstdefinelanguage{clingos}{language=clingo,basicstyle=\small\ttfamily }

\begin{document}

\title{How to build your own ASP-based system\,?!}

\author[Roland Kaminski et al.]{ROLAND~KAMINSKI,
  \
  JAVIER~ROMERO,
  \
  TORSTEN~SCHAUB,\
  and
  PHILIPP~WANKO\\
  University of Potsdam, Germany
}

\maketitle
\begin{abstract}
Answer Set Programming, or ASP for short, has become a popular and sophisticated approach to declarative problem solving.
Its popularity is due to its attractive modeling-grounding-solving workflow that provides an easy approach to
problem solving, even for laypersons outside computer science.
However, in contrast to ASP's ease of use,
the high degree of sophistication of the underlying technology makes it even hard for ASP experts
to put ideas into practice whenever this involves modifying ASP's machinery.

For addressing this issue,
this tutorial aims at enabling users to build their own ASP-based systems.
More precisely,
we show how the ASP system \clingo\ can be used
for extending ASP and
for implementing customized special-purpose systems.
To this end, we propose two alternatives.
We begin with a traditional AI technique and show
how meta programming can be used for extending ASP.
This is a rather light approach that relies on \clingo's reification feature
to use ASP itself for expressing new functionalities.
The second part of this tutorial uses traditional programming (in \python) for manipulating \clingo\
via its application programming interface.
This approach allows for changing and controlling the entire model-ground-solve workflow of ASP.
Central to this is \clingo's new \lstinline{Application} class that
allows us to draw on \clingo's infrastructure by customizing processes similar to the one in \clingo.
For instance,
we may
apply manipulations to programs' abstract syntax trees,
control various forms of multi-shot solving, and
set up theory propagators for foreign inferences.
A cross-sectional structure, spanning meta as well as application programming,
is \clingo's intermediate format, \aspif, that specifies the interface among
the underlying grounder and solver.
We illustrate the aforementioned concepts and techniques throughout this tutorial by means of
examples and several non-trivial case-studies.
In particular, we show
how \clingo\ can be extended by difference constraints and
how guess-and-check programming can be implemented with both meta and application programming.
\end{abstract}

\section{Introduction}\label{sec:introduction}

Answer Set Programming (ASP;~\citeNP{baral02a,gekakasc12a,gelkah14a,lifschitz19a})
has become an established approach to declarative problem solving,
experiencing an increasing popularity in academia as well as industry,
and sometimes even beyond Artificial Intelligence and Computer Science.
This is arguably due to its pursuit of an integrated modeling-grounding-solving paradigm~\cite{gebsch16a,kalepesc16a}
that enables laypersons to use ASP systems off-the-shelf.
However, the underlying technology is highly involved and thus much less accessible even for ASP experts.
This is also reflected by the fact that there are only two genuine ASP systems nowadays,
namely \dlv~\cite{dlv03a,alcadofuleperiveza17a} and \clingo~\cite{gekakasc17a},
while other computational approaches mostly rely on extensions to these systems or translations into neighboring solving paradigms.
This is not without reason and rather due to the high technical sophistication of full-fledged ASP systems.
Hence, it is all the more important to keep this technology open and extensible and so to enable the community
to participate in the continuous enhancement of ASP technology.
If neglected, we risk a technological gulf that is prone to cut off advances in ASP in the future.
Moreover, the extension and integration of ASP technology is indispensable in many real-world applications.
Examples include
decision support systems for the space shuttle~\cite{nobagewaba01a},
metabolic network completion~\cite{frscscsiwa18a}, and
train scheduling~\cite{abjoossctowa21a}.
Hence,
empowering the community to master ASP technology
also makes it fit for addressing applications at industrial scale.

This empowerment was a guiding motive in the development of the core ASP systems of the
{\em Potsdam Answer Set Solving Collection},
or \potassco\ for short~\cite{gekakaosscsc11a,gekakaluobosroscscwa18a},
and has meanwhile led to numerous extensions,
either being part of \potassco\ at \url{potassco.org} or conducted by other scientists worldwide.
To further foster such advancements and transfer of ASP technology,
we provide in this tutorial an introduction to key techniques allowing advanced users to construct their own ASP systems
by building upon \potassco\ tools.

No matter whether the envisaged system aims at
extending ASP or
using it as an implementation platform,
the key issue is how to capture the added functionality.

To this end, we propose two alternatives.

We begin with a traditional AI technique and show in Section~\ref{sec:meta}
how meta programming can be used for extending ASP.
This is clearly the lightest approach in which ASP itself is used to express new functionalities.
It draws upon \clingo's reification feature for representing the result of
grounding a logic program as a set of facts.
The original program is then given as data to a meta program that implements the new functionality.
In this way, we use \clingo\ as a black box and implement all examples by consecutive \clingo\ calls.
Meta programming is for example used in \asprin~\cite{brderosc15b} and \plasp~\cite{digelurosc18a}.

We then move to the other focus of our tutorial, namely,
the use of traditional programming for manipulating \clingo\ via its application programming interface (API).
This can be seen as treating \clingo\ as a gray box,
whose modifications are guided through a well-defined interface.
Before that, all functionality had to be done by re-programming,
a white box approach that needed quite good programming skills.
For application interface programming, we have chosen \python\ as our example language, although other choices exist
(e.g.\ \C, \cpp, \lua, and \rust).
This approach allows us to make changes to the entire model-ground-solve workflow of ASP.
We detail capabilities and interfaces supporting
the implementation of novel ASP technology
such as
extending the modeling language of \clingo\ by means of grammar-based specifications,
manipulating the abstract syntax trees of (non-ground) logic programs,
as well as multi-shot and theory solving.
While multi-shot solving provides us with fine-grained control of ASP reasoning processes,
theory solving allows for refining basic ASP solving by incorporating foreign types of constraints.
Central to this is \clingo's new application class that allows for deriving customized applications from the
one of \clingo.
This class constitutes the cornerstone of all recent \potassco\ systems such as
\clingcon~\cite{bakaossc16a},
\clingoM{dl}~\cite{jakaosscscwa17a},
\eclingo~\cite{cafagarosc20a}, and
\telingo~\cite{cakamosc19a}.
We discuss its role in Section~\ref{sec:app} and use it throughout the remaining sections.

Both meta and application interface programming allow for changing the functionality of ASP systems.
One difference manifests itself in their degree of elaboration-tolerance~\cite{mccarthy98a}.
While a meta encoding benefits from ASP's elaboration-tolerance,
this feature is less pronounced in non-declarative programming languages.
Here, however, an API makes the difference since it greatly simplifies programming by
abstracting from an underlying implementation.
Although an API is usually less accessible to ASP users than a meta encoding,
it is much easier to handle than any intervention into the programming of the actual ASP system.
This differentiation reflects the above distinction of treating an ASP system as a black, gray, or white box, respectively.
Also, the possibility of changing an ASP system's functionality brings about the new role of an ASP engineer,
which is situated between
basic ASP users, using ASP systems as such,
and ASP system builders.
Obviously, meta programming offers a light entry point for basic users, and
its easy accessibility makes it well suited for prototyping new functionalities.
On the other hand, API programming can be accomplished with much less programming skills than ASP system building.
Also, the usage of an ASP system's API is the predominant use case in industrial applications since it allows for a
flexible integration into an existing IT infrastructure.
Last but not least, it is instructive to realize that the effectiveness of the chosen alternative
depends on the nature of the added functionality.
Whenever it can be mapped back onto ASP, meta reasoning might be quite efficient
since it harnesses the power of modern ASP systems.
Any functionality exceeding ASP's capabilities, however, needs an extension to the system as such.

A cross-sectional structure, spanning meta as well as application interface programming,
is \clingo's intermediate format, \aspif, that specifies the interface among
the underlying grounder and solver, namely, \gringo\ and \clasp.
This is relevant whenever one deals with ground logic programs,
be it as reified rules, in machine-readable format, or via the application interface.
The whole input, including rules, customized language expressions, as well as all types of directives,
is expressed in their ground form in the \aspif\ format.
The complete \aspif\ specification is given in \ref{sec:aspif}.
A system that relies on translating ground logic programs in \aspif\ format is \lctocasp~\cite{cakaossc16a}.

We illustrate the aforementioned techniques throughout this tutorial by means of examples and
several non-trivial case-studies.
This includes the computation of classical, supported, here-and-there, and diverse models with meta programming in Section~\ref{sec:meta},
optimization and incremental solving with multi-shot solving in Section~\ref{sec:multi},
hybrid solving and optimization with theory solving in Section~\ref{sec:case},
and finally
guess-and-check programming with both meta and application programming in Sections~\ref{sec:meta:gc} and~\ref{sec:gnt},
respectively.
The source code of all examples is available online~\cite{meta-encodings,clingo-opt,clingo-dl,clingo-gnt}.

In what follows,
we refrain from distinguishing features of \gringo\ and \clasp\ and simply refer to them as features of \clingo.
We deal in this tutorial with \clingo\ series~5, in particular, \clingo\ version~5.5;
its installation instructions can be found on its webpage~\cite{clingo}.
A complete documentation of \clingo's API is also available online~\cite{clingo-api-temp}.
We rely on a basic familiarity with ASP and its underlying concepts.
Comprehensive introductions can be found in several textbooks~\cite{baral02a,gekakasc12a,gelkah14a,lifschitz19a}.
Accordingly, we only sketch \clingo's input language and refer for details to the
\emph{Potassco User Guide}~\cite{PotasscoUserGuide}.
Otherwise,
we presuppose some computer science training that permits a basic understanding of shell and \python\ programming
(for meta and application interface programming, respectively).

The core of this tutorial is based on material stemming from an earlier edition on hybrid answer set solving~\cite{kascwa17a}.
The tutorial at hand provides itself the basis of an advanced ASP course,
offering complementary teaching material~\cite{potassco-teaching}.

 \section{Answer set programming}\label{sec:background}

A {logic program} consists of {rules} of the form
\begin{lstlisting}[mathescape,numbers=none]
   a$_1$;...;a$_m$ :- a$_{m+1}$,...,a$_n$,not a$_{n+1}$,...,not a$_o$
\end{lstlisting}
where each \lstinline[mathescape]{a$_i$} is
an {atom} of form \lstinline[mathescape]{p(t$_1$,...,t$_k$)}
and all \lstinline[mathescape]{t$_i$} are terms,
composed of function symbols and variables.
For $0 \leq m \leq n \leq o$,
atoms \lstinline[mathescape]{a$_1$} to \lstinline[mathescape]{a$_m$} are often called head atoms,
while \lstinline[mathescape]{a$_{m+1}$} to \lstinline[mathescape]{a$_n$}
and \lstinline[mathescape]{not a$_{n+1}$} to \lstinline[mathescape]{not a$_o$}
are also referred to as positive and negative body literals, respectively.
An expression is said to be {ground}, if it contains no variables.
As usual, \lstinline[mathescape]{not} denotes (default) {negation}.
A rule is called a {fact} if $m=n=o=1$,
normal if $m=1$, and
an integrity constraint if $m=0$.
Semantically, a logic program induces a set of {stable models},
being distinguished models of the program determined by the stable models semantics~\cite{gellif90a}.

To ease the use of ASP in practice,
several extensions have been developed.
First of all, rules with variables are viewed as shorthands for the set of their ground instances.
Further language constructs include
{conditional literals} and {cardinality constraints}~\cite{siniso02a}.
The former are of the form\footnote{In rule bodies, they are terminated by `\lstinline{;}' or `\lstinline{.}'~\cite{PotasscoUserGuide}.}
\lstinline[mathescape]{a:b$_1$,...,b$_m$},
the latter can be written as\footnote{More elaborate forms of aggregates are obtained by explicitly using function (e.g.\ \lstinline{#count}) and relation symbols (e.g.\ \lstinline{<=})~\cite{PotasscoUserGuide}.}
\lstinline[mathescape]+s$\,${d$_1$;...;d$_n$}$\,$t+,
where \lstinline{a} and \lstinline[mathescape]{b$_i$} are possibly default-negated (regular) literals  and each \lstinline[mathescape]{d$_j$} is a conditional literal; \lstinline{s} and \lstinline{t} provide optional lower and upper bounds on the number of satisfied literals in the cardinality constraint.
We refer to \lstinline[mathescape]{b$_1$,...,b$_m$} as a {condition}.
The practical value of both constructs becomes apparent when used with variables.
For instance, a conditional literal like
\lstinline[mathescape]{a(X):b(X)}
in a rule's body expands to the conjunction of all instances of \lstinline{a(X)} for which the corresponding instance of \lstinline{b(X)} holds.
Similarly,
\lstinline[mathescape]+2$\;${a(X):b(X)}$\;$4+
is true whenever at least two and at most four instances of \lstinline{a(X)} (subject to \lstinline{b(X)}) are true.
More sophisticated examples are given in Section~\ref{sec:meta}, e.g.\ in Listing~\ref{prg:meta}.
Finally,
objective functions minimizing the sum over the first argument $w_i$ of a set of weighted tuples $(w_i,\boldsymbol{t}_i)$,
whose membership is subject to condition $c_i$,
are expressed as
\lstinline[mathescape]!#minimize{$w_1$@$l_1$,$\boldsymbol{t}_1$:$c_1$;$\dots$;$w_n$@$l_n$,$\boldsymbol{t}_n$:$c_n$}!.
Lexicographically ordered objective functions are (optionally) distinguished via levels indicated by $l_i$.
An omitted level defaults to 0.

As an example, consider the rule in Line~\ref{fig:toh:opt:enc:move} of Listing~\ref{fig:toh:opt:enc}:
\begin{lstlisting}[numbers=none,language=clingos]
1 { move(D,P,T): disk(D), peg(P) } 1 :- ngoal(T-1), T<=n.
\end{lstlisting}
This rule has a single head atom consisting of a cardinality constraint;
it comprises all instances of \lstinline{move(D,P,T)},
where \lstinline{T} is constrained by the two body literals,
and \lstinline{D} and \lstinline{P} vary over all instantiations of predicates \lstinline{disk} and \lstinline{peg}, respectively.
Given 3 pegs and 4 disks as in Listing~\ref{fig:toh:opt:ins}, this results in 12 instances of \lstinline{move(D,P,T)} for each valid replacement of
\lstinline{T}, among which exactly one must be chosen according to the above rule.

Full details on the input language of \clingo\ along with various examples can be found
in the \emph{Potassco User Guide}~\cite{PotasscoUserGuide};
its semantics is given by \citeANP{gehakalisc15a}~\citeNN{gehakalisc15a} in terms of infinitary formulas.

\section{Meta programming}
\label{sec:meta}

Meta programming is a technique in which computer programs treat other programs as data
\cite{enwiki:1001427050}.
Although this includes traditional compilers and interpreters,
it has always played a prominent role in AI languages, such as Lisp or Prolog,
since their syntactic proximity of program and data offers an easy way of self-modification.
For instance in Prolog, meta programs allow for manipulating the execution of logic programs
and constitute an easy way to extend programs with debugging information.
Moreover, special-purpose predicates enable the conversion of data into new program parts during run-time.

Similarly, meta programming can be used in ASP to change the semantics of language constructs and/or implement new ones.
Examples include
reasoning about action and change~\cite{bargel00a,sobanamc03a,digelurosc18a},
debugging~\cite{gepuscto08a},
preferences~\cite{gelson97a,descto02a,eifalepf03a} and optimization~\cite{gekasc11b}, as well as
guess-and-check programming~\cite{eitpol06a}.
The latter is of particular interest to us since we detail its implementation below with meta programming as well as
through application interfaces in Section~\ref{sec:gnt}.

A common difficulty of such approaches is the conversion of programs into data, or in terms of ASP,
the transformation of (non-ground) logic programs into sets of facts.
Either a dedicated parser is built or a user is expected to write a program in terms of a prescribed fact format.
Consequently, the resulting systems are mostly propositional and only offer a limited set of language constructs.
This is because they cannot draw upon the infrastructure of an ASP system for parsing and grounding.

This issue is addressed in \clingo, or more precisely its grounder \gringo,
by means of a fact-based representation of the grounded logic program.
This enables sophisticated meta programming
that may draw on the full-featured non-ground input language of \clingo,
a highly effective grounding procedure, and ultimately
a factual representation reflecting all features of the input language.
The remainder of this section provides an introduction to meta programming with \clingo.
The extension of logic programs during run-time is explained in the subsequent sections.

\subsection{Reification format}
\label{sec:meta:format}

The process of turning a (ground) logic program into a set of facts is also called \emph{reification}.
\Clingo's fact format of reified programs follows its intermediate language \aspif,
detailed in \ref{sec:aspif}.\footnote{\label{fn:reification:clingo:four}Reification was originally introduced in \clingo~4.
  However, the corresponding format is different since
  it is modeled after the intermediate format of \smodels~\cite{lparseManual}.
  For details, we refer to the paper by~\citeANP{gekasc11b}~\citeNN{gekasc11b}.}
This is no coincidence since both languages must capture ground programs in their full generality.
In what follows, however, we concentrate on dealing with the actual logic program part
and disregard non-logical statements except for \lstinline{#show}.

A logic program consists of a set of rules,
each of which is composed of a head and a body.
While heads are formed from atoms, bodies are made of literals.

The fact format is the result of serializing the syntax tree of the ground logic program rule by rule.
To this end, heads and bodies are identified via non-negative integers.
Also,
positive and negative integers are used to represent positive or  negative literals, respectively.
Hence, \lstinline{0} is not a valid literal.

A rule is represented as a binary fact, using predicate
\lstinline{rule/2},
whose arguments reflect the head and the body of the rule.
Following the rule format of \aspif, a head is either a disjunction or a choice,
which is indicated by the unary function symbols \lstinline{disjunction/1} and \lstinline{choice/1}.
Similarly, a body is either a collection of literals or a weight constraint,
indicated by functions \lstinline{normal/1} and \lstinline{sum/1}, respectively.
All four constituents are treated as tuples,
the two former consisting of atoms and the latter either of regular or weighted literals, respectively.

\begin{table}
\begin{lstlisting}[caption={A simple logic program (\texttt{ezy.lp})},label={prg:ezy},language=clingos]
{a}.%%#(\label{lp:example:rule:one}#)
b :- a.
c :- not a.
\end{lstlisting}
\begin{lstlisting}[caption={System call to reify the logic program in Listing~\ref{prg:ezy}},label={prg:ezy:reify},language=others]
clingo --output=reify ezy.lp
\end{lstlisting}
\begin{lstlisting}[caption={The result of the system call in Listing~\ref{prg:ezy:reify} (\texttt{ezy.rlp})},label={prg:ezy:reified},language=clingos]
rule(choice(0),normal(0)).             %%#(\label{rlp:example:rule:one:one}#)
atom_tuple(0).     literal_tuple(0).   %%#(\label{rlp:example:rule:one:two}#)
atom_tuple(0,1).                       %%#(\label{rlp:example:rule:one:tri}#)

rule(disjunction(1),normal(1)).        %%#(\label{rlp:example:rule:two:one}#)
atom_tuple(1).     literal_tuple(1).   %%#(\label{rlp:example:rule:two:two}#)
atom_tuple(1,2).   literal_tuple(1,-1).%%#(\label{rlp:example:rule:two:tri}#)

rule(disjunction(2),normal(2)).        %%#(\label{rlp:example:rule:tri:one}#)
atom_tuple(2).     literal_tuple(2).   %%#(\label{rlp:example:rule:tri:two}#)
atom_tuple(2,3).   literal_tuple(2,1). %%#(\label{rlp:example:rule:tri:tri}#)

output(a,2).                           %%#(\label{rlp:example:output:one:one}#)

output(b,3).       literal_tuple(3).   %%#(\label{rlp:example:output:two:one}#)
                   literal_tuple(3,3). %%#(\label{rlp:example:output:two:two}#)

output(c,4).       literal_tuple(4).   %%#(\label{rlp:example:output:tri:one}#)
                   literal_tuple(4,2). %%#(\label{rlp:example:output:tri:two}#)
\end{lstlisting}
\end{table}

Let us illustrate this with the example program \lstinline{ezy.lp} in Listing~\ref{prg:ezy}.
Its (reformatted) reified program given in Listing~\ref{prg:ezy:reified} is obtained by the command in Listing~\ref{prg:ezy:reify}.
More precisely,
the first rule in program~\lstinline{ezy.lp} is represented by
the facts in Lines~\ref{rlp:example:rule:one:one} to~\ref{rlp:example:rule:one:tri} of Listing~\ref{prg:ezy:reified}.
The fact `\lstinline{rule(choice(0),normal(0)).}'
states that the first rule in Listing~\ref{prg:ezy} has
a choice atom in the head (indicated by \lstinline{choice(0)}) and
a normal body (denoted by \lstinline{normal(0)}).
The atoms associated with a head are grouped in tuples, which are identified by a non-negative integer.
This is analogous to the treatment of literals in the body.
In our example,
the choice in the head is linked via the identifier \lstinline{0}
to a tuple of atoms declared by the fact `\lstinline{atom_tuple(0).}'
in Line~\ref{rlp:example:rule:one:two} of Listing~\ref{prg:ezy:reified}.
The members of such an atom tuple are represented by all instances of predicate \lstinline{atom_tuple/2}
and share the same tuple identifier as their first argument.
The atom tuple \lstinline{0} has one member, as indicated by the single instance \lstinline{atom_tuple(0,1)}
in Line~\ref{rlp:example:rule:one:tri},
where
\lstinline{0} stands for the tuple and \lstinline{1} is the integer identifying atom \lstinline{a}.
To summarize,
the choice atom concerns only one regular atom and this atom is identified by \lstinline{1}
(thus connecting \lstinline{a} and \lstinline{1}; see below).

Analogously, the (empty) body of the choice rule is represented by the tuple of literals that is also identified by \lstinline{0}.
This tuple happens to be empty, as reflected by the lack of corresponding instances of \lstinline{literal_tuple/2}.
Note that this tuple-centered representation treats atom and literal tuples independently and numbers them
consecutively.
Heads and bodies of the same rule may thus be represented by atom and literal tuples having distinct identifiers.
Rules themselves have no identifier.

The second rule in Listing~\ref{prg:ezy} is represented by the facts
in Lines~\ref{rlp:example:rule:tri:one} to~\ref{rlp:example:rule:tri:tri} of Listing~\ref{prg:ezy:reified}.
Unlike above, its head is a single atom and is thus represented as a one-element \lstinline{disjunction} associated with atom tuple \lstinline{2}.
This tuple has a single element, which is accounted for by the fact \lstinline{atom_tuple(2,3).}
Hence, \lstinline{b} is represented by \lstinline{3}.
Similarly, its body, also marked with \lstinline{2}, comprises a single literal captured by \lstinline{literal_tuple(2,1).}
As above, \lstinline{1} stands for the positive body literal \lstinline{a}.
The last rule in Listing~\ref{prg:ezy} is represented analogously
in Lines~\ref{rlp:example:rule:two:one} to~\ref{rlp:example:rule:two:tri},
just that its negative body literal is mapped to a negative integer, namely,
\lstinline{not a} is associated with \lstinline{-1}.

The remaining facts in Lines~\ref{rlp:example:output:one:one} to~\ref{rlp:example:output:tri:two}
account for implicit output statements.
This mimics the default behavior of \clingo, outputting all atoms
unless a restrictive \lstinline{#show} statement is given.
That is, unless any restrictions are formulated, all satisfied atoms can potentially be output.
This is done by means of the binary predicate \lstinline{output/2}.
For instance, the output of atom \lstinline{c} is linked via literal tuple \lstinline{4} to integer \lstinline{2}
(cf.\ Lines~\ref{rlp:example:output:tri:one} and~\ref{rlp:example:output:tri:two}).
The indirection via the tuple representation is due to the fact that \clingo's output statements may be conditioned by several literals (cf.~\ref{sec:aspif}).
Finally, it is interesting to observe that no new literal tuple is generated for \lstinline{output(a,2)}.
Rather the one in Lines~\ref{rlp:example:rule:tri:two} and~\ref{rlp:example:rule:tri:tri} is reused.
This redundancy-free representation is a general feature of \clingo's reification.

\begin{remark}\label{rem:reify}
Although we do not detail this here, it is worth mentioning that
the reified output format of \clingo\ accounts for the full spectrum of language constructs supported by \clingo's input language
(including its generic grammar-based theory language).

In addition, \clingo\ offers the options
\lstinline{--reify-sccs} and
\lstinline{--reify-steps}
to calculate the strongly connected components of the ground logic program's (positive) dependency graph
and
to add step numbers to the reified output, in case multi-shot solving is used, respectively.
\end{remark}

\subsection{Meta encoding}
\label{sec:meta:encoding}

The facts obtained from reifying a logic program can now be used to instantiate meta encodings.
Such encodings allow us to reestablish the original or attribute a different meaning to program constructs.

\begin{table}[t]
\begin{lstlisting}[caption={A simple meta program interpreting reified logic programs (\texttt{meta.lp})},label={prg:meta},language=clingos]
conjunction(B) :- literal_tuple(B),%%#(\label{meta:conjunction}#)
      hold(L): literal_tuple(B, L), L>0;%%#(\label{meta:conjunction:hold}#)
  not hold(L): literal_tuple(B,-L), L>0.%%#(\label{meta:conjunction:end}#)

body(normal(B)) :- rule(_,normal(B)), conjunction(B).%%#(\label{meta:normal}#)
body(sum(B,G))  :- rule(_,sum(B,G)),%%#(\label{meta:sum}#)
  #sum {
    W,L:     hold(L), weighted_literal_tuple(B, L,W), L>0;%%#(\label{meta:sum:hold}#)
    W,L: not hold(L), weighted_literal_tuple(B,-L,W), L>0
  } >= G.%%#(\label{meta:sum:end}#)

  hold(A): atom_tuple(H,A)   :- rule(disjunction(H),B), body(B).%%#(\label{meta:disjunction}#)
{ hold(A): atom_tuple(H,A) } :- rule(     choice(H),B), body(B).%%#(\label{meta:choice}#)

#show.%%#(\label{meta:show}#)
#show T: output(T,B), conjunction(B).%%#(\label{meta:show:conditional}#)
\end{lstlisting}
\begin{lstlisting}[language=others,caption={Two steps system call using reification},label={call:reification}]
clingo --output=reify ezy.lp | clingo - meta.lp 0
\end{lstlisting}
\end{table}

To illustrate this,
let us start with the simple meta encoding in Listing~\ref{prg:meta},
which supports all above mentioned language constructs according to their original meaning.
This encoding is only a subset; the full encoding also accounts for optimization statements~\cite{meta-encodings}.

Before detailing how the encoding works, let us describe its usage.
To compute the stable models of the logic program \lstinline{ezy.lp} in Listing~\ref{prg:ezy} via meta programming,
we proceed in two steps.\footnote{\label{fn:warning}Adding option \lstinline{-Wno-atom-undefined} to the second call suppresses warnings due to missing definitions.
  The same effect is obtained by using \lstinline{#defined} declarations for selected predicates.}
At first, program \lstinline{ezy.lp} is reified as described above, and
then the resulting set of facts along with the meta encoding \lstinline{meta.lp}
are passed to \clingo.
The corresponding system call is given in Listing~\ref{call:reification}.
The possibility of modifying the semantics of language constructs in \lstinline{meta.lp} is paid by twice as much grounding effort.
Interestingly, keeping the semantics as done in Listing~\ref{prg:meta} results in roughly the same solver constraints,
no matter whether meta-programming is used or not.
Hence, the overall overhead of meta programming is often negligible.

\begin{remark}
We use pipes to avoid auxiliary files.
An alternative to the command in Listing~\ref{call:reification} is
\begin{lstlisting}[language=others]
clingo --output=reify ezy.lp > ezy.rlp
clingo ezy.rlp meta.lp 0
\end{lstlisting}
in which the auxiliary file~\lstinline{ezy.rlp} is used to capture the facts in Listing~\ref{prg:ezy:reified}.
Note how the use of `\lstinline{-}' in the pipe captures the output of the command before `\lstinline{|}'.
\end{remark}

Let us now turn to the actual meta encoding.
The logic program in Listing~\ref{prg:ezy} uses the unary predicate \lstinline{hold/1} to express that an atom is true.
Such atoms are derived in Lines~\ref{meta:disjunction} and~\ref{meta:choice},
provided there is an original choice or disjunctive rule,
whose body is satisfied.
Both rules use conditional literals to gather all \lstinline{hold} atoms belonging to the same atom tuple \lstinline{H},
identifying the head of the original rule.
In Line~\ref{meta:disjunction}, this results in a disjunction of atoms,
while in Line~\ref{meta:choice} all such atoms form a set of choosable atoms.
In both cases, several, one, or no \lstinline{hold} atoms may manifest themselves, depending on the size of the atom tuple.
The satisfaction of the body of the original rule, identified by \lstinline{B},
is indicated in both lines by the positive body literal \lstinline{body(B)}.
The corresponding atoms are derived by the two rules in Lines~\ref{meta:normal} to~\ref{meta:sum:end}.

In Line~\ref{meta:normal},
a \lstinline{normal} body, composed of regular literals, is satisfied
whenever all its literals are found to be true.
This is realized by the rule in Lines~\ref{meta:conjunction} to~\ref{meta:conjunction:end}
by using again conditional literals to gather
all instances of \lstinline{hold} atoms induced by a tuple of literals.
Depending on whether the integer representing a literal is positive or negative,
the corresponding \lstinline{hold} atoms must be satisfied or must not be satisfied.
Similarly, the rule in Lines~\ref{meta:sum} to~\ref{meta:sum:end} implements a weight constraint,
just that the \lstinline{hold} atoms are collected within a sum constraint along with their associated weights.
This information is extracted from instances of the ternary predicate \lstinline{weighted_literal_tuple/3}
just as with \lstinline{literal_tuple/2}.

Note that conjunctions of literal tuples may stand not only for rule bodies
but also occur in other constructs like the conditional output statement in Line~\ref{meta:show:conditional}.
Hence, it makes sense to account for them separately in Lines~\ref{meta:conjunction} to~\ref{meta:conjunction:end}.

\subsection{Examples}
\label{sec:meta:examples}

The next subsections give meta encodings computing classical, supported, here-and-there, and diverse models.
Moreover, we show how guess-and-check programming can be addressed with both meta and application programming in
Sections~\ref{sec:meta:gc} and~\ref{sec:gnt}, respectively.

\subsubsection{Classical and supported models}
\label{sec:meta:supported}

Let us start with some simple modifications to our meta encoding in Listing~\ref{prg:meta} that change the semantics of logic programs.

For illustration, we consider classical and supported models of logic programs.
Take the logic program consisting of the following three rules:
\begin{lstlisting}[numbers=none,language=clingos]
 a :- not b.    b :- c.    c :- b.
\end{lstlisting}
It has
one stable model, $\{\mathtt{a}\}$,
two supported models, $\{\mathtt{a}\}$ and $\{\mathtt{b},\mathtt{c}\}$,
and
three classical models, $\{\mathtt{a}\}$, $\{\mathtt{b},\mathtt{c}\}$ and $\{\mathtt{a},\mathtt{b},\mathtt{c}\}$.

This example already illustrates a general relationship between all three semantics:
a stable model is also a supported model, which in turn is also a classical model but not vice versa.
Intuitively, this difference is the result of
how tight each semantics relates the truth of an atom to its derivability (through rules and positive body literals).
While no such relation is imposed in the classical setting,
supported models require that each of their atoms is supported by a rule having the atom as head and a body satisfied by the model at hand.
Stable models reinforce this by stipulating that furthermore all positive body literals of the supporting rule
have themselves a supporting rule and that this ends in facts
(and thus yields a finite derivation).
In our example,
only $\mathtt{a}$ warrants this within the only stable model,
while $\mathtt{b}$ and $\mathtt{c}$ only satisfy the supportedness criterion in $\{\mathtt{b},\mathtt{c}\}$ but lack a finite derivation.
The detachment of truth from derivations (via rules) is exemplified by $\mathtt{a}$ in the classical model $\{\mathtt{a},\mathtt{b},\mathtt{c}\}$.

\begin{table}
\begin{lstlisting}[caption={Meta encoding computing classical models of logic programs (\texttt{classic.lp})},label={prg:meta:classic},language=clingos]
#include "meta.lp".%%#(\label{classic:meta}#)

atom( A ) :- atom_tuple(_,A).%%#(\label{classic:atom:begin}#)
atom(|L|) :-          literal_tuple(_,L).
atom(|L|) :- weighted_literal_tuple(_,L).%%#(\label{classic:atom:end}#)

{ hold(A) } :- atom(A).%%#(\label{classic:choice}#)
\end{lstlisting}
\begin{lstlisting}[caption={Meta encoding computing supported models of logic programs (\texttt{supported.lp})},label={prg:meta:supported},language=clingos]
conjunction(B) :- literal_tuple(B),
  not not hold(L): literal_tuple(B, L), L>0;%%#(\label{supported:conjunction:hold}#)
      not hold(L): literal_tuple(B,-L), L>0.

body(normal(B)) :- rule(_,normal(B)), conjunction(B).
body(sum(B,G))  :- rule(_,sum(B,G)),
  #sum {
    W,L: not not hold(L), weighted_literal_tuple(B, L,W), L>0;%%#(\label{supported:sum:hold}#)
    W,L:     not hold(L), weighted_literal_tuple(B,-L,W), L>0
  } >= G.

  hold(A): atom_tuple(H,A)   :- rule(disjunction(H),B), body(B).
{ hold(A): atom_tuple(H,A) } :- rule(     choice(H),B), body(B).

#show.
#show T: output(T,B), conjunction(B).
\end{lstlisting}
\end{table}

For computing classical models in ASP, we have to lift the ban of derivability from atoms.
To this end,
we extend in Listing~\ref{prg:meta:classic} our previous meta encoding (via Line~\ref{classic:meta}) with the choice rule in Line~\ref{classic:choice};
its subjects are gathered in Lines~\ref{classic:atom:begin} to~\ref{classic:atom:end}
(by extracting the atom underlying a literal \lstinline{L} by taking its absolute value \lstinline{|L|}).
This choice rule exempts atoms of predicate \lstinline{hold/1} from having a derivation by allowing for their inclusion into a stable model at will.
Classical models of a logic program can then be computed as in Listing~\ref{call:reification},
just by replacing \lstinline{meta.lp} with \lstinline{classic.lp},
given in Listing~\ref{prg:meta:classic}.
Clearly, more direct meta encodings can be devised, for instance, by turning rules into integrity constraints.

For computing supported models,
we have to make sure that each included \lstinline{hold} atom is supported by some rule.
The body of this rule must be satisfied, and its positive literals must themselves have supporting rules,
but they do not necessarily have to yield a finite derivation.
This can be accomplished by replacing the positive occurrences of \lstinline{hold} literals in Lines~\ref{meta:conjunction:hold} and~\ref{meta:sum:hold} in Listing~\ref{prg:meta}
by their double negation.
In fact, in ASP, each true positive literal must have a non-cyclic derivation, while its double negated variant is freed from this requirement.
The resulting meta encoding is given in Listing~\ref{prg:meta:supported}; cf.\ Lines~\ref{supported:conjunction:hold} and~\ref{supported:sum:hold} in both encodings.
As above,
supported models are computed by replacing \lstinline{meta.lp} by \lstinline{supported.lp} in the system call in Listing~\ref{call:reification}.

\begin{remark}\label{rem:simplification}
Note that the methods for computing classical and supported models may fall short in practice
since reification is subject to stable-model preserving simplifications
that are usually too strong for such weaker semantics.
In simple cases, like ours, this can be counterbalanced by declaring atoms as being externally defined.
For instance, adding `\lstinline{#external b.}' (cf.\ Section~\ref{sec:glance}) to our example program
spares $\mathtt{b}$ from simplification and produces the above results;
otherwise not all models are obtained.

Unfortunately,
such techniques become infeasible with programs using integers or function symbols
since they may possess infinitely many models in general.
For instance, the program consisting of
`\lstinline{q(f(a)).}'
and
`\lstinline{p(X) :- p(X).}'
has a single stable
but infinitely many supported and classical models.

Also, note that grounding may introduce auxiliary atoms that can lead to duplicate models.
To counterbalance this,
one could restrict the choice in Line~\ref{classic:choice} in Listing~\ref{prg:meta:classic} to output atoms.
\end{remark}

\subsubsection{Diverse models}
\label{sec:meta:diverse}

Our next example application of meta programming is about computing several diverse stable models of a logic program.
General approaches to computing diverse stable models can be found in the literature~\cite{eiererfi13a,roscwa16a}.

To do so within ASP rather than by external scripting,
we consider several reified stable models within one.
To this end,
we turn the predicate \lstinline{hold} into a binary predicate, whose second argument identifies the respective stable model.
These identifiers are generated in Line~\ref{many:models} of Listing~\ref{prg:meta:diverse},
where the parameter~\lstinline{m} limits the number of reified stable models.
\begin{lstlisting}[float,caption={Meta encoding computing several (diverse) stable models (\texttt{many.lp})},label={prg:meta:diverse},language=clingos]
model(1..m).%%#(\label{many:models}#)

conjunction(B,M) :- model(M), literal_tuple(B),%%#(\label{many:meta:begin}#)
      hold(L,M) : literal_tuple(B, L), L>0;
  not hold(L,M) : literal_tuple(B,-L), L>0.

body(normal(B),M) :- rule(_,normal(B)), conjunction(B,M).
body(sum(B,G),M)  :- model(M), rule(_,sum(B,G)),
  #sum {
    W,L:     hold(L,M), weighted_literal_tuple(B, L,W), L>0;
    W,L: not hold(L,M), weighted_literal_tuple(B,-L,W), L>0
  } >= G.

  hold(A,M): atom_tuple(H,A)   :- rule(disjunction(H),B), body(B,M).
{ hold(A,M): atom_tuple(H,A) } :- rule(     choice(H),B), body(B,M).

show(T,M) :- output(T,B), conjunction(B,M).
#show.
#show (T,M): show(T,M).%%#(\label{many:meta:end}#)#(\label{many:meta:show}#)

:- model(M), model(N), M<N, option=1,%%#(\label{many:diverse:k:beg}#)
   #sum {%%#(\label{many:diverse:k:sum:beg}#)
     1,T:     show(T,M), not show(T,N);
     1,T: not show(T,M),     show(T,N)
   } < k.%%#(\label{many:diverse:k:end}#)%%#(\label{many:diverse:k:sum:end}#)

#maximize { 1,T,M,N: show(T,M), not show(T,N), model(N), option=2 }.%%#(\label{many:maximize}#)
\end{lstlisting}
The following Lines~\ref{many:meta:begin} to~\ref{many:meta:end} constitute an extension of the original meta encoding obtained by
adding an additional argument to all non-structural predicates for identifying the associated stable model.
This is done throughout with variable \lstinline{M}, sometimes bound by \lstinline{model(M)}.
Taking a logic program with $n$ stable models and setting \lstinline{m} to~2
makes Lines~\ref{many:models} to~\ref{many:meta:end} produce $n^2$ stable models,
each of which comprises two stable models of the original program.
To distinguish the comprised models, Line~\ref{many:meta:show} outputs each atom with its associated model identifier.

This initial part of Listing~\ref{prg:meta:diverse} acts as a generator of combinations of \lstinline{m} stable models of the original program.
In this spirit,
the remainder selects two types of model combinations depending upon
the setting of parameter \lstinline{option} (and \lstinline{k} in the first case).
More precisely, the selected \lstinline{m} reified stable models are
\begin{itemize}
\item \lstinline{k}-diverse, if \lstinline{option=1},
  that is,
  the Hamming distance between each pair of the \lstinline{m} stable models is greater or equal than \lstinline{k}, and
\item most-diverse, if \lstinline{option=2},
  that is,
  the \lstinline{m} stable models maximize the sum of the Hamming distances between each pair of stable models.
\end{itemize}
Moreover,
the implementation considers only atoms declared to be shown
(by using predicate \lstinline{show/2} rather than \lstinline{hold/2} in Lines~\ref{many:diverse:k:beg} to~\ref{many:maximize}).

The selection of model collections having a pairwise Hamming distance greater or equal than \lstinline{k} is represented
by the sum constraint in Lines~\ref{many:diverse:k:sum:beg} and~\ref{many:diverse:k:sum:end}.
The condition is embedded into an integrity constraint ruling out all pairs of models
\lstinline{M} and \lstinline{N} that differ on less than \lstinline{k} shown atoms.
Similarly, the optimization statement in Line~\ref{many:maximize} maximizes the difference between two models;
it attributes one point per difference.
Given that such a statement exists for each pair of models the overall sum of differences is maximized.
Note that in each case the value of parameter \lstinline{option} is tested,
so that at most one of them applies.

As an example,
consider the logic program in Listing~\ref{prg:meta:diverse:example}.
\begin{lstlisting}[float,caption={Mark \texttt{c} cells of an \texttt{n}$\times$\texttt{n} grid that must be connected to each other (\texttt{cells.lp})},label={prg:meta:diverse:example},language=clingos]
{ x((1..n,1..n)) } = c.

connect(C) :- x(C), C<=C': x(C').
connect((X',Y')) :- connect((X,Y)), x((X',Y')), |X-X'|+|Y-Y'|=1.

:- x(C), not connect(C).

#show x/1.
\end{lstlisting}
The goal in this example is to mark \lstinline{c} cells of an \lstinline{n}$\times$\lstinline{n} grid
such that
the marked cells are connected to each other
(where \lstinline{c} and \lstinline{n} are parameters).

Let us begin by computing three different stable models with a pairwise Hamming distance greater or equal than one:
\begin{lstlisting}[language=shells]
UNIX> clingo --output=reify cells.lp -c n=3 -c c=3 | \
      clingo - many.lp -c option=1 -c k=1 -c m=3
clingo version 5.5.0
Reading from - ...
Solving...
Answer: 1
(x((2,3)),1) (x((3,2)),1) (x((3,3)),1) \
(x((1,3)),2) (x((2,3)),2) (x((3,3)),2) \
(x((3,1)),3) (x((3,2)),3) (x((3,3)),3)
SATISFIABLE
\end{lstlisting}
The obtained three reified stable models can be visualized as follows
(by letting (1,1) be the lower left corner):
\[
  \begin{array}{|c|c|c|}
    \cline{1-3} \phantom{\mathtt{x}}&\mathtt{x}&\mathtt{x}\\
    \cline{1-3} &&\mathtt{x}\\
    \cline{1-3} &&\\
    \cline{1-3}
  \end{array}
  \qquad
  \begin{array}{|c|c|c|}
    \cline{1-3} \mathtt{x}&\mathtt{x}&\mathtt{x}\\
    \cline{1-3} &&\\
    \cline{1-3} &&\\
    \cline{1-3}
  \end{array}
  \qquad
  \begin{array}{|c|c|c|}
    \cline{1-3} \phantom{\mathtt{x}}&\phantom{\mathtt{x}}&\mathtt{x}\\
    \cline{1-3} &&\mathtt{x}\\
    \cline{1-3} &&\mathtt{x}\\
    \cline{1-3}
  \end{array}
\]

Next, consider the result obtained by imposing a Hamming distance of \lstinline{6}:
\begin{lstlisting}[language=shells]
UNIX> clingo --output=reify cells.lp -c n=3 -c c=3 | \
      clingo - many.lp -c option=1 -c m=3 -c k=6
clingo version 5.5.0
Reading from - ...
Solving...
Answer: 1
(x((2,3)),1) (x((3,2)),1) (x((3,3)),1) \
(x((1,1)),2) (x((1,2)),2) (x((1,3)),2) \
(x((2,1)),3) (x((2,2)),3) (x((3,1)),3)
SATISFIABLE
\end{lstlisting}
Unlike above, the three solutions do not overlap and possess the imposed pairwise Hamming distance:
\[
  \begin{array}{|c|c|c|}
    \cline{1-3} \phantom{\mathtt{x}}&\mathtt{x}&\mathtt{x}\\
    \cline{1-3} &&\mathtt{x}\\
    \cline{1-3} &&\\
    \cline{1-3}
  \end{array}
  \qquad
  \begin{array}{|c|c|c|}
    \cline{1-3} \mathtt{x}&\phantom{\mathtt{x}}&\phantom{\mathtt{x}}\\
    \cline{1-3} \mathtt{x}&&\\
    \cline{1-3} \mathtt{x}&&\\
    \cline{1-3}
  \end{array}
  \qquad
  \begin{array}{|c|c|c|}
    \cline{1-3} \phantom{\mathtt{x}}&&\\
    \cline{1-3} &\mathtt{x}&\\
    \cline{1-3} &\mathtt{x}&\mathtt{x}\\
    \cline{1-3}
  \end{array}
\]
Note that no three stable models are obtainable with a pairwise Hamming distance exceeding 6.

A similar result is obtained by maximizing the sum of Hamming distances,
as shown next.
We use~\lstinline{--quiet=1,2,2} to suppress intermediate models:
\begin{lstlisting}[language=shells]
UNIX> clingo --output=reify cells.lp -c n=3 -c c=3       | \
      clingo - many.lp -c option=2 -c m=3 --quiet=1,2,2
clingo version 5.5.0
Reading from - ...
Solving...
Answer: 5
(x((2,2)),1) (x((3,1)),1) (x((3,2)),1) \
(x((1,1)),2) (x((1,2)),2) (x((2,1)),2) \
(x((1,3)),3) (x((2,3)),3) (x((3,3)),3)
OPTIMUM FOUND
\end{lstlisting}
The obtained solution has the same quality as the previous one, namely, \lstinline{18},
three times a Hamming distance of six.
Note that this property is a particularity of our example:
\[
  \begin{array}{|c|c|c|}
    \cline{1-3} \phantom{\mathtt{x}}&&\\
    \cline{1-3} &\mathtt{x}&\mathtt{x}\\
    \cline{1-3} &&\mathtt{x}\\
    \cline{1-3}
  \end{array}
  \qquad
  \begin{array}{|c|c|c|}
    \cline{1-3} &&\phantom{\mathtt{x}}\\
    \cline{1-3} \mathtt{x}&&\\
    \cline{1-3} \mathtt{x}&\mathtt{x}&\\
    \cline{1-3}
  \end{array}
  \qquad
  \begin{array}{|c|c|c|}
    \cline{1-3} \mathtt{x}&\mathtt{x}&\mathtt{x}\\
    \cline{1-3} &&\\
    \cline{1-3} &&\\
    \cline{1-3}
  \end{array}
\]

\subsubsection{Here-and-there models}
\label{sec:meta:ht}

The logical foundations of ASP rest upon the logic of Here-and-There~(\HT;~\citeNP{heyting30a}),
or more specifically its non-monotonic extension called Equilibrium Logic~\cite{pearce96a}.
Informally,
interpretations in \HT\ consist of pairs of interpretations $(H,T)$ such that $H\subseteq T$.
The intuition of using two such sets is that
atoms in $H$ are the ones that can be proved,
atoms not in $T$ are those for which no proof exists, and,
finally, atoms in $T\setminus H$ are assumed to hold but have not been proved.
A total \HT~model $(T,T)$ is an equilibrium model if there is no \HT~model $(H,T)$ with $H\subset T$.
In such a case, $T$ is also called a stable model.
A comprehensive account of \HT\ is given by~\citeANP{pearce96a}~\citeNN{pearce96a}.

The meta encoding for computing \HT~models of logic programs is given in Listing~\ref{prg:meta:ht};
it builds upon several constructions already used in the previous meta encodings.
\begin{lstlisting}[float,caption={A meta encoding for computing here-and-there models (\texttt{ht.lp})},label={prg:meta:ht},language=clingos]
atom( A ) :- atom_tuple(_,A).%%#(\label{ht:models:atoms}#)
atom(|L|) :-          literal_tuple(_,L).
atom(|L|) :- weighted_literal_tuple(_,L).

model(h).  model(t).%%#(\label{ht:models}#)

{ hold(A,h) } :- atom(A),    option=1.%%#(\label{ht:models:h}#)
{ hold(A,t) } :- atom(A).%%#(\label{ht:models:t}#)
:- hold(L,h), not hold(L,t).%%#(\label{ht:models:subseteq}#)

:- not hold(L,h), hold(L,t), option=3.%%#(\label{ht:models:supseteq}#)

conjunction(B,M) :- model(M), literal_tuple(B),%%#(\label{ht:satisfaction:begin}#)
      hold(L,M): literal_tuple(B, L), L>0;
  not hold(L,t): literal_tuple(B,-L), L>0.%%#(\label{ht:satisfaction:conjunction:negation}#)

body(normal(B),M) :- rule(_,normal(B)), conjunction(B,M).
body(sum(B,G),M)  :- model(M), rule(_,sum(B,G)),
  #sum {
    W,L:     hold(L,M), weighted_literal_tuple(B, L,W), L>0;
    W,L: not hold(L,t), weighted_literal_tuple(B,-L,W), L>0%%#(\label{ht:satisfaction:sum:negation}#)
  } >= G.

hold(A,M): atom_tuple(H,A) :- rule(disjunction(H),B), body(B,M).
hold(A,M); not hold(A,t)   :- atom_tuple(H,A),#(\label{ht:satisfaction:choice:beg}#)
                              rule(     choice(H),B), body(B,M).%%#(\label{ht:satisfaction:end}#)#(\label{ht:satisfaction:choice:end}#)

#show.
#show (T,M): output(T,B), conjunction(B,M).
\end{lstlisting}
The first part in Lines~\ref{ht:models:atoms} to~\ref{ht:models:t} is similar to
the computation of classical models in Listing~\ref{prg:meta:classic}
in generating all admissible \HT~interpretations.
As above, we use terms, namely, \lstinline{h} and \lstinline{t}, to distinguish the components in \HT~interpretations such as $(H,T)$.
Line~\ref{ht:models:h} generates atoms in $H$,
Line~\ref{ht:models:t} the ones in $T$, and
the integrity constraint in Line~\ref{ht:models:subseteq} makes sure that $H\subseteq T$.
The type of generated \HT~interpretations is once more determined by parameter \lstinline{option}.
A generated pair of interpretations $(H,T)$ is
\begin{itemize}
\item an \HT~interpretation, if \lstinline{option=1},
\item an \HT~interpretation with minimal $H\subseteq T$, if \lstinline{option=2} (or undefined), or
\item an equilibrium model, if \lstinline{option=3}.
\end{itemize}
The aforementioned description captures the first case.
In both remaining cases,
the free generation of atoms in $H$ is dropped (cf.~Line~\ref{ht:models:h}) and
they must rather be derived via program rules.
In addition, option value~\lstinline{3} enforces $T\subseteq H$ via the integrity constraint in Line~\ref{ht:models:supseteq}.

The satisfaction of rules and the derivation of \lstinline{hold} atoms
in Lines~\ref{ht:satisfaction:begin} to~\ref{ht:satisfaction:end}
is similar to  the computation of diverse models in Listing~\ref{prg:meta:diverse}
with a few exceptions due to the different semantic setting.
First of all,
a negative literal only holds in an \HT~interpretation $(H,T)$ if its atom does not belong to $T$.
Accordingly, the tests for negative body literals
in Lines~\ref{ht:satisfaction:conjunction:negation} and~\ref{ht:satisfaction:sum:negation}
only refer to \lstinline{hold} atoms associated with \lstinline{t}.
Second,
the choice of an atom~$a$ amounts in \HT\ to an instance of the law of the excluded middle $a\vee\neg a$.
This classical treatment of an atom leaves only two possibilities in Line~\ref{ht:satisfaction:choice:beg},
either $a$ is false, and thus does not belong to $T$ (and neither to $H$),
or $a$ is true in $H$ or $T$.\footnote{Strictly speaking, the rule in Lines~\ref{ht:satisfaction:choice:beg} and~\ref{ht:satisfaction:choice:end} could be specialized by adding \lstinline{M=h},
  since the instance with \lstinline{M=t} is subsumed by the rule in Line~\ref{ht:models:t},
  which is equivalent to `\lstinline{hold(A,t); not hold(A,t) :- atom(A).}'.}
Note that whenever \lstinline{option=1},
this variant of our meta encoding merely checks the satisfaction of rules in both components of an \HT~interpretation.
Unlike this,
\lstinline{hold} atoms associated with \lstinline{h} must be derived via the meta encoding in all other cases.

For illustration,
consider the logic programs
\texttt{or.lp} containing the disjunction `\lstinline{a;b.}'
and
\texttt{even.lp} with rules `\lstinline{a :- not b. b :- not a.}'.
Both programs are equivalent in the sense that they share the same stable models
$\{\mathtt{a}\}$ and $\{\mathtt{b}\}$.
However, putting each program together with rules
`\lstinline{a :- b. b :- a.}'
yields different stable models, indicating that both programs are not interchangeable in an encompassing program;
in formal terms, they are not strongly equivalent~\cite{lipeva01a}.
Interestingly, the strong equivalence of logic programs corresponds to their equivalence in \HT\
(their mere equivalence can be read off the same set of equilibrium models).

Let us verify this by means of our meta encoding in Listing~\ref{prg:meta:ht}.

Program \lstinline{or.lp} has five \HT~models:
\begin{lstlisting}[language=shells]
UNIX> clingo --output=reify or.lp | \
      clingo - ht.lp 0 -c option=1
clingo version 5.5.0
Reading from - ...
Solving...
Answer: 1
(b,t) (b,h)
Answer: 2
(b,t) (a,t) (b,h)
Answer: 3
(a,t) (a,h)
Answer: 4
(b,t) (a,t) (a,h)
Answer: 5
(b,t) (a,t) (b,h) (a,h)
SATISFIABLE
\end{lstlisting}
In contrast to `\lstinline{a;b.}', program \lstinline{even.lp} has the additional \HT~model
\(
(\emptyset,\{\mathtt{a},\mathtt{b}\})
\),
as witnessed by the first of its six \HT~models:
\begin{lstlisting}[language=shells]
UNIX> clingo --output=reify even.lp | \
      clingo - ht.lp 0 -c option=1
clingo version 5.5.0
Reading from - ...
Solving...
Answer: 1
(a,t) (b,t)
Answer: 2
(a,t) (a,h)
Answer: 3
(a,t) (b,t) (a,h)
Answer: 4
(b,t) (b,h)
Answer: 5
(a,t) (b,t) (b,h)
Answer: 6
(a,t) (b,t) (a,h) (b,h)
SATISFIABLE
\end{lstlisting}

Unlike above, both programs have the same equilibrium models,
which establishes their equivalence.
This can be verified using option value \lstinline{3} instead of \lstinline{1} above.

Further examples of meta programming are available online~\cite{meta-encodings}.

\subsection{Guess-and-check programming}
\label{sec:meta:gc}

So far, we addressed problems at the first level of the polynomial hierarchy, sharing the same complexity as normal logic programs in ASP~\cite{daeigovo01a}.
In fact, ASP can also be used for expressing problems at the second level, when disjunctive heads or non-monotonic aggregates are used.

An interesting class of such problems consists of two subproblems~\cite{eitpol06a}:
A guess-and-check logic program is a pair $(P,Q)$ of normal logic programs whose
solution is a stable model of $P$ that results in an unsatisfiable program once its atoms are added as facts to $Q$.
This combines a satisfiability problem with an unsatisfiability problem,
in the most interesting case, an \textit{NP} with a \textit{coNP} problem
since this lifts the joined problem to the second level of the polynomial hierarchy.
An example of this is preference handling,
where the first problem defines feasible solutions, while the second one ensures that
there are no better solutions.
Another example is (bounded) conformant planning under incomplete information,
where the first problem gives a plan in some scenario,
while the second one makes sure that it is not invalidated in any other scenario.
We are interested in finding a single ASP encoding using disjunctive heads that
combines both problems and
yields solutions despite the unsatisfiability of the second subproblem.

For implementing such problems, \citeANP{eitgot95a}~\citeNN{eitgot95a} invented
the \textit{saturation} technique, using the elevated complexity of disjunctive logic programming.
In stark contrast to the ease of common ASP modeling,
however,
this technique is rather involved and hardly usable by ASP laymen.
This shortcoming was addressed by means of meta programming by \citeANP{eitpol06a}~\citeNN{eitpol06a}.
The idea is to represent both components of a guess-and-check program as facts and
to combine them with a meta encoding to obtain a single joint program after grounding.
The meta encoding implements the saturation technique and discharges users from this intricate modeling task.
Now, the reification functionalities of \clingo\ allow us to simplify this even further
by relieving users from the specification of guess-and-check program in terms of facts.

In what follows,
we showcase the computation of solutions to guess-and-check programs by means of meta programming with \clingo.
To this end,
let us start with a sketch of the meta encoding implementing saturation and
tie it up to guess-and-check programming afterwards.
We build on an encoding of saturation put forward by \citeANP{gekasc11b}~\citeNN{gekasc11b}
and present in Listing~\ref{prg:meta:D} in \ref{sec:meta:D} a revised version based on the \aspif\ format.\footnote{The original meta encoding~\cite{gekasc11b} relies on the \smodels\ format
  used for reification in \clingo~4 (cf.\ Footnote~\ref{fn:reification:clingo:four}).}
A detailed account of the saturation technique goes beyond the scope of this tutorial
(see the papers by \citeANP{eitgot95a}~\citeNN{eitgot95a} and \citeANP{eitpol06a}~\citeNN{eitpol06a} for details).
From the perspective of plain ASP,
a saturation-based meta encoding acts as an ordinary one just that it yields the set of all atoms whenever a
program is unsatisfiable.
That is, it generates one stable model for each original stable model, whenever the program is satisfiable,
and otherwise, it yields a unique stable model containing the set of all atoms of the program.
Also, unlike Listing~\ref{prg:meta}, the encoding in Listing~\ref{prg:meta:D} takes a reified normal logic program as
input and results in a disjunctive logic program after grounding.

Now, for implementing guess-and-check programming,
we exploit the property of saturation-based meta encodings that
the non-existence of stable models results in a set containing all atoms.
In such a case, the encoding in Listing~\ref{prg:meta:D} additionally produces
the special-purpose atom \lstinline{bot} for indicating unsatisfiability;
this atom never appears in a genuine stable model.
Hence, to make sure that an (augmented) check program is unsatisfiable,
we can simply add an integrity constraint that enforces \lstinline{bot} to hold (cf. Listing~\ref{prg:meta:bot}).

What remains to be done is to account for the import of the guessed atoms into the check program and
to align the stable models of the guess and the check program.
For simplicity, we address both tasks in a rather direct way and show below a more principled alternative.
For addressing the first task,
we simply add a choice rule over the possible guess atoms to the check program
to allow for exchanging all possible interpretations.
For this to work, however, the guess atoms must not occur among the heads of the check program.
Otherwise, a guess atom may be false but become true in the check program.
For illustration, consider the simple guess and check programs in Listings~\ref{prg:meta:guess} and~\ref{prg:meta:check}
along with the import of guess atoms into the check program in Listing~\ref{prg:meta:in}.

\begin{table}[t]
\begin{minipage}[b]{.5\linewidth}
\begin{lstlisting}[caption={Guess program (\texttt{guess.lp})},label={prg:meta:guess},numbers=none,language=clingos]
1 { a(1..2) }.
\end{lstlisting}
\end{minipage}\begin{minipage}[b]{.5\linewidth}
\begin{lstlisting}[caption={Check program (\texttt{check.lp})},label={prg:meta:check},numbers=none,language=clingos]
:- not a(1).
\end{lstlisting}
\end{minipage}\\
\begin{minipage}[b]{.5\linewidth}
\begin{lstlisting}[caption={Import guess atoms (\texttt{in.lp})},label={prg:meta:in},numbers=none,language=clingos]
{ a(1..2) }.
\end{lstlisting}
\end{minipage}\begin{minipage}[b]{.5\linewidth}
\begin{lstlisting}[caption={Enforce \lstinline{bot} atom (\texttt{bot.lp})},label={prg:meta:bot},numbers=none,language=clingos]
:- not bot.
\end{lstlisting}
\end{minipage}
\begin{lstlisting}[caption={Synchronize stable models (\texttt{glue.lp})},label={prg:meta:glue},numbers=none,language=clingos]
bot :- output(a(X),B),     a(X), fail(normal(B)).
bot :- output(a(X),B), not a(X), true(normal(B)).
\end{lstlisting}
\end{table}

The second task, synchronizing the respective stable models, is handled by saturation.
Remember that the check program uses a fresh set of guess atoms derived via the choice rule in Listing~\ref{prg:meta:in}.
Hence, we have two different sets of atoms in both the guess and the check programs, which have to be synchronized.
In valid counter examples, the truth values of both sets of atoms have to agree.
We simply derive the \lstinline{bot} in Listing~\ref{prg:meta:glue} if the atoms differ,
which triggers saturation and discards invalid candidates for counter examples.

The body literals of the rules in Listing~\ref{prg:meta:glue} already hint at the design decision to only reify the check program,
while leaving the guess program intact.
At last, let us put all this together for our example guess and check program,
where program \lstinline{metaD.lp} is given in Listing~\ref{prg:meta:D}:
\begin{lstlisting}[language=shells]
UNIX> clingo --output=reify --reify-sccs check.lp in.lp | \
      clingo - metaD.lp bot.lp glue.lp guess.lp 0
clingo version 5.5.0
Reading from - ...
Solving...
Answer: 1
a(2)
SATISFIABLE
\end{lstlisting}
Combining program \lstinline{guess.lp} and \lstinline{check.lp}
from Listings~\ref{prg:meta:guess} and~\ref{prg:meta:check} in a guess-and-check program
eliminates all stable models of \lstinline{guess.lp} that contain \lstinline{a(1)}.
Note that joining both in a regular program eliminates models excluding~\lstinline{a(1)}.

The first call to \clingo\ merely reifies program \lstinline{check.lp},
so that it gets interpreted by the meta encoding \lstinline{metaD.lp} in the second call.
Unlike this, program \lstinline{guess.lp} remains untouched.
Hence, the problem handed to the second call of \clingo\ combines both the guess and the check program,
whereby the latter is encoded via saturation in order to succeed upon unsatisfiability.
And finally, program \lstinline{glue.lp}
is in charge of aligning the guessed stable models to the checker
by saturating the non-aligned ones.
Clearly, our illustrative example is very simple and could also be solved directly.
In what follows, we present some more substantial use-cases.

\begin{remark}
It is instructive to observe the effect of programs \lstinline{bot.lp} and \lstinline{glue.lp}
on the formation of the joint solving result:
\begin{itemize}
\item
Without both \lstinline{bot.lp} and \lstinline{glue.lp},
we get all combinations of
stable models of the guess program
with stable models of the check program (choice rules included), if the check program is satisfiable, and
with the unique saturated set of atoms containing \lstinline{bot}, otherwise.
\item
Without \lstinline{bot.lp} but with \lstinline{glue.lp},
we get all combinations of
stable models of the guess program
with stable models of the check program (conjoined with all facts stemming from guess atoms),
if this program is satisfiable, and
with the unique saturated set of atoms containing \lstinline{bot}, otherwise.
\item
With both \lstinline{bot.lp} and \lstinline{glue.lp},
we get all stable models of the previous item that contain \lstinline{bot}.
That is,
we get all combinations of
stable models of the guess program
with the unique saturated set of atoms containing \lstinline{bot} whenever
the check program (conjoined with all facts stemming from guess atoms) is unsatisfiable,
and no stable models, otherwise.
\end{itemize}
\end{remark}

A more generic approach can be obtained by capturing the guessed atoms by a dedicated predicate, say \lstinline{guess/1}.
This also eases the restriction of the relayed atoms to a distinguished subset.

\begin{table}[t]
\begin{minipage}[b]{.5\linewidth}
\begin{lstlisting}[caption={Export guess atoms (\texttt{out.lp})},label={prg:meta:guess:out},language=clingos,numbers=none]
guess(a(X)) :- a(X).%%#(\label{meta:guess:out}#)
\end{lstlisting}
\end{minipage}\begin{minipage}[b]{.5\linewidth}
\begin{lstlisting}[caption={Import guess atoms (\texttt{in.lp})},label={prg:meta:guess:in},language=clingos,numbers=none]
{ guess(X) } :- output(guess(X),_).%%#(\label{meta:guess:in}#)
\end{lstlisting}
\end{minipage}\\
\begin{lstlisting}[caption={Generic synchronization of stable models (\texttt{superglue.lp})},label={prg:meta:glue:super},language=clingos,numbers=none]
bot :- output(guess(X),B),     guess(X), fail(normal(B)).%%#(\label{meta:superglue:one}#)
bot :- output(guess(X),B), not guess(X), true(normal(B)).%%#(\label{meta:superglue:two}#)
\end{lstlisting}
\end{table}
In our example, the guess program in Listing~\ref{prg:meta:guess}
is then augmented with the specification of the actual guess in Listing~\ref{prg:meta:guess:out}.
Accordingly,
the guessed atoms become enveloped in predicate \lstinline{guess/1} in the check program from Listing~\ref{prg:meta:check}, viz.
`\lstinline[language=clingos]{:- not guess(a(1)).}'.
Similarly, the import of guessed atoms is modified in the program in Listing~\ref{prg:meta:guess:in}.
This version of program~\lstinline{in.lp} replaces the
specific choice rule `\lstinline!{ a(1..2) }.!' in Listing~\ref{prg:meta:in}
by a generic rule.
The instantiation of the first rule is obtained from the symbol table of the guess program,
as shown below.
With this approach, both the guess as well as the check programs come in pairs,
the former accompanied by the specification of guessed atoms
in \lstinline{out.lp} in Listing~\ref{prg:meta:guess:out},
and the latter with choices delineating all possible sets of guessed atoms
in \lstinline{in.lp} in Listing~\ref{prg:meta:guess:in}.
While the first fixes all guessable atoms, the second one is generic.
And finally, also the synchronization of stable models can now be expressed in a generic way via the rules in Listing~\ref{prg:meta:glue:super}.
Note that except for Listing~\ref{prg:meta:guess:out}, all auxiliary programs are problem independent.

Putting everything together, we can solve the example in Listings~\ref{prg:meta:guess} to~\ref{prg:meta:check}:
\begin{lstlisting}[language=shells]
UNIX> clingo --output=reify                                      \
             guess.lp out.lp <(echo "#show guess/1.")          | \
      grep "output(guess(.*))"                                 | \
      clingo --output=reify --reify-sccs                         \
             - check.lp in.lp                                  | \
      clingo - metaD.lp bot.lp superglue.lp guess.lp out.lp 0
clingo version 5.5.0
Reading from - ...
Solving...
Answer: 1
a(2)
SATISFIABLE
\end{lstlisting}

The purpose of the first two lines in the system call above is to extract all guessable atoms from the symbol table of the guess program.
(In passing, this illustrates a pragmatic way of exploiting the symbol table of reified programs.)
In our example, this results in the facts \lstinline{output(guess(a(1)),5).} and \lstinline{output(guess(a(2)),6).}
These atoms are then used to instantiate the rules in Listing~\ref{prg:meta:guess:in} and~\ref{prg:meta:glue:super}.
Otherwise, the two remaining calls work just as described above, and obviously yield the same result.

In fact, the effort of encapsulating the guessed atoms along with the resulting generality pays off,
as we demonstrate in the following two use-cases.
More examples of guess-and-check programming are available online \cite{meta-encodings}.

\paragraph{Preferences.}In this example,
we use the above setup to compute subset maximal stable models of logic programs.
The idea is to guess a candidate stable model and
to check whether any of its proper supersets is a stable model of the program, too.
If this fails, the candidate is subset maximal.

To illustrate this,
reconsider program~\lstinline{guess.lp} in Listing~\ref{prg:meta:guess} along with
the specification of guessable atoms in Listing~\ref{prg:meta:guess:out}.
The check program consists once more of program \lstinline{guess.lp}
yet extended with the rules in Listing~\ref{prg:meta:superset}.

\begin{table}
\begin{lstlisting}[caption={Identifying subset maximal stable models (\texttt{superset.lp})},label={prg:meta:superset},language=clingos]
better :- a(Y), not guess(a(Y)), a(X): guess(a(X)).
:- not better.%%#(\label{meta:superset:not:better}#)
\end{lstlisting}
\begin{minipage}[b]{0.5\linewidth}
\begin{lstlisting}[caption={Player O (\texttt{playero.lp})},label={prg:meta:tictactoe:player:o},language=clingos]
#const n=3.
n { o(1..n,1..n) } n.
:- not win.
win :- I=1..n, o(I,J): J=1..n.
win :- J=1..n, o(I,J): I=1..n.
win :- o(I,I): I=1..n.
win :- o(I,n+1-I): I=1..n.

#show o/2.
\end{lstlisting}
\end{minipage}\begin{minipage}[b]{0.5\linewidth}
\begin{lstlisting}[caption={Player X (\texttt{playerx.lp})},label={prg:meta:tictactoe:player:x},language=clingos,numbers=right]
#const n=3.
n { x(1..n,1..n) } n.
:- not win.
win :- I=1..n, x(I,J): J=1..n.
win :- J=1..n, x(I,J): I=1..n.
win :- x(I,I): I=1..n.
win :- x(I,n+1-I): I=1..n.

:- guess(o(I,J)), x(I,J).%%#(\label{meta:ttt:o:blocked}#)
\end{lstlisting}
\end{minipage}
\begin{lstlisting}[caption={Exporting guess atoms (\texttt{out.lp})},label={prg:meta:tictactoe:out},language=clingos]
guess(o(I,J)) :- o(I,J).
\end{lstlisting}
\end{table}

The atom \lstinline{better} is derived whenever the stable model generated by the check program is a proper superset of
the guessed stable model.
This candidate model is not subset maximal if some strictly larger model is obtainable,
as checked in Line~\ref{meta:superset:not:better} of Listing~\ref{prg:meta:superset}.

What makes this example different from the one above (and below) is that the guess and the check programs
are based on the same logic program and thus deal with the same set of stable models.
A guessed stable model is optimal, if no better model can be obtained by the checker.
This nicely reflects the checker's role of generating potential counterexamples.

Proceeding in the same manner as above,
we compute the only subset maximal model of program \lstinline{guess.lp} as follows:
\begin{lstlisting}[language=shells]
UNIX> clingo --output=reify guess.lp out.lp                      \
             <(echo "#show guess/1.")                          | \
      grep "output(guess(.*))"                                 | \
      clingo --output=reify --reify-sccs                         \
             - guess.lp superset.lp in.lp                      | \
      clingo - metaD.lp bot.lp superglue.lp guess.lp out.lp 0
clingo version 5.5.0
Reading from - ...
Solving...
Answer: 1
a(1) a(2)
SATISFIABLE
\end{lstlisting}
\paragraph{Tic-tac-toe.}In this example, we consider a simplified $3\times3$ Tic-tac-toe puzzle.
Player~O has to place their three tokens in a winning configuration such that
afterwards Player~X cannot place theirs in a winning position.
Hence, the game is not played in turns and
just guaranteed winning configurations of Player~O are determined.

This example involves two similar yet different programs.
Unlike above, a guess involves only a subset of the atoms of the generated stable models.

The encodings of Player~O and~X are given in Listings~\ref{prg:meta:tictactoe:player:o}
and~\ref{prg:meta:tictactoe:player:x}.
The former acts as a guess program and the latter as a checker program;
only the positions of Player~O are passed to Player~X,
as fixed in Listing~\ref{prg:meta:tictactoe:out}.
In this way,
the positions occupied by~O become blocked for Player~X in Line~\ref{meta:ttt:o:blocked} of
Listing~\ref{prg:meta:tictactoe:player:x}.

Relying on the above setup,
we can then compute the two undefeatable diagonal configurations in Tic-tac-toe:
\begin{lstlisting}[language=shells]
UNIX> clingo --output=reify playero.lp out.lp                     \
             <(echo "#show guess/1.")                           | \
      grep "output(guess(.*))"                                  | \
      clingo --output=reify --reify-sccs                          \
             - playerx.lp in.lp                                 | \
      clingo - metaD.lp bot.lp superglue.lp playero.lp out.lp 0
clingo version 5.5.0
Reading from - ...
Solving...
Answer: 1
o(1,1) o(2,2) o(3,3)
Answer: 2
o(1,3) o(2,2) o(3,1)
SATISFIABLE
\end{lstlisting}

\section{About \clingo\ applications}
\label{sec:app}

We have seen in the last section how the functionality of ASP systems can be changed by using meta programming.
In particular,
the reification of logic programs allows us to control ASP by means of ASP.
The remainder of this tutorial parallels this by showcasing several ways of how ASP can be managed with other programming languages.
As mentioned in the introduction, we have chosen \python\ as our example language, although other choices are possible.

This section focuses on the overall setup.
The following ones delve into particular functionalities and case studies.

\Clingo\ offers three ways of combining ASP with other programming languages,
either via
an embedded script,
module import, or
its application class.
Although all three options allow us to change the behavior of \clingo\ by overwriting its main function,
they aim at rather different use cases.
We discuss the three choices below and show how they treat the common example in Listing~\ref{lst:extending:example}.
\begin{lstlisting}[
  float=ht,
  label={lst:extending:example},
  language=clingos,
  basicstyle=\small\ttfamily,
  caption={Example with external function (\lstinline{example.lp})}]
num(3).
num(6).
div(N,@divisors(N)) :- num(N).%%#(\label{ex:divisor}#)
\end{lstlisting}
The idea is to outsource the computation of divisors to \python\ via an external function call.
Such calls look like function terms but are preceded by \lstinline{@}~\cite{PotasscoUserGuide}.
In our example,
the term \lstinline{@divisors(N)} in Line~\ref{ex:divisor} assumes the definition of a corresponding method in \python\
that takes the instantiation of \lstinline{N} as argument;
its results are added as a term pool~\cite{gehakalisc15a}
so that one or several values can be accommodated.
In our example, the head of the third rule thus results in the atoms
\lstinline{div(3,(1;3))} and \lstinline{div(6,(1;2;3;6))}.

\subsection{Embedded \python\ code}
\label{sec:app:script}

The simplest way to extend an ASP encoding with a \python\ method is to add it as an embedded script.

In our example, this can be done by supplying \clingo\ with the embedded \python\ script in Listing~\ref{lst:extending:embedded}.
\begin{lstlisting}[
  float=ht,
  label={lst:extending:embedded},
  language=clingos,
  commentstyle={},caption={Embedded Python code (\lstinline{embedded.lp})}]
#script (python)

from clingo.symbol import Number

def divisors(a):
    a = a.number
    for i in range(1, a+1):
        if a % i == 0:
          yield Number(i)

#end.
\end{lstlisting}
It shows that foreign language scripts are enclosed in
\lstinline{#script} and \lstinline{#end.} and supplied with an argument indicating the used language.
The code in this block is arbitrary and executed before grounding.
Functions defined in it, such as the \lstinline{divisors} function,
can be called from ASP by prepending an \lstinline{@} symbol.
Finally,
such scripts are meant to be part of the input of \clingo\ just as the encoding in Listing~\ref{lst:extending:example},
as shown below:
\begin{lstlisting}[language=shell]
UNIX> clingo example.lp embedded.lp
clingo version 5.5.0
Reading from example.lp ...
Solving...
Answer: 1
num(3) num(6) div(3,1) div(3,3) div(6,1) div(6,2) div(6,3) div(6,6)
SATISFIABLE
\end{lstlisting}
Such external term evaluation (during grounding) is an intended use case for embedded scripts.
The \clingo\ object invoked on the command line is in charge of loading, grounding, and solving,
unless these tasks are taken from it~\cite{gekakasc17a}.
Hence, the usage of embedded scripts is generally most suitable for small extensions to logic programs,
anything on the term level during grounding.
Often they are used to perform calculations that are difficult or inconvenient to express in ASP.

\subsection{The \clingo\ \python\ module}
\label{sec:app:module}

The second alternative is to write a \python\ script using \clingo's \python\ module.
The module provides high level functions to interact with the grounder and solver
including input and output processing as well as fine-grained control over the grounding and solving process.
This provides a convenient way to use \clingo\ as part of a larger project.
The surrounding application is in charge of the control flow and ASP is used to perform specific computations.
Even for simple computations, this avoids error prone string processing
like transforming data into ASP facts or parsing the solver's output.

Listing~\ref{lst:extending:module} implements the previous example by using the \python\ module.
\begin{lstlisting}[
  float=ht,
  literate={\%\%}{}{0},
  escapeinside={\#(}{\#)},
  label={lst:extending:module},
  language=pythons,
  caption={Example with external function (\lstinline{module.py})}]
from clingo.symbol import Number
from clingo.control import Control

class ExampleApp:
    @staticmethod
    def divisors(a):
        a = a.number
        for i in range(1, a+1):
            if a % i == 0:
              yield Number(i)

    def run(self):
        ctl = Control()
        ctl.load("example.lp")
        ctl.ground([("base", [])], context=self)
        ctl.solve(on_model=print)

if __name__ == "__main__":
    ExampleApp().run()
\end{lstlisting}
The main difference is that we have to construct a \lstinline{Control} object and take care of the control flow ourselves.
Such a \lstinline{Control} object encapsulates an instance of \clingo.
First, the program is added to the \lstinline{Control} object using its \lstinline{load} function.
Then, the \lstinline{base} part is grounded.\footnote{Without any declarations, rules belong to a logic program referred to as \lstinline{base}
  (cf.\ Section~\ref{sec:glance}).}
To be able to call the \lstinline{divisors} function, we pass the self argument as context to the ground function.
Finally, note that we use \python's \lstinline{print} function as \lstinline{on_model} callback.
An \lstinline{on_model} callback is a function passed to \lstinline{solve} that is called for each model.
It allows for inspecting (and printing) the current model.
This construction is necessary because, unlike with the \clingo\ system,
there is no output foreseen when using the \clingo\ module in \python.
This is nicely reflected by the plain output produced by \python's \lstinline{print} function
when solving our example with Listing~\ref{lst:extending:module}:
\begin{lstlisting}[language=shell]
UNIX> python module.py
num(3) num(6) div(3,1) div(3,3) div(6,1) div(6,2) div(6,3) div(6,6)
\end{lstlisting}
This is different from the ASP-specific output produced by \clingo\ in the previous section.
While there ASP is extended with \python, here it is the other way around.
This renders the use of ASP completely opaque.

\subsection{Implementing a system based on \clingo}
\label{sec:app:app}

Finally, we present a third way that aims at building custom systems based on \clingo.
This is similar to embedded \python\ code but gives more control to customize the system.
For example, parts of the text output can be modified, additional options can be registered, or the way input files are
treated can be changed completely.

Unlike the previous example,
we now derive our \lstinline{ExampleApp} class from \clingo's application class.
The resulting class can then be used with the \lstinline{clingo_main} function,
which starts a process similar to the one in \clingo\ but possibly with some overwritten functions.
First, the program name and version are defined.
These values are then used in the status, help, and version output of \clingo.
Furthermore, each application class must implement a main function,
which is called right after option parsing and is in charge of the grounding and solving process.
The function receives a \lstinline{Control} object and a list of paths to the files passed on the command line.
The subsequent code processes the files just as \clingo\ would.
Similar to the previous example, we pass in Listing~\ref{lst:extending:app} the \lstinline{ExampleApp} object to the
ground function to be able to call its \lstinline{divisors} method during grounding.
\begin{lstlisting}[
  float=ht,
  label={lst:extending:app},
  language=pythons,
  caption={Example application (\lstinline{app.py})}]
import sys
from clingo.symbol import Number
from clingo.application import Application, clingo_main

class ExampleApp(Application):
    program_name = "example"
    version = "1.0"

    @staticmethod
    def divisors(a):
        a = a.number
        for i in range(1, a+1):
            if a % i == 0:
                yield Number(i)

    def main(self, ctl, files):
        for path in files: ctl.load(path)
        if not files:
            ctl.load("-")
        ctl.ground([("base", [])], context=self)
        ctl.solve()

if __name__ == "__main__":
    clingo_main(ExampleApp(), sys.argv[1:])
\end{lstlisting}

Note that we do not need to use a print function to output models.
Rather the one of \clingo\ is used in a seamless way.
Again, this is reflected by the output of solving our example with Listing~\ref{lst:extending:app}:
\begin{lstlisting}[language=shell]
UNIX> python app.py example.lp
example version 1.0
Reading from example.lp
Solving...
Answer: 1
num(3) num(6) div(3,1) div(3,3) div(6,1) div(6,2) div(6,3) div(6,6)
SATISFIABLE
\end{lstlisting}
Furthermore, this output already hints at the benefits of using \clingo's \lstinline{Application}.
While control is exercised from \python, as in the last section,
it allows us to draw on \clingo's infrastructure, similar to using embedded scripting.

The utility of this class becomes apparent in the rest of the tutorial,
where it is used throughout as the basic building block of all \clingo\ based systems.

\section{Multi-shot ASP solving}
\label{sec:multi}

Multi-shot solving allows for solving continuously changing logic programs in an operative way.
This can be controlled via APIs implementing reactive procedures that loop on grounding and solving while reacting,
for instance, to outside changes or previous solving results.
Such reactions may entail the addition or retraction of rules that \clingo's operative approach can accommodate
while leaving unaffected program parts intact within the solver.
This avoids re-grounding and solving benefits from heuristic scores and constraints learned over time.

We begin with an informal overview of the central features and language constructs of \clingo's multi-shot solving capabilities.
We illustrate them in Sections~\ref{sec:optimization} and~\ref{sec:iclingo} by showcasing two exemplary reasoning modes,
namely branch-and-bound-based optimization and incremental ASP solving.
A comprehensive introduction to multi-shot solving with \clingo\ is given by~\citeANP{gekakasc17a}~\citeNN{gekakasc17a}.

\subsection{A gentle introduction}
\label{sec:glance}

\Clingo\ allows us to structure (non-ground) rules into subprograms.
To this end,
a program can be partitioned into several subprograms by means of the directive \lstinline{#program};
it comes with a name and an optional list of parameters.
Once given in the input,
the directive gathers all rules up to the next such directive (or the end of file)
within a subprogram identified by the supplied name and parameter list.
As an example,
we specify two subprograms \lstinline{base} and \lstinline{acid(k)} in file \lstinline{chemistry.lp} in Listing~\ref{lst:chemistry}.
\begin{lstlisting}[float=h,language=clingos,label={lst:chemistry},caption={Subprograms \lstinline{base} and \lstinline{acid(k)} (\lstinline{chemistry.lp})}]
a(1).
#program acid(k).
b(k).
c(X,k) :- a(X).
#program base.#(\label{lst:chemistry:base}#)
a(2).
\end{lstlisting}
Note that \lstinline{base} is a special subprogram (with an empty parameter list).
In addition to the rules in its scope,
it gathers all rules not preceded by any \lstinline{#program} directive.
Hence, in the above example, the \lstinline{base} subprogram includes the facts \lstinline{a(1)} and \lstinline{a(2)},
although, only the latter is in the actual scope of the directive in Line~\ref{lst:chemistry:base}.
Without further control instructions (see below),
\clingo\ grounds and solves the \lstinline{base} subprogram only,
essentially, yielding the standard behavior of ASP systems.
The processing of other subprograms such as \lstinline{acid(k)} is subject to external governance.

We first have a look at customizing grounding and solving by creating a \lstinline{Control} object,
as put forward in Section~\ref{sec:app:module}.
For illustration, let us consider two \python\ code snippets:\footnote{The \lstinline{ground} routine takes a list of pairs as argument.
  Each such pair consists of a subprogram name (e.g.\ \lstinline{base} or \lstinline{acid}) and a list of actual parameters
  (e.g.\ \lstinline{[]} or \lstinline{[Number(42)]}).}
\begin{lstlisting}[language=pythons]
from clingo.control import Control
ctl = Control()
ctl.load("chemistry.lp")
ctl.ground([("base", [])])
ctl.solve(on_model=print)
\end{lstlisting}
While the above control program matches the default behavior of \clingo,
the one below ignores all rules in the \lstinline{base} program but rather
contains a \lstinline{ground} instruction for \lstinline{acid(k)} in Line~\ref{lst:chemistry:base:ground},
where the parameter~\lstinline{k} is to be instantiated with the term \lstinline{42}.
\begin{lstlisting}[language=pythons]
from clingo.symbol import Number
from clingo.control import Control
ctl = Control()
ctl.load("chemistry.lp")
ctl.ground([("acid",[Number(42)])])#(\label{lst:chemistry:base:ground}#)
ctl.solve(on_model=print)#(\label{lst:chemistry:acid:solve}#)
\end{lstlisting}
The treatment of parameter~\lstinline{k} is similar to that of a constant,
defined with \lstinline{#const}~\cite{PotasscoUserGuide},
yet restricted to the rules in the scope of the respective subprogram.
Accordingly, the schematic fact \lstinline{b(k)} is turned into \lstinline{b(42)}.
No ground rule is obtained from `\lstinline{c(X,k) :- a(X)}' due to lacking instances of \lstinline{a(X)}.
Hence, the \lstinline{solve} call in Line~\ref{lst:chemistry:acid:solve} yields a stable model consisting of
\lstinline{b(42)} only.
Note that \lstinline{ground} instructions apply to the subprograms given as arguments,
while \lstinline{solve} triggers reasoning with respect to all accumulated ground rules.

For more elaborate reasoning processes,
it is indispensable to activate and/or deactivate ground rules on demand.
For instance, former initial or goal state conditions may need to be
relaxed or completely replaced when modifying a planning problem, e.g.,
by extending the plan length.
To expire transient rules,
\clingo\ provides the \lstinline{#external} directive.
This directive goes back to \lparse~\cite{lparseManual},
where it was used to exempt (input) atoms from simplifications during grounding.
Its functionality is generalized in \clingo\
to provide a flexible handling of yet undefined atoms in the course of grounding and solving.

For continuously assembling ground rules evolving at different stages of a reasoning process,
\lstinline{#external} directives declare atoms that may still be defined by rules added later on.
In terms of module theory~\cite{oikjan06a}
such atoms correspond to inputs, which (unlike undefined output atoms) must not be simplified.
For declaring such input atoms,
\clingo\ offers schematic \lstinline{#external} directives that are instantiated along with
the rules of their respective subprograms.

For instance, the directive in the second line below
\begin{lstlisting}[firstnumber=8,language=clingos]
#program acid(k).
#external d(X,k): c(X,k).
e(X,k) :- d(X,k).
\end{lstlisting}
is treated similar to the rule
`\lstinline{d(X,k) :- c(X,k)}'
during grounding,
just that only the head atoms of the resulting ground instances are collected as inputs.
Hence, adding the above lines to program \lstinline{chemistry.lp} and grounding both subprograms~\lstinline{base} and~\lstinline{acid(42)} yields
the external atoms~\lstinline{d(1,42)} and~\lstinline{d(2,42)}.
Thus, we furthermore obtain the ground rules~`\lstinline{e(1,42) :- d(1,42)}' and~`\lstinline{e(2,42) :- d(2,42)}'.

Once grounded, the truth value of external atoms can be changed via \clingo's API
(until the atoms become defined by corresponding rules or are released).
By default, the initial truth value of external atoms is set to false.
Then, for example, with \clingo's \python\ API,
the call~\lstinline{ctl.assign_external(d(2,42),True)}\footnote{For constructing atoms, symbolic terms, or function terms, respectively,
  the \clingo\ API function \lstinline{Function} has to be used.
  Similarly, numeric terms are constructed using function \lstinline{Number}.
  Hence, the expression \lstinline{d(2,42)} actually stands for \lstinline{Function("d", [Number(2),Number(42)])}.}
can be used to set the truth value of the external atom \lstinline{d(2,42)} to true.
This can be used to activate and deactivate rules in logic programs.
For instance,
the rule `\lstinline{e(1,42) :- d(1,42)}' is ineffective because \lstinline{d(1,42)} is false by default.
Hence, a subsequent \lstinline{solve} call yields the stable model consisting of
atoms~\lstinline{a(1)}, \lstinline{a(2)}, \lstinline{c(1,42)}, \lstinline{c(2,42)}, \lstinline{d(2,42)}, and~\lstinline{e(2,42)}.
One further interesting use case is to release external atoms.
The call \lstinline{ctl.release_external(d(1,42))} removes the external atom and the rule `\lstinline{e(1,42) :- d(1,42)}' from the \lstinline{Control} object.

\begin{remark}
Module theory~\cite{oikjan06a} is used to characterize the composition of ground subprograms during multi-shot solving.
For this, each ground subprogram is associated with a module.
Accordingly, the restrictions of module composition apply:
First, no two subprograms may define the same atom.
Second, loops cannot spread across subprograms.
We refer to the paper by \citeANP{gekakasc17a}~\citeNN{gekakasc17a} for details.

The first condition can be enforced by equipping rule heads with parameters, as done above.
Often a natural choice for this is an argument identifying a step, as \lstinline{t} in Listing~\ref{fig:toh:enc}.
\end{remark}

\subsection{Branch-and-bound-based optimization}
\label{sec:optimization}

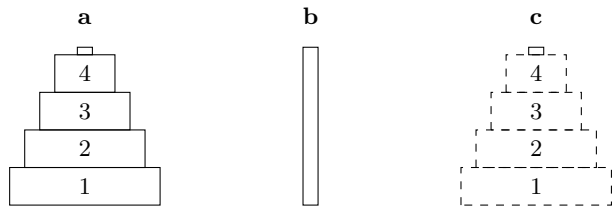
\begin{figure}[ht]
\centering
	\begin{tikzpicture}
\draw (0.5,0)   rectangle node {1} (2.5,0.5);
		\draw (0.7,0.5) rectangle node {2} (2.3,1);
		\draw (0.9,1)   rectangle node {3} (2.1,1.5);
		\draw (1.1,1.5) rectangle node {4} (1.9,2);
		\draw (1.4,2)   rectangle          (1.6,2.1);
		\draw [-] (1.5,2.3) node [above] {\textbf{a}};
\draw (4.4,0.0)   rectangle          (4.6,2.1);
		\draw [-] (4.5,2.3) node [above] {\textbf{b}};
\draw [dashed] (6.5,0)   rectangle node {1} (8.5,0.5);
		\draw [dashed] (6.7,0.5) rectangle node {2} (8.3,1);
		\draw [dashed] (6.9,1)   rectangle node {3} (8.1,1.5);
		\draw [dashed] (7.1,1.5) rectangle node {4} (7.9,2);
		\draw (7.4,2.0) rectangle          (7.6,2.1);
		\draw [-] (7.5,2.3) node [above] {\textbf{c}};
	\end{tikzpicture}
\caption{Towers of Hanoi: initial and goal situation}
\label{fig:hanoi}
\end{figure}

In this section and the following one,
we illustrate \clingo's multi-shot solving machinery
by applying it to a Towers of Hanoi puzzle~\cite{clingo-opt}.
Our example consists of three pegs and four disks of different size; it is shown in Figure~\ref{fig:hanoi}.
The goal is to move all disks from the left peg to the right one.
Only the topmost disk of a peg can be moved at a time.
 Furthermore,
a disk cannot be moved to a peg already containing a disk of smaller size.
Although there is an efficient algorithm to solve our simple puzzle,
we do not exploit it here and merely specify conditions for
sequences of moves being solutions.
The Towers of Hanoi puzzle constitutes a typical planning problem,
aiming at finding a plan, that is, a sequence of actions, that leads from an initial state to a state satisfying a goal.

\begin{table}
\begin{lstlisting}[label={fig:toh:opt:enc},caption={Bounded towers of hanoi encoding (\lstinline{tohB.lp})},language=clingos]
on(D,P,0) :- init_on(D,P).%%#(\label{fig:toh:opt:enc:init}#)

ngoal(T) :- goal_on(D,P), T=0..n, not on(D,P,T).%%#(\label{fig:toh:opt:enc:goal:begin}#)
:- ngoal(n).%%#(\label{fig:toh:opt:enc:goal:end}#)

1 { move(D,P,T): disk(D), peg(P) } 1 :- ngoal(T-1), T<=n.%%#(\label{fig:toh:opt:enc:trans:begin} \label{fig:toh:opt:enc:move}#)

move(D,T)        :- move(D,P,T).
on(D,P,T)        :- move(D,P,T).
on(D,P,T)        :- on(D,P,T-1), not move(D,T), T<=n.
blocked(D-1,P,T) :- on(D,P,T-1).
blocked(D-1,P,T) :- blocked(D,P,T), disk(D).

:- move(D,P,T), blocked(D-1,P,T).
:- move(D,T), on(D,P,T-1), blocked(D,P,T).
:- disk(D), not 1 { on(D,P,T) } 1, T=1..n.%%#(\label{fig:toh:opt:enc:trans:end}#)

#show move/3.

_minimize(1,T) :- ngoal(T).%%#(\label{fig:toh:opt:enc:minimize}#)
\end{lstlisting}
\begin{lstlisting}[label={fig:toh:opt:ins},language=clingos,caption={Towers of hanoi instance (\lstinline{tohI.lp})}]
peg(a;b;c).
disk(1..4).
init_on(1..4,a).
goal_on(1..4,c).

\end{lstlisting}
\end{table}

To illustrate how multi-shot solving can be used for realizing branch-and-bound-based optimization,
we consider the problem of finding the shortest plan solving our puzzle within a given horizon.
To this end,
we adapt the Towers of Hanoi encoding by \citeANP{gekakasc17a}~\citeNN{gekakasc17a} in Listing~\ref{fig:toh:opt:enc}.
Here, the length of the horizon is given by parameter \lstinline{n}.
The problem instance in Listing~\ref{fig:toh:opt:ins} together with
Line~\ref{fig:toh:opt:enc:init} in Listing~\ref{fig:toh:opt:enc}
gives the initial configuration of disks in Figure~\ref{fig:hanoi}.
Similarly,
the goal is checked in Lines~\ref{fig:toh:opt:enc:goal:begin}--\ref{fig:toh:opt:enc:goal:end} of Listing~\ref{fig:toh:opt:enc}
(by drawing on the problem instance in Listing~\ref{fig:toh:opt:ins}).
Because the overall objective is to solve the problem in the minimum number of steps within a given bound,
the goal is tested at each time step in Line~\ref{fig:toh:opt:enc:goal:begin}.
Once it is established, we do not permit any further moves and the goal persists in the following steps.
This allows us to read off whether the goal was reached at the planning horizon (in Line~\ref{fig:toh:opt:enc:goal:end}).
The state transition function along with state constraints are described in Lines~\ref{fig:toh:opt:enc:trans:begin}--\ref{fig:toh:opt:enc:trans:end}.
Since the encoding of the Towers of Hanoi problem is fairly standard,
we focus in the sequel on implementing branch-and-bound-based minimization.
In view of this, note that Line~\ref{fig:toh:opt:enc:move} ensures that moves are only permitted if the goal is not yet achieved in the previous state.
Thus, the subsequent states do not change anymore, which allows us to express the optimization function in
Line~\ref{fig:toh:opt:enc:minimize} as minimizing the number of steps in which the goal is not reached.

\begin{lstlisting}[float=tb,label={prg:opt:main},caption={Branch-and-bound optimization (\lstinline{opt.py})},language=pythons]
import sys
from clingo.symbol import Number, SymbolType
from clingo.application import Application, clingo_main

class OptExampleApp(Application):
    program_name = "opt-example"
    version = "1.0"

    def __init__(self):
        self._bound = None

    def _on_model(self, model):#(\label{prg:opt:main:models:begin}#)
        self._bound = 0
        for atom in model.symbols(atoms=True):#(\label{prg:opt:main:atoms:begin}#)
            if (atom.match("_minimize", 2)
            and atom.arguments[0].type
            is  SymbolType.Number):#(\label{prg:opt:main:minimize:end}#)
                self._bound += atom.arguments[0].number#(\label{prg:opt:main:atoms:end}\label{prg:opt:main:models:end}#)

    def main(self, ctl, files):#(\label{prg:opt:main:begin}#)
        if not files:#(\label{prg:opt:main:read:begin}#)
            files = ["-"]
        for f in files:
            ctl.load(f)#(\label{prg:opt:main:read:end}#)
        ctl.add("bound", ["b"],#(\label{prg:opt:main:add}#)
            ":- #sum { V,I: _minimize(V,I) } >= b.")#(\label{prg:opt:main:sum}#)

        ctl.ground([("base", [])])#(\label{prg:opt:main:ground}#)
        while ctl.solve(on_model=self._on_model).satisfiable:#(\label{prg:opt:loop:begin}\label{prg:opt:main:solve}#)
            print("Found new bound: {}".format(self._bound))#(\label{prg:opt:main:print_bound}#)
            ctl.ground([("bound", [Number(self._bound)])])#(\label{prg:opt:main:ground_bound}\label{prg:opt:loop:end}#)

        if self._bound is not None:#(\label{prg:opt:main:finish:begin}#)
            print("Optimum found")#(\label{prg:opt:main:finish:end}#)

clingo_main(OptExampleApp(), sys.argv[1:])
\end{lstlisting}

The idea of branch-and-bound-based optimization is to compute an optimal solution by
producing a series of increasingly better solutions until no better solution is found.
The solution obtained last is then an optimal one.
Listing~\ref{prg:opt:main} implements the corresponding optimization algorithm
via \clingo's \lstinline{Application} class from Section~\ref{sec:app}.
It starts by overriding \clingo's \lstinline{main} function in Line~\ref{prg:opt:main:begin} and
begins with reading the input either
from \lstinline{files} provided on the command line or from standard input
in Lines~\ref{prg:opt:main:read:begin} and~\ref{prg:opt:main:read:end}.

The basic building block of our algorithm consists of a weight constraint embedded in the following subprogram.
\begin{lstlisting}[language=clingos]
#program bound(b).
:- #sum { V,I: _minimize(V,I) } >= b.%%#(\label{prg:opt:sum}#)
\end{lstlisting}
This program ensures that the next stable model yields a better bound than the one of the solution at hand.
More precisely,
it expects a bound \lstinline{b} as parameter and adds the integrity constraint in Line~\ref{prg:opt:sum}
to enforce a better solution.
A new instance of this program is added for each consecutive stable model.
The addition of the (non-ground) constraint in Line~\ref{prg:opt:sum} as part of program \lstinline{bound(b)}
is accomplished in Lines~\ref{prg:opt:main:add} and~\ref{prg:opt:main:sum}.

The actual minimization algorithm starts by grounding the \lstinline{base} program in Line~\ref{prg:opt:main:ground}
before it enters the loop in Lines~\ref{prg:opt:loop:begin}--\ref{prg:opt:loop:end}.
This loop implements the branch-and-bound-based search for the minimum
by searching for stable models while updating the bound until the problem is unsatisfiable.
Note that we pass a callback to the \lstinline{solve} function in Line~\ref{prg:opt:main:solve}.
With it,
the \lstinline{_on_model} function in Lines~\ref{prg:opt:main:models:begin}--\ref{prg:opt:main:models:end} is called for every model found.
If there is a stable model, Lines~\ref{prg:opt:main:atoms:begin}--\ref{prg:opt:main:atoms:end} iterate over the atoms of the stable model
while summing up the current bound by extracting the weight of atoms over predicates \lstinline[mathescape=true]{_minimize/$2$}.
We check that the first argument of the atom is an integer and ignore atoms where this is not the case;
just like the \lstinline{#sum} aggregate in Line~\ref{prg:opt:main:sum}.
When a model was found, the body of the loop in Lines~\ref{prg:opt:loop:begin}--\ref{prg:opt:loop:end} is processed.
First, the algorithm prints the bound in Line~\ref{prg:opt:main:print_bound}.
Then, it adds an integrity constraint in Line~\ref{prg:opt:main:ground_bound}
making sure that the next stable model is strictly better than the current one.
Finally,
if the program becomes unsatisfiable, the branch-and-bound loop in Lines~\ref{prg:opt:loop:begin}--\ref{prg:opt:loop:end} ends
and Lines~\ref{prg:opt:main:finish:begin}--\ref{prg:opt:main:finish:end}
print that the previously found stable model (if any) is the optimal solution.

When running the augmented logic program in Listings~\ref{fig:toh:opt:enc}, \ref{fig:toh:opt:ins}, and~\ref{prg:opt:main} with a horizon of $17$,
the solver finds plans of length 17, 16, and 15 and shows that no plan of length 14 exists.
This is reflected by \clingo's output indicating four solver calls and three stable models:
\begin{lstlisting}[language=shell]
UNIX> python opt.py tohB.lp tohI.lp -c n=17
opt-example version 1.0
Reading from tohB.lp ...
Solving...
Answer: 1
move(4,b,1)  move(3,c,2)  move(4,a,3)  move(4,c,4)  move(2,b,5)  \
move(4,a,6)  move(3,b,7)  move(4,c,8)  move(4,b,9)  move(1,c,10) \
move(4,c,11) move(3,a,12) move(4,a,13) move(2,c,14) move(4,b,15)
move(3,c,16) move(4,c,17)
Found new bound: 17
Solving...
Answer: 1
move(4,b,1)  move(3,c,2)  move(4,c,3)  move(2,b,4)  move(4,a,5)  \
move(3,b,6)  move(4,c,7)  move(4,b,8)  move(1,c,9)  move(4,c,10) \
move(3,a,11) move(4,a,12) move(2,c,13) move(4,b,14) move(3,c,15)
move(4,c,16)
Found new bound: 16
Solving...
Answer: 1
move(4,b,1)  move(3,c,2)  move(4,c,3)  move(2,b,4)  move(4,a,5)  \
move(3,b,6)  move(4,b,7)  move(1,c,8)  move(4,c,9)  move(3,a,10) \
move(4,a,11) move(2,c,12) move(4,b,13) move(3,c,14) move(4,c,15)
Found new bound: 15
Solving...
Optimum found
UNSATISFIABLE
\end{lstlisting}
Note that at the end the solver deals with the ground program obtained from Listings~\ref{fig:toh:opt:enc}
and~\ref{fig:toh:opt:ins} grounded with parameter \lstinline{n=17}
along with the three integrity constraints
obtained from subprograms \lstinline{bound(17)}, \lstinline{bound(16)}, and \lstinline{bound(15)}.

Last but not least,
note that the functionality implemented above is equivalent to using \clingo's inbuilt optimization mode
by replacing Line~\ref{fig:toh:opt:enc:minimize} in Listing~\ref{fig:toh:opt:enc} with
\begin{lstlisting}[language=clingos,firstnumber=20]
#minimize { 1,T: ngoal(T) }.
\end{lstlisting}

\subsection{Incremental ASP solving}
\label{sec:iclingo}

Incremental ASP solving offers a step-oriented approach to ASP that avoids redundancies by
gradually processing the extensions to a problem
rather than repeatedly re-processing the entire growing problem.
To this end, a program is partitioned into a
base part, describing static knowledge independent of the step parameter~\lstinline{t},
a cumulative part, capturing knowledge accumulating with increasing~\lstinline{t},
and
a volatile part specific for each value of~\lstinline{t}.
In \clingo, all three parts are captured by \lstinline{#program} declarations
along with \lstinline{#external} atoms for handling volatile rules, namely,
subprograms named \lstinline{base}, \lstinline{step}, and \lstinline{check}
along with external atoms of form \lstinline{query(t)}.
Note that these names have no general, predefined meaning;
their composition is usually defined in an associated script.

We illustrate this by adapting the Towers of Hanoi encoding from Listing~\ref{fig:toh:opt:enc}
to an incremental version in Listing~\ref{fig:toh:enc}.
\begin{lstlisting}[float=tb,caption={Towers of hanoi incremental encoding (\lstinline{tohE.lp})},label={fig:toh:enc},language=clingos]
#program base.
on(D,P,0) :- init_on(D,P).%%#(\label{fig:toh:enc:static}#)

#program check(t).
:- goal_on(D,P), not on(D,P,t), query(t).%%#(\label{fig:toh:enc:goal}#)

#program step(t).%%#(\label{fig:toh:enc:step:begin}#)
1 { move(D,P,t): disk(D), peg(P) } 1.

move(D,t)        :- move(D,P,t).
on(D,P,t)        :- move(D,P,t).
on(D,P,t)        :- on(D,P,t-1), not move(D,t).
blocked(D-1,P,t) :- on(D,P,t-1).
blocked(D-1,P,t) :- blocked(D,P,t), disk(D).

:- move(D,P,t), blocked(D-1,P,t).
:- move(D,t), on(D,P,t-1), blocked(D,P,t).
:- disk(D), not 1 { on(D,P,t) } 1.%%#(\label{fig:toh:enc:step:end}#)

#show move/3.
\end{lstlisting}
To this end,
we arrange the original encoding in program parts \lstinline{base}, \lstinline{check(t)}, and \lstinline{step(t)},
use \lstinline{t} instead of \lstinline{T} as time parameter, and
simplify testing the goal.
Checking the goal is easier here because the incremental approach guarantees a shortest plan and,
hence, does not require additional minimization.

At first, we observe that
the problem instance in Listing~\ref{fig:toh:opt:ins} as well as Line~\ref{fig:toh:enc:static} in Listing~\ref{fig:toh:enc}
constitute static knowledge and thus belong to the \lstinline{base} program.
More interestingly, the query is expressed in Line~\ref{fig:toh:enc:goal} of Listing~\ref{fig:toh:enc}.
Its volatility is realized by making it subject to the truth assignment to the external atom \lstinline{query(t)}.
For convenience,
this atom is predefined in Line~\ref{fig:iclingo:python:query} in Listing~\ref{fig:iclingo:python} as part of the \lstinline{check} program.
Hence, for illustration, subprogram \lstinline{check} consists of a user- and a pre-defined part.
Finally,
the transition function along with state constraints are described in the subprogram \lstinline{step} in Lines~\ref{fig:toh:enc:step:begin}--\ref{fig:toh:enc:step:end}.

The idea is now to control the successive grounding and solving of the program parts in Listings~\ref{fig:toh:opt:ins} and~\ref{fig:toh:enc}
by the \python\ script in Listing~\ref{fig:iclingo:python}.\footnote{\label{fn:no:parsing}For brevity, we have stripped class \lstinline{IncConfig} in Line~\ref{fig:iclingo:python:parsing}
  from the parsing methods \lstinline{parse_int} and \lstinline{parse_stop}.
  The full source code is available online~\cite{clingo-opt}.}
\begin{lstlisting}[float=p,caption={\python\ script implementing incremental ASP solving in \clingo\ (\lstinline{inc.py})},label=fig:iclingo:python,language=pythons,linerange={1-9,29-81},basicstyle=\ttfamily\footnotesize]
import sys
from clingo.symbol import Function, Number
from clingo.application import Application, clingo_main

class IncConfig:
    def __init__(self):
        self.imin, self.imax, self.istop = 1, None, "SAT"

#(\textit{[...]}#)#(\label{fig:iclingo:python:parsing}#)

class IncExampleApp(Application):
    program_name = "inc-example"
    version = "1.0"

    def __init__(self):
        self._conf = IncConfig()

    def register_options(self, options):
        group = "Inc-Example Options"
        options.add(#(\label{fig:iclingo:python:opt:imin:begin}#)
            group, "imin",
            f"Minimum number of steps [{self._conf.imin}]",
            parse_int(self._conf, "imin", min_val=0),
            argument="<n>")#(\label{fig:iclingo:python:opt:imin:end}#)
        options.add(
            group, "imax",
            f"Maximum number of steps [{self._conf.imax}]",
            parse_int(self._conf, "imax", min_val=0, optional=True),
            argument="<n>")
        options.add(
            group, "istop",
            f"Stop criterion [{self._conf.istop}]",
            parse_stop(self._conf, "istop"))

    def main(self, ctl, files):
        if not files: files = ["-"]
        for f in files: ctl.load(f)
        ctl.add("check", ["t"], "#external query(t).")#(\label{fig:iclingo:python:check}\label{fig:iclingo:python:query}#)

        conf = self._conf
        imin, imax, istop = conf.imin, conf.imax, conf.istop#(\label{fig:iclingo:python:const}#)

        step, ret = 0, None#(\label{fig:iclingo:python:vars}#)
        while ((imax is None or step < imax) and#(\label{fig:iclingo:python:loop:begin}#)
               (step == 0 or step < imin or (
                  (istop == "SAT"     and not ret.satisfiable) or
                  (istop == "UNSAT"   and not ret.unsatisfiable) or
                  (istop == "UNKNOWN" and not ret.unknown)))):
            parts = []#(\label{fig:iclingo:python:parts}#)
            parts.append(("check", [Number(step)]))
            if step > 0:
                query = Function("query", [Number(step - 1)])
                ctl.release_external(query)
                parts.append(("step", [Number(step)]))
            else:
                parts.append(("base", []))
            ctl.ground(parts)
            query = Function("query", [Number(step)])
            ctl.assign_external(query, True)#(\label{fig:iclingo:python:assign:query}#)
            ret, step = ctl.solve(), step + 1#(\label{fig:iclingo:python:loop:end}\label{fig:iclingo:python:solve}#)

clingo_main(IncExampleApp(), sys.argv[1:])
\end{lstlisting}
To this end,
we use five variables to govern the loop in Lines~\ref{fig:iclingo:python:loop:begin}--\ref{fig:iclingo:python:loop:end}.\footnote{This follows the original implementation of incremental ASP solving in \iclingo~\cite{gekakaosscth08a}}
Variables \lstinline{imin} and \lstinline{imax} prescribe a minimum and maximum number of iterations, respectively;
\lstinline{istop} gives a termination criterion, e.g., \lstinline{"SAT"} or \lstinline{"UNSAT"}.
The value of \lstinline{step} is used to instantiate the parametrized subprograms
and \lstinline{ret} gives the solving result.
While
the initial values of \lstinline{step} and \lstinline{ret} are set in Line~\ref{fig:iclingo:python:vars},
the first three variables are user-defined.
We show how such user-defined variables are integrated into \clingo's option handling below,
but first describe the actual implementation of incremental ASP solving.

The subprograms grounded in each iteration are accumulated in the list \lstinline{parts}
(cf.\ Line~\ref{fig:iclingo:python:parts}).
Each of its entries is a pair consisting of a subprogram name along with its list of actual parameters.
In the very first iteration, the subprograms \lstinline{base} and \lstinline{check(0)} are grounded.
Note that this involves the declaration of the external atom \lstinline{query(0)} and the assignment of its default value false.
The latter is changed in Line~\ref{fig:iclingo:python:assign:query} to true in order to activate the actual query.
The \lstinline{solve} call in Line~\ref{fig:iclingo:python:solve} then amounts to checking whether the goal situation is already satisfied in the initial state.
As well, the value of \lstinline{step} is incremented to \lstinline{1}.

As long as the termination condition remains unfulfilled,
each following iteration takes the respective value of variable \lstinline{step}
to replace the parameter in subprograms \lstinline{step} and \lstinline{check}
during grounding.
In addition,
the current external atom \lstinline{query(t)} is set to true,
while the previous one is permanently set to false.
This disables the corresponding instance of the integrity constraint in Line~\ref{fig:toh:enc:goal} of Listing~\ref{fig:toh:enc} before it is replaced in the next iteration.
In this way,
the query condition only applies to the current horizon.

In our example, the \lstinline{solver} is called 16 times before a plan of length 15 is found:
\begin{lstlisting}[language=shell]
UNIX> python inc.py tohE.lp tohI.lp
inc-example version 1.0
Reading from tohE.lp ...
Solving...
|\textit{[\dots]}|
Solving...
Answer: 1
move(4,b,1)  move(3,c,2)  move(4,c,3)  move(2,b,4)  \
move(4,a,5)  move(3,b,6)  move(4,b,7)  move(1,c,8)  \
move(4,c,9)  move(3,a,10) move(4,a,11) move(2,c,12) \
move(4,b,13) move(3,c,14) move(4,c,15)
SATISFIABLE

Models       : 1+
Calls        : 16
\end{lstlisting}

Last but not least,
let us explain how option processing can be added by sketching how the three variables
\lstinline{imin}, \lstinline{imax}, and \lstinline{istop} can be set from the command line.
For this purpose, \clingo's API offers the two methods \lstinline{register_options} and \lstinline{validate_options}.
In our simple example,
we only use the former since the latter is meant to handle situations with conflicting options.\footnote{We also omit describing the underlying option parsers; cf.\ Footnote~\ref{fn:no:parsing}.}
In fact, \lstinline{register_options} receives an \lstinline{ApplicationOptions} object as parameter to register
additional options.
For example, option \lstinline{--imin} is added in
Lines~\ref{fig:iclingo:python:opt:imin:begin}--\ref{fig:iclingo:python:opt:imin:end}
by calling \lstinline{ApplicationOptions.add}.
Its first parameter is the name of an option group,
used as the section heading when printing the application's help when called with \lstinline{--help}.
In our example, all options are grouped in the ``\lstinline{Inc-Example Options}'' section.
More concretely, we get the following help output (omitting descriptions for \clingo's options):
\begin{lstlisting}[language=shell]
UNIX> python inc.py --help
|\textit{[...]}|
Inc-Example Options:

  --imin=<n>              : Minimum number of steps [1]
  --imax=<n>              : Maximum number of steps [None]
  --istop=<arg>           : Stop criterion [SAT]

|\textit{[...]}|
\end{lstlisting}
The second parameter is the name of the option on the command line.
Here, we pass \lstinline{"imin"} to add option \lstinline{--imin}.
This is followed by the option description and the (omitted) option parser.
Registration and parsing of options happens once before the \lstinline{IncExampleApp.main} method is called.
Hence, at that time the attributes of the applications \lstinline{IncExampleApp._conf} object
either have their default values or were overwritten by options passed on the command line.
The last keyword parameter configures the placeholder, which is printed in the help output.
Since we are adding a numeric option, it is good practice to call it \lstinline{<n>}.
By default it is called \lstinline{<arg>}.

 \section{Theory-enhanced ASP solving}
\label{sec:theory}

This section provides fundamental concepts for extending \clingo\ with foreign types of constraints,
also referred to as theories.
To begin with, it is important to bear in mind that ASP is a modeling-grounding-solving paradigm,
in contrast to, for instance, the solving-centered approach of SAT Modulo Theories (SMT;~\citeNP{niolti06a}).
Hence extensions of ASP are rarely limited to single components
but rather spread throughout the whole workflow.
This begins with the addition of new language constructs to the input language,
requiring in turn enhancements to the grounder as well as
syntactic means for passing the ground constructs to a downstream system.
In case the latter is an ASP solver,
it must be enabled to handle the specific input and incorporate corresponding solving capacities.
Finally,
each such extension is rather specific and thus requires different means at all ends.

We begin by showing how \clingo's input language can be customized with theory-specific constructs.
We then outline \clingo's algorithmic approach to ASP solving with theory propagation
to put the description of \clingo's theory reasoning interface on firm grounds.

\subsection{Input language}
\label{sec:language}

We begin by introducing the theory-related features of \clingo's input language.
They are situated in the underlying grounder \gringo\ and can thus
also be used independently of \clingo.
We start with a detailed description of the generic means for defining theories
and complement this in \ref{sec:aspif} with an overview of the corresponding intermediate language \aspif.

The generic approach to theory specification rests upon two languages:
the one defining theory languages and the theory language itself.
Both borrow elements from the underlying ASP language,
foremost an aggregate-like syntax for formulating variable length expressions.
To illustrate this, consider Listing~\ref{prg:lc},
where a logic program is extended by constructs for handling difference and linear constraints.
While the former are binary constraints of the form
$x_1-x_2\leq k$, the latter have a variable size and are of the form
\(
a_1x_1+\dots+a_nx_n\circ k
\),
where $x_i$ are integer variables, $a_i$ and $k$ are integers, and $\circ\in\{\leq,\geq,<,>,=\}$
for $1\leq i\leq n$.
Note that solving difference constraints is polynomial, while solving linear equations (over integers) is NP-complete.
The theory language for expressing both types of constraints is defined in Lines~\ref{prg:lc:begin-theory}--\ref{prg:lc:end-theory} and preceded by the directive \lstinline{#theory}.
The elements of the resulting theory language are preceded by \lstinline{&} and used as regular atoms in the logic program in Lines~\ref{prg:lc:begin-program}--\ref{prg:lc:end-program}.
\begin{lstlisting}[language=clingos,morekeywords={n,m},float=t,label=prg:lc,caption=Logic program enhanced with difference and linear constraints (\lstinline{lc.lp})]
#theory lc { %%#( \label{prg:lc:begin-theory} #)
  constant    { - : 0, unary }; %%#( \label{prg:lc:begin-def-term} #)
  diff_term   { - : 0, binary, left };
  linear_term { + : 2, unary; - : 2, unary; %%#( \label{prg:lc:op-add} #)
                * : 1, binary, left; %%#( \label{prg:lc:op-mul} #)
                + : 0, binary, left;
                - : 0, binary, left };
  domain_term { .. : 1, binary, left }; %%#( \label{prg:lc:op-dot} #)
  show_term   { / : 1, binary, left }; %%#( \label{prg:lc:end-def-term} #)

  &dom/0 : domain_term, {=}, linear_term, any; %%#( \label{lst:ex1-begin-def-atom} #)
  &sum/0 : linear_term, {<=,=,>=,<,>,!=}, linear_term, any;
  &diff/0 : diff_term, {<=}, constant, any;
  &show/0 : show_term, directive %%#( \label{lst:ex1-end-def-atom} #)
}. %%#( \label{prg:lc:end-theory} #)

#const n=2.  #const m=1000.   %%#( \label{prg:lc:begin-program} #)

task(1..n).                   %%#( \label{prg:lc:rule-task} #)
duration(T,200*T) :- task(T). %%#( \label{prg:lc:rule-duration} #)

&dom  { 1..m } = start(T) :- task(T). %%#( \label{prg:lc:begin-use-theory} #)
&dom  { 1..m } = end(T)   :- task(T). 
&diff { end(T)-start(T) } <= D :- duration(T,D). %%#( \label{prg:lc:rule-diff} #)
&sum  { end(T): task(T); -start(T): task(T) } <= m.

&show { start/1; end/1 }. %%#( \label{prg:lc:end-use-theory} \label{prg:lc:end-program} #)
\end{lstlisting}

To be more precise,
a \emph{theory definition} has the form
\begin{lstlisting}[numbers=none,mathescape=t,language=clingo]
#theory $T$ {$D_1$;$\dots$;$D_n$}.
\end{lstlisting}
where $T$ is the theory name and each $D_i$ is a definition for a theory term or a theory atom for $1\leq i\leq n$.
The language induced by a theory definition is the set of all theory atoms constructible from its theory atom definitions.

A \emph{theory atom definition} has the form
\begin{lstlisting}[numbers=none,mathescape=t]
&$p$/$k$ : $t$,$o$ $\quad\textrm{ or }\quad$ &$p$/$k$ : $t$,{$\diamond_1$,$\dots$,$\diamond_m$},$t'$,$o$
\end{lstlisting}
where $p$ is a predicate symbol and $k$ its arity,
$t$ and $t'$ are names of theory term definitions,
each $\diamond_i$ is a theory operator for $m\geq 1$,
and
\(
o\in\{\text{\lstinline{head}},\text{\lstinline{body}},\text{\lstinline{any}}, \text{\lstinline{directive}}\}
\)
determines where theory atoms may occur in a rule.
Examples of theory atom definitions are given in Lines~\ref{lst:ex1-begin-def-atom}--\ref{lst:ex1-end-def-atom} of Listing~\ref{prg:lc}.
The language of a theory atom definition as given above contains all \emph{theory atoms} of the form
\begin{lstlisting}[numbers=none,mathescape=t]
&$a$ {$C_1$:$L_1$;$\dots$;$C_n$:$L_n$}$\quad\textrm{ or }\quad$ &$a$ {$C_1$:$L_1$;$\dots$;$C_n$:$L_n$} $\diamond$ $c$
\end{lstlisting}
where $a$ is an atom over predicate $p$ of arity $k$,
each $C_i$ is a tuple of theory terms in the language for~$t$,
$c$ is a theory term in the language for~$t'$,
$\diamond$ is a theory operator among $\{ \diamond_1, \dots, \diamond_m \}$,
and each $L_i$ is a regular condition (i.e., a tuple of regular literals)
for $1\leq i\leq n$.
Whether the last part `${}\diamond c$' is included depends on the form of a theory atom definition.
Further, observe that theory atoms with occurrence type \lstinline{any} can be used both in the head and body of a rule;
with occurrence types \lstinline{head} and \lstinline{body}, their usage can be restricted to rule heads and bodies only.
Occurrence type \lstinline{directive} is similar to type \lstinline{head} but additionally requires that the rule body must be completely evaluated during grounding.
Five occurrences of theory atoms can be found in Lines~\ref{prg:lc:begin-use-theory}--\ref{prg:lc:end-use-theory} of Listing~\ref{prg:lc}.

\begin{remark}
Having conditions, such as $L_i$ above, is useful to address variable length constraints as in Listing~\ref{prg:lc}.
However, atoms like \lstinline{task(T)} are given as facts and therefore are not subject to solving.
In general, however, conditions may involve atoms that are not decided during grounding.
If this is the case, they have to be handled with care since there is no predefined behavior.
For instance, a rule \lstinline|a :- &sum{ x : a } >= 0| may be unsatisfiable under certain semantic
principles~\cite{gelzha19a}.
Formal foundations and implementation techniques of conditional theory atoms were developed
by~\citeANP{cafascwa20a}~\citeNN{cafascwa20a} and~\citeANP{cafascwa20b}~\citeNN{cafascwa20b}.
\end{remark}

A \emph{theory term definition} has the form
\begin{lstlisting}[numbers=none,mathescape=t]
$t$ {$D_1$;$\dots$;$D_n$}
\end{lstlisting}
where $t$ is a name for the defined terms and each $D_i$ is a theory operator definition
for $1\leq i\leq n$.
A corresponding definition specifies the language of all theory terms
that can be constructed via its operators.
Examples of theory term definitions are given in Lines~\ref{prg:lc:begin-def-term}--\ref{prg:lc:end-def-term} of Listing~\ref{prg:lc}.
Each resulting \emph{theory term} is one of the following:
\par\medskip\noindent
\begin{minipage}{0.5\linewidth}
  \begin{itemize}
  \item a constant term: \ $c$
  \item a variable term: \ $v$
  \item a binary theory term: \  $t_1 \diamond t_2$
  \item a unary theory term: \  ${}\diamond t_1$
  \end{itemize}
\end{minipage}
\begin{minipage}{0.5\linewidth}
\begin{itemize}
  \item a function theory term: \  $f(t_1,\dots,t_k)$
  \item a tuple theory term: \  $(t_1,\dots,t_l,)$
  \item a set theory term: \  $\{t_1,\dots,t_l\}$
  \item a list theory term: \  $[t_1,\dots,t_l]$
\end{itemize}
\end{minipage}
\par\medskip\noindent
where each $t_i$ is a theory term,
$\diamond$ is a theory operator defined by some $D_i$,
$c$ and $f$ are symbolic constants,
$v$ is a first-order variable,
$k\geq 1$, and
$l\geq 0$.
(The trailing comma in tuple theory terms is optional if $l \neq 1$.)
Parentheses can be used to specify operator precedence.

A \emph{theory operator definition} has the form
\begin{lstlisting}[numbers=none,mathescape=t]
$\diamond$ : $p$,unary $\quad\textrm{ or }\quad$ $\diamond$ : $p$,binary,$a$
\end{lstlisting}
where $\diamond$ is a unary or binary theory operator with precedence $p\geq 0$
(determining implicit parentheses). Binary theory operators are additionally characterized by an associativity
$a\in\{\text{\lstinline{right}},\linebreak[1]\text{\lstinline{left}}\}$.
As an example,
consider Lines~\ref{prg:lc:op-add}--\ref{prg:lc:op-mul} of Listing~\ref{prg:lc},
where the \lstinline{left} associative \lstinline{binary} operators~\lstinline{+} and~\lstinline{*} are defined with
precedence~\lstinline{2} and~\lstinline{1}, respectively.
Hence, parentheses in terms like `\lstinline{(X+(2*Y))+Z}' can be omitted.
In total, Lines~\ref{prg:lc:begin-def-term}--\ref{prg:lc:end-def-term} of Listing~\ref{prg:lc}
include nine theory operator definitions.
Specific \emph{theory operators} can be assembled (written consecutively without spaces) from the symbols
`\lstinline{!}',
`\lstinline{<}',
`\lstinline{=}',
`\lstinline{>}',
`\lstinline{+}',
`\lstinline{-}',
`\lstinline{*}',
`\lstinline{/}',
`\lstinline{\}',
`\lstinline{?}',
`\lstinline{&}',
`\lstinline{|}',
`\lstinline{.}',
`\lstinline{:}',
`\lstinline{;}',
`\lstinline{~}', and
`\lstinline{^}'.
For instance,
in Line~\ref{prg:lc:op-dot} of Listing~\ref{prg:lc}, the operator `\lstinline{..}' is defined as the concatenation of two periods.
The tokens
`\lstinline{.}',
`\lstinline{:}',
`\lstinline{;}', and
`\lstinline{:-}'
must be combined with other symbols
due to their dedicated usage.
Instead, one may write
`\lstinline{..}',
`\lstinline{::}',
`\lstinline{;;}',
`\lstinline{::-}',
etc.

While theory terms are formed similar to regular ones,
theory atoms rely upon an aggregate-like construction for forming variable-length theory expressions.
In this way, standard grounding techniques can be used for gathering theory terms.
The treatment of theory terms still differs from their regular counterparts
in that the grounder skips simplifications like, e.g., arithmetic evaluation.
This can be seen on the different results in Listing~\ref{prg:grd:diff} of grounding terms formed with
the regular and theory-specific variants of operator `\lstinline{..}'.
Observe that
the fact \lstinline{task(1..n)} in Line~\ref{prg:lc:rule-task} of Listing~\ref{prg:lc} results in \lstinline{n} ground facts,
viz.\ \lstinline{task(1)} and \lstinline{task(2)} because of \lstinline{n=2}.
Unlike this, the theory expression \lstinline{1..m} stays structurally intact and is only transformed into \lstinline{1..1000} in view of \lstinline{m=1000}.
That is, the grounder does not evaluate the theory term \lstinline{1..1000} and
leaves its interpretation to a downstream theory solver.
\begin{lstlisting}[language=clingos,float=t,label=prg:grd:diff,caption={Human-readable result of grounding Listing~\ref{prg:lc} via `\lstinline{gringo --text lc.lp}'}]
task(1).%%#( \label{grd:diff:fact:one} #)
task(2).%%#( \label{grd:diff:fact:two} #)
duration(1,200).%%#( \label{grd:diff:fact:tri} #)
duration(2,400).%%#( \label{grd:diff:fact:for} #)

&dom { 1..1000 } = start(1).
&dom { 1..1000 } = start(2).
&dom { 1..1000 } = end(1).
&dom { 1..1000 } = end(2).

&diff { end(1)-start(1) } <= 200.%%#( \label{grd:diff:fact:fiv} #)
&diff { end(2)-start(2) } <= 400.%%#( \label{grd:diff:fact:six} #)

&sum { end(1); end(2); -start(1); -start(2) } <= 1000.

&show { start/1; end/1 }.%%#( \label{grd:diff:show} #)
\end{lstlisting}
A similar situation is encountered when comparing the treatment of the regular term `\lstinline{200*T}' in Line~\ref{prg:lc:rule-duration} of Listing~\ref{prg:lc} to the
theory term `\lstinline{end(T)-start(T)}' in Line~\ref{prg:lc:rule-diff}.
While each instance of `\lstinline{200*T}' is evaluated during grounding,
instances of the theory term `\lstinline{end(T)-start(T)}' are left intact in Lines~\ref{grd:diff:fact:fiv} and~\ref{grd:diff:fact:six} of Listing~\ref{prg:grd:diff}.
In fact, if `\lstinline{200*T}' had been a theory term as well,
it would have resulted in the unevaluated instances  `\lstinline{200*1}' and `\lstinline{200*2}'.

\subsection{Semantic principles}
\label{sec:semantics}

Given the hands-on nature of this work,
we only give an informal idea of the semantic principles underlying theory solving in ASP.

A logic program induces a set of stable models.
To extend this concept to logic programs with theory expressions,
we follow the approach of lazy theory solving~\cite{baseseti09a}.
We abstract from the specific semantics of a theory by considering the theory atoms representing the underlying theory constraints.
The idea is that a regular stable model of a program over regular and theory atoms is only valid with respect to a theory,
if the constraints induced by the truth assignment to the theory atoms are satisfiable in the theory.

In the example above,
this amounts to finding a numeric assignment to all theory variables satisfying all difference and linear constraints associated with theory atoms.
The ground program in Listing~\ref{prg:grd:diff} has a single stable model consisting of all regular and theory atoms in Lines~\ref{grd:diff:fact:one}--\ref{grd:diff:show}.
Here, we easily find assignments satisfying the induced constraints,
e.g.,\
\(
\text{\lstinline{start(1)}}\mapsto 1
\),
\(
\text{\lstinline{end(1)}}\mapsto 2
\),
\(
\text{\lstinline{start(2)}}\mapsto 2
\), and
\(
\text{\lstinline{end(2)}}\mapsto 3
\).

In fact,
there are alternative semantic options for capturing theory atoms, as discussed by~\citeANP{gekakaosscwa16a}~\citeNN{gekakaosscwa16a}.
First of all,
we may distinguish whether imposed constraints are only determined outside or additionally inside a logic program.
This leads to the distinction between \emph{defined} and \emph{external} theory atoms
(analogous to rule heads and input atoms defined by \lstinline{#external} directives).
While external theory atoms must only be satisfied by the respective theory,
defined ones must additionally be derivable through rules in the program.
A second distinction concerns the interplay of ASP with theories.
More precisely, it is about the logical correspondence between theory atoms and theory constraints.
This leads us to the distinction between \emph{strict} and \emph{non-strict} theory atoms.
The strict correspondence requires
a constraint to be satisfied
\textit{iff}
the associated theory atom is true.
A weaker since only implicative condition is imposed in the non-strict case.
Here, a constraint must hold
\textit{only if}
the associated theory atom is true.
In other words, only non-strict theory atoms assigned true impose requirements,
while constraints associated with falsified non-strict theory atoms are free to hold or not.
However, by contraposition, a violated constraint leads to a false non-strict theory atom.

\subsection{Algorithmic aspects}
\label{sec:algo}

The algorithmic approach to ASP solving modulo theories of \clingo,
or more precisely that of its underlying ASP solver \clasp,
follows the lazy approach to solving in Satisfiability Modulo Theories (SMT;~\citeNP{baseseti09a}).
We give below an abstract overview that serves as light algorithmic underpinning for the description of \clingo's implementation
given in the next section.

A ground program~$P$ induces \emph{completion} and \emph{loop nogoods}, called $\Delta_P$ or $\Lambda_P$, respectively,
that can be used for computing stable models of $P$~\cite{gekasc09c}.
Nogoods represent invalid partial assignments and can be thought of as negative Boolean constraints.
We represent (partial) assignments as consistent sets of literals.
An assignment is total if it contains either the positive or negative literal of each atom.
We say that a nogood is violated by an assignment if the former is contained in the latter;
a nogood is unit if all but one of its literals are in the assignment.
Each total assignment not violating any nogood in $\Delta_P\cup\Lambda_P$
yields a regular stable model of~$P$, and such an assignment is called a solution
(for $\Delta_P\cup\Lambda_P$).
To accommodate theories,
we identify a theory $T$ with a set $\Delta_T$ of \emph{theory nogoods},
and extend the concept of a solution in the straightforward way.

The nogoods in $\Delta_P\cup\Lambda_P\cup\Delta_T$ provide the
logical foundation for the Conflict-Driven Constraint Learning (CDCL) procedure~\cite{malyma09a,gekasc09c}
outlined in Figure~\ref{fig:cdcl}.
While the completion nogoods in~$\Delta_P$ are usually made explicit and subject to
unit propagation,\footnote{Unit propagation extends an assignment with literals complementary to the ones missing in unit nogoods.}
the loop nogoods in~$\Lambda_P$ as well as theory nogoods in~$\Delta_T$ are typically
handled by dedicated propagators and particular members are selectively recorded.
\begin{figure}[t]
  \newcounter{cddl}
  \setcounter{cddl}{8}
  \renewcommand{\thecddl}{\Alph{cddl}}
  \begin{tabbing}
   (X) \= Xx \= Xx \= Xx \= \kill \refstepcounter{cddl}(\thecddl)\label{fig:cdcl:init}
         \> \textit{initialize} \` // register theory propagators and initialize watches \\
         \> \textbf{loop} \\
         \> \> \textit{propagate} completion, loop, and recorded nogoods
               \` // deterministically assign literals \\
         \> \> \textbf{if} no conflict \textbf{then} \\
         \> \> \> \textbf{if} all variables assigned \textbf{then} \\
   \setcounter{cddl}{2}\refstepcounter{cddl}(\thecddl)\label{fig:cdcl:check}
         \> \> \> \> \textbf{if} some $\delta\in\Delta_T$ is violated \textbf{then} record $\delta$
                     \` // theory propagators check $\Delta_T$ \\
         \> \> \> \> \textbf{else} \textbf{return} variable assignment
                   \`// theory-based stable model found \\
         \> \> \> \textbf{else} \\
   \setcounter{cddl}{15}\refstepcounter{cddl}(\thecddl)\label{fig:cdcl:propagate}
         \> \> \> \> \textit{propagate} theories
                     \`// theory propagators may record theory nogoods from $\Delta_T$ \\
         \> \> \> \> \textbf{if} no nogood recorded \textbf{then} \textit{decide}
                     \`// non-deterministically assign some literal \\
         \> \> \textbf{else} \\
         \> \> \> \textbf{if} top-level conflict \textbf{then} \textbf{return} unsatisfiable \\
         \> \> \> \textbf{else} \\
         \> \> \> \> \textit{analyze} \`// resolve conflict and record a conflict constraint \\
   \setcounter{cddl}{20}\refstepcounter{cddl}(\thecddl)\label{fig:cdcl:undo}
         \> \> \> \> \textit{backjump} \`// undo assignments until conflict constraint is unit
 \end{tabbing}
\caption{Basic algorithm for Conflict-Driven Constraint Learning (CDCL) modulo theories}
\label{fig:cdcl}
\end{figure}While a dedicated propagator for loop nogoods is built-in in systems like \clingo,
those for theories are provided via the interface \textbf{Propagator} in Figure~\ref{fig:interface}.
To utilize custom propagators,
the algorithm in Figure~\ref{fig:cdcl} includes an \emph{initialization} step in Line~(\ref{fig:cdcl:init}).
In addition to the ``registration'' of a propagator for a theory
as an extension of the basic CDCL procedure,
common   tasks performed in this step include setting up internal data structures and
so-called watches for (a subset of) the theory atoms,
so that the propagator is invoked (only) when some watched literal gets assigned.

As usual,
the main CDCL loop starts with unit propagation on completion and loop nogoods,
the latter handled by the respective built-in propagator, as well as any nogoods
already recorded.
If this results in a non-total assignment without conflict,
theory propagators for which some of their watched literals have been assigned
are invoked in Line~(\ref{fig:cdcl:propagate}).
A propagator for a theory~$T$ can then inspect the current assignment,
update its data structures accordingly, and most importantly,
perform \emph{theory propagation} determining theory nogoods $\delta\in\Delta_T$ to record.
Usually, any such nogood~$\delta$ is unit or conflicting in order to trigger unit propagation or conflict resolution,
although this is not a necessary condition.
The interplay of unit and theory propagation continues until a conflict or
a total assignment arises,
or no (further) watched literals of theory propagators get assigned by unit propagation.
In the latter case, some non-deterministic decision is made to extend the partial
assignment at hand and then to proceed with unit and theory propagation.

If no conflict arises and an assignment is total,
in Line~(\ref{fig:cdcl:check}), theory propagators are called, one by one,
for a final \emph{check}.
The idea is that, e.g., a ``lazy'' propagator for a theory~$T$
that does not exhaustively test violations of its theory nogoods by partial assignments
can make sure that the assignment is indeed a solution for~$\Delta_T$, or record some
violated nogood(s) from $\Delta_T$ otherwise.
Even in case theory propagation on partial assignments is exhaustive and a final
check is not needed to detect conflicts,
the information that search led to a total assignment can be useful in practice, e.g.,
to store values for integer variables like
\lstinline{start(1)},
\lstinline{start(2)},
\lstinline{end(1)}, and
\lstinline{end(2)} in Listing~\ref{prg:grd:diff}
that witness the existence of a solution for $T$.

Finally, in case of a conflict, i.e., some completion or recorded nogood is violated by
the current assignment,
provided that some non-deterministic decision is involved in the conflict,
a new conflict constraint is recorded and utilized to guide backjumping
in Line~(\ref{fig:cdcl:undo}), as usual with CDCL.
In a similar fashion as the assignment of watched literals serves as
trigger for theory propagation, theory propagators are informed when they become
unassigned upon backjumping.
This allows the propagators to \emph{undo} earlier operations, e.g., internal data structures can be
reset to return   to a state taken prior to the assignment of watches.

In summary, the basic CDCL procedure is extended in four places to account for
custom propagators: initialization, propagation of (partial) assignments,
final check of total assignments, and undo steps upon backjumping.

\subsection{Propagator interface}
\label{sec:system}

\begin{figure}
  \includegraphics[width=\textwidth]{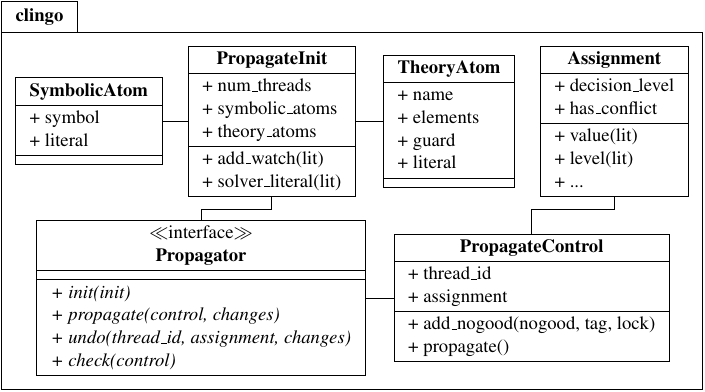}
  \caption{Class diagram of \clingo's (theory) propagator interface\label{fig:interface}}
\end{figure}

We now turn to the implementation of theory propagation in \clingo\
and detail the structure of its interface depicted in Figure~\ref{fig:interface}.
The interface \lstinline{Propagator} has to be implemented by each custom propagator.
After registering such a propagator with \clingo,
its functions are called during initialization and search as indicated in Figure~\ref{fig:cdcl}.
Function \lstinline{init}
is called once before solving (Line~(\ref{fig:cdcl:init}) in Figure~\ref{fig:cdcl})
to allow for initializing data structures used during theory propagation.
It is invoked with a \lstinline{PropagateInit} object providing access to symbolic (\lstinline{SymbolicAtom}) as well as theory (\lstinline{TheoryAtom}) atoms.
Both kinds of atoms are associated with program literals,
which are in turn associated with solver literals.
Program as well as solver literals are identified by non-zero integers, where positive and negative numbers represent positive or  negative literals, respectively.
In order to get notified about assignment changes, a propagator can set up watches on solver literals during initialization.

During search, function \lstinline{propagate} is called with a \lstinline{PropagateControl} object
and a (non-empty) list of watched literals that got assigned in the recent round of unit propagation (Line~(\ref{fig:cdcl:propagate}) in Figure~\ref{fig:cdcl}).
The \lstinline{PropagateControl} object can be used to inspect the current assignment, record nogoods, and trigger unit propagation.
Furthermore, to support multi-threaded solving,
its \lstinline{thread_id} property identifies the currently active thread,
each of which can be viewed as an independent instance of the CDCL algorithm in Figure~\ref{fig:cdcl}.\footnote{Depending on the configuration of \clasp, threads can communicate with each other.
For example, some of the recorded nogoods can be shared.
This is transparent from the perspective of theory propagators.}
Function \lstinline{undo} is the counterpart of \lstinline{propagate}
and called whenever the solver retracts assignments to watched literals (Line~(\ref{fig:cdcl:undo}) in Figure~\ref{fig:cdcl}).
In addition to the list of watched literals that have been retracted (in chronological order),
it receives the identifier and the assignment of the active thread.
Finally, function \lstinline{check} is similar to \lstinline{propagate},
yet invoked without a list of changes.
Instead, it is (only) called on total assignments
(Line~(\ref{fig:cdcl:check}) in Figure~\ref{fig:cdcl}), independently of watches.
Overriding the empty default implementations of propagator methods is optional.

\section{Extending ASP with difference constraints}
\label{sec:case}

In this section,
we develop a case-study featuring the extension of ASP with difference constraints\footnote{In SMT, the underlying formal system is also referred to as \emph{quantifier free integer difference logic}.}
by augmenting \clingo\ with a corresponding propagator.
To this end, we extend the language of Section~\ref{sec:background}
with difference constraint atoms of form
\begin{lstlisting}[mathescape,numbers=none]
&diff { $u$-$v$ } <= $d$
\end{lstlisting}
where $u$ and $v$ are (regular) terms and
$d$ is an integer constant.
Such atoms may either occur in the head or the body of a rule.
Hence,
stable models may now also include theory atoms of form
`\lstinline[mathescape]|&diff { $u$-$v$ } <= $d$|'.
More precisely,
for a stable model $X$, let $C_X$ be the set of \emph{difference constraints} $u-v \leq d$ associated with theory atoms
`\lstinline[mathescape]|&diff { $u$-$v$ } <= $d$|' in $X$
and $V_X$ be the set of all (integer) variables occurring in the difference constraints in $C_X$.
In our case, a stable model $X$ is then \emph{\textsc{dc}-stable},
if there is a mapping from $V_X$ to the integers, first,
satisfying all constraints in $C_X$,
and second, falsifying all constraints not in $C_X$ that are associated with a \emph{strict} difference
constraint atom~\cite{jakaosscscwa17a}.

Next, let us discuss the semantic principles guiding our implementation.
Recall from Section~\ref{sec:semantics} that theory atoms
may have different semantic properties,
either defined or external   depending upon their occurrence, or
either strict  or non-strict depending upon their logical relation to the represented constraint.
For difference constraints,
the combinations of \emph{strict} and \emph{external} as well as \emph{non-strict} and \emph{defined}
appear to be the most intuitive combinations~\cite{jakaosscscwa17a}.

To illustrate this, consider the following example:
\begin{lstlisting}[label={example:strict},language=clingos]
&diff { 0-x } <= -2.
a :- &diff { 0-x } <= -1.
\end{lstlisting}
This program states that $x$ is greater or equal 2 and that \lstinline{a} is derived if $x$ is greater or equal than 1.
An intuitive result is to assign $x$ a value greater or equal than 2 and to derive \lstinline{a}.
However, in case atom
`\lstinline|&diff { 0-x } <= -1|'
is non-strict,
we also obtain answer sets without \lstinline{a}.
This is because the falsity of
`\lstinline|&diff { 0-x } <= -1|'
does not imply that $0-x \leq -1$ is false as well.
This is enforced by interpreting the relation between
`\lstinline|&diff { 0-x } <= -1|' and $0-x \leq -1$
as strict, and then $a$ is obtained.
Intuitively,
the combination of external and strict can be seen as interpreting theory atoms relative to an external oracle,
according to which all possibilities have to be considered as the logic program is oblivious to the meaning of the theory atom.

For a complement to the above example, consider the following one:
\begin{lstlisting}[label={example:non_strict},language=clingos]
&diff { 0-x } <= -2.
&diff { 0-x } <= -1 :- a.
\end{lstlisting}
Again, the program states that $x$ is greater equal 2 but now atom \lstinline{a} derives that $x$ is greater or equal than 1.
As we do not have a definition of \lstinline{a},
we expect answer sets not containing \lstinline{a} and assignments where $x$ is greater or equal 2.
However, once we interpret the atom
`\lstinline|&diff { 0-x } <= -1|'
as strict, the program becomes unsatisfiable.
In this case the falsity of
`\lstinline|&diff { 0-x } <= -1|'
implies that $0-x \leq -1$ is false as well.
A non-strict interpretation avoids this and yields the expected result.
The combination of non-strict and defined lets the logic program decide which theory atoms hold.
Specifically,
the absence of an atom in an answer set does not imply that the constraint is false but rather that it is not enforced.
As a result,
we handle occurrences of difference constraint atoms in the head as defined and non-strict, and
atom occurrences in the body as external and strict.\footnote{Note that this amounts to treating occurrences of the same constraint atom in different ways.
This is not unusual since the same constraint may be represented by syntactically different constraint atoms.}

Let us now turn to the actual extension of \clingo.
The overall implementation is divided in two,
on the one hand,
the actual application class \lstinline{DLApp} addressing grounding and solving in Listing~\ref{prg:dl:theory},
and on the other hand,
six classes dealing with various aspects of difference constraints.
The complete source code is available online~\cite{clingo-dl}.

In what follows,
we concentrate on the \lstinline{HeadBodyTransformer} class in Listing~\ref{prg:dl:transformer},
illustrating the manipulation of a logic program's abstract syntax tree (AST),
as well as the \lstinline{DLPropagator} class in Listing~\ref{prg:dl:propagator},
showcasing a propagator adding foreign inferences to ASP.
To support this, we also describe the interface of the \lstinline{Graph} class but refrain from presenting its implementation.
\begin{lstlisting}[float=tp,label={prg:dl:language},caption={Theory language~\lstinline{dl} for difference constraints (\lstinline{dl.py}, Lines {\ref{prg:dl:theory:begin}--\ref{prg:dl:theory:end}})},linerange={10-23},firstnumber=10,language=pythons,literate={_}{\Underscore}1 {\%\%}{}0]
THEORY = """%%#(\label{prg:dl:theory:begin}#)
#theory dl{
    diff_term {
    -  : 3, unary;
    ** : 2, binary, right;
    *  : 1, binary, left;
    /  : 1, binary, left;
    \\ : 1, binary, left;
    +  : 0, binary, left;
    -  : 0, binary, left
    };
    &diff/1 : diff_term, {<=}, diff_term, any
}.
"""#(\label{prg:dl:theory:end}#)
\end{lstlisting}
For expressing difference constraints,
we define the theory language~\lstinline{dl} in Listing~\ref{prg:dl:language},
a subset of the theory language~\lstinline{lc} presented in Listing~\ref{prg:lc} above.
Note that to accommodate the additional argument indicating the location of the difference constraint atom,
we have replaced \lstinline{&diff/0} by \lstinline{&diff/1}.

To achieve the above distinction between head and body occurrences of theory atoms without changing the input language,
we use \clingo's functionalities to modify the AST of non-ground programs
for tagging theory atoms with their respective occurrence.
Although pragmatic, the annotation of theory atoms has turned out to be very useful in several implementations.
Moreover, it serves us as a first example of how non-ground programs can be modified through \clingo's API.

\begin{remark}
The user still writes difference constraints over \lstinline{&diff/0}.
The AST modification occurs on the non-ground level during parsing.
Once grounded, it is checked whether all theory atoms are valid with regards to a theory language.
At this point, difference constraints are constructed over \lstinline{&diff/1},
which is opaque to the user.
\end{remark}
As mentioned,
this is accomplished by the \lstinline{HeadBodyTransformer} class in Listing~\ref{prg:dl:transformer}.
\begin{lstlisting}[float=tp,label={prg:dl:transformer},caption={\lstinline{HeadBodyTransformer} class for tagging occurrence of theory atoms (\lstinline{dl.py}, Lines {\ref{prg:dl:transformer:begin}--\ref{prg:dl:transformer:end}})},linerange={79-91},firstnumber=79,language=pythons]
class HeadBodyTransformer(ast.Transformer):#(\label{prg:dl:transformer:begin}#)
    def visit_Literal(self, lit, in_lit=False):#(\label{prg:dl:transformer:visit_lit:begin}#)
        return lit.update(**self.visit_children(lit, True))#(\label{prg:dl:transformer:lit:visit}\label{prg:dl:transformer:visit_lit:end}#)

    def visit_TheoryAtom(self, atom, in_lit=False):#(\label{prg:dl:transformer:tatom:begin}#)
        term = atom.term
        if term.name == "diff" and not term.arguments:
            loc = "body" if in_lit else "head"
            atom = atom.update(term=ast.Function(
                term.location, term.name,
                [ast.Function(term.location, loc, [], False)],
                False))
        return atom#(\label{prg:dl:transformer:tatom:end}\label{prg:dl:transformer:end}#)
\end{lstlisting}
This class nicely illustrates how the \emph{visitor design pattern} is used by \clingo\ to manipulate the
AST of (non-ground) logic programs.
The transformer uses the property that theory atoms in rule heads or bodies are never or always children of literal nodes, respectively.
Assuming that the root node of the AST is visited with \lstinline{in_lit=False},
function \lstinline{visit_Literal} in Lines~\ref{prg:dl:transformer:visit_lit:begin}--\ref{prg:dl:transformer:visit_lit:end} visits its children with \lstinline{in_lit=False}.
Hence, function \lstinline{visit_TheoryAtom} in
Lines~\ref{prg:dl:transformer:tatom:begin}--\ref{prg:dl:transformer:tatom:end} is visited with the parameter \lstinline{in_lit} indicating a head or body occurrence and
returns the theory atom with the location (either \lstinline{head} or \lstinline{body}) as an argument.
For instance,
treating the example above using this class results in the following program:
\begin{lstlisting}[label={example:strict:parsed},language=clingos]
&diff(head) { 0-x } <= -2.
a :- &diff(body) { 0-x } <= -1.
\end{lstlisting}

\begin{lstlisting}[float=tp,label={prg:dl:theory},caption={Application class \lstinline{DLApp} with main loop for difference constraints (\lstinline{dl-app.py})},language=pythons]
import sys
from clingo.application import Application, clingo_main
from clingo.ast import ProgramBuilder, parse_files
from dl import DLPropagator, HeadBodyTransformer, THEORY

class DLApp(Application):
    program_name = "clingo-dl"
    version = "1.0"

    def __init__(self):
        self._propagator = DLPropagator()

    def on_model(self, model):#(\label{prg:dl:theory:on_model}#)
        self._propagator.on_model(model)

    def main(self, ctl, files):#(\label{prg:dl:theory:main:begin}#)
        ctl.register_propagator(self._propagator)
        ctl.add("base", [], THEORY)

        with ProgramBuilder(ctl) as bld:
            hbt = HeadBodyTransformer()
            parse_files(files, lambda stm: bld.add(hbt.visit(stm)))

        ctl.ground([("base", [])])
        ctl.solve(on_model=self.on_model)#(\label{prg:dl:theory:main:end}#)

sys.exit(int(clingo_main(DLApp(), sys.argv[1:])))
\end{lstlisting}
Listing~\ref{prg:dl:theory} shows the application that addresses grounding and solving.
Lines~\ref{prg:dl:theory:main:begin}--\ref{prg:dl:theory:main:end} implement a customized main function.
The difference to \clingo's regular one is that a propagator for difference constraints is registered,
the string variable \lstinline{THEORY} containing the above theory language is added as a program, and
the input programs are rewritten adding locations to the difference constraint atoms;
grounding and solving then follow as usual.
Note that the \lstinline{solve} function in Line~\ref{prg:dl:theory:main:end} takes a model callback as argument.
Whenever a \textsc{dc}-stable model $X$ is found,
this callback adds symbols to the answer set representing a mapping satisfying the corresponding difference constraints~$C_X$.
The model $X$ (excluding theory atoms) is printed as part of \clingo's default output.
The callback function \lstinline{on_model} in Line~\ref{prg:dl:theory:on_model} calls in turn the \lstinline{on_model}
function of the propagator (Line~\ref{prg:dl:theory:on_model} in Listing~\ref{prg:dl:propagator}) that
adds symbols of the form $\mathtt{dl}(x,v)$ to the model,
where $x$ is the name of an integer variable and $v$ the assigned value in $X$.

Our example propagator for difference constraints in Listing~\ref{prg:dl:propagator}
implements the algorithm presented by~\citeANP{cotmal06a}~\citeNN{cotmal06a}.
\begin{lstlisting}[float=tp,label={prg:dl:propagator},caption={\lstinline{DLPropagator} class for difference constraints (\lstinline{dl.py}, Lines {\ref{prg:dl:propagator:begin}--\ref{prg:dl:propagator:end}})},linerange={192-244},firstnumber=192,language=pythons]
class DLPropagator(Propagator):#(\label{prg:dl:propagator:begin}#)
    def __init__(self):
        self._l2e = {}    # {literal: [(node, node, weight)]}#(\label{prg:dl:propagator:member:l2e}#)
        self._e2l = {}    # {(node, node, weight): [literal]}#(\label{prg:dl:propagator:member:e2l}#)
        self._states = [] # [Graph]#(\label{prg:dl:propagator:member:state}#)

    def _state(self, thread_id):
        while len(self._states) <= thread_id:
            self._states.append(Graph())
        return self._states[thread_id]

    def _lit(self, control, edge):
        for lit in self._e2l[edge]:
            if control.assignment.is_true(lit):
                return lit
        assert False

    def _add_edge(self, init, lit, u, v, w):
        edge = (u, v, w)
        self._l2e.setdefault(lit, []).append(edge)#(\label{prg:dl:propagator:init:l2e}#)
        self._e2l.setdefault(edge, []).append(lit)#(\label{prg:dl:propagator:init:e2l}#)
        init.add_watch(lit)#(\label{prg:dl:propagator:init:watch}#)

    def init(self, init):#(\label{prg:dl:propagator:init:begin}#)
        for atom in init.theory_atoms:#(\label{prg:dl:propagator:init:loop:begin}#)
            term = atom.term
            if term.name == "diff" and len(term.arguments) == 1:
                u = _eval(atom.elements[0].terms[0].arguments[0])#(\label{prg:dl:propagator:init:edge:begin}#)
                v = _eval(atom.elements[0].terms[0].arguments[1])
                w = _eval(atom.guard[1]).number#(\label{prg:dl:propagator:init:edge:end}#)
                lit = init.solver_literal(atom.literal)#(\label{prg:dl:propagator:init:map-literal}#)
                self._add_edge(init, lit, u, v, w)
                if term.arguments[0].name == "body":#(\label{prg:dl:propagator:init:strict}#)
                    self._add_edge(init, -lit, v, u, -w - 1)#(\label{prg:dl:propagator:init:loop:end}#)#(\label{prg:dl:propagator:init:end}#)

    def propagate(self, control, changes):#(\label{prg:dl:propagator:propagate:begin}#)
        state = self._state(control.thread_id)#(\label{prg:dl:propagator:propagate:state}#)
        level = control.assignment.decision_level
        for lit in changes:#(\label{prg:dl:propagator:propagate:loop:begin}#)
            for edge in self._l2e[lit]:
                cycle = state.add_edge(level, edge)#(\label{prg:dl:propagator:propagate:add-edge}#)
                if cycle is not None:
                    c = [self._lit(control, e) for e in cycle]
                    control.add_nogood(c) and control.propagate()#(\label{prg:dl:propagator:propagate:add-nogood}#)
                    return#(\label{prg:dl:propagator:propagate:end}#)#(\label{prg:dl:propagator:propagate:loop:end}#)

    def undo(self, thread_id, assign, changes):#(\label{prg:dl:propagator:undo:begin}#)
        self._state(thread_id).backtrack(assign.decision_level)#(\label{prg:dl:propagator:undo:end}#)

    def on_model(self, model):#(\label{prg:dl:propagator:on_model:begin}#)
        assignment = self._state(model.thread_id).get_assignment()
        model.extend([Function("dl", [var, Number(value)])
                      for var, value in assignment])#(\label{prg:dl:propagator:on_model:end}\label{prg:dl:propagator:end}#)
\end{lstlisting}
The idea is that deciding whether a set of difference constraints is satisfiable can be mapped to a graph problem.
Given a set of difference constraints, let $(V,E)$ be the weighted directed graph
such that
$V$ is the set of variables occurring in the constraints and
$E$ the set of weighted edges $(u, v, d)$ for each constraint $u-v\leq d$.
The set of difference constraints is satisfiable if the corresponding graph does not contain a negative cycle
(i.e.\ a cycle whose sum of edge labels is negative).
\begin{figure}[htb]
\centering
\includegraphics[]{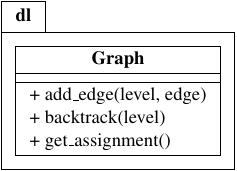}
\caption{Class diagram for the \lstinline{Graph} class\label{fig:class:graph}}
\end{figure}
The \lstinline{} class is in charge of cycle detection; its interface is given in Figure~\ref{fig:class:graph}.
We refrain from giving its code and rather concentrate on describing its interface:
\begin{itemize}
\item
  Function \lstinline{add_edge} adds an edge of form \lstinline{(u,v,d)} to the graph.
  If adding an edge to the graph leads to a negative cycle,
  the function returns the cycle in form of a list of edges;
  otherwise, it returns \lstinline{None}.
  Furthermore, each edge added to the graph is associated with a decision level\footnote{The ASP solver's assignment comprises the decision level;
    it is incremented for each decision made and decremented for each decision undone while backjumping;
    initially, the decision level is zero.}.
  This additional information is used to backtrack to a previous state of the graph,
  whenever the solver has to backtrack to recover from a conflict.
\item
  Function \lstinline{backtrack} takes a decision level as argument.
  It removes all edges added on that level from the graph.
  For this to work, decision levels have to be backtracked in chronological order.
  Note that the CDCL algorithm in Figure~\ref{fig:cdcl} calling our propagator also backtracks decision levels in chronological order.
\item
  The \lstinline{Graph} class internally maintains an assignment of integers to nodes.
  This assignment can be turned into an assignment to the variables such that the difference constraints corresponding to the edges of the graph are satisfied.
  Function \lstinline{get_assignment} returns this assignment in form of a list of pairs of variables and integers.
\end{itemize}

The difference logic propagator implements the \lstinline{Propagator} interface (except for \lstinline{check}) in Figure~\ref{fig:interface}
in Lines~\ref{prg:dl:propagator:init:begin}--\ref{prg:dl:propagator:undo:end};
it features aspects like
incremental propagation and backtracking,
while supporting solving with multiple threads, and
multi-shot solving.
Whenever the set of edges associated with the current partial assignment of a solver induces a negative cycle
and, hence, the corresponding difference constraints are unsatisfiable,
it adds a nogood forbidding the negative cycle.
To this end,
it maintains data structures for detecting whether there is a conflict upon the addition of new edges.
More precisely, the propagator has three data members:
\begin{enumerate}
\item
  The \lstinline{self._l2e} dictionary in Line~\ref{prg:dl:propagator:member:l2e} maps solver literals
  for difference constraint theory atoms to their corresponding edges,\footnote{A solver literal might be associated with multiple edges.}
\item
  the \lstinline{self._e2l} dictionary in Line~\ref{prg:dl:propagator:member:e2l} maps edges back to solver literals,\footnote{In one solving step,
    the \clingo\ API guarantees that a (grounded) theory atom is associated with exactly one solver literal.
    Theory atoms grounded in later solving steps can be associated with fresh solver literals though.}
  and
\item
  the \lstinline{self._states} list in Line~\ref{prg:dl:propagator:member:state} stores for each solver thread its current graph
  with the edges assigned so far.
\end{enumerate}

Function \lstinline{init} in Lines~\ref{prg:dl:propagator:init:begin}--\ref{prg:dl:propagator:init:end}
sets up watches as well as the dictionaries \lstinline{self._l2e} and \lstinline{self._e2l}.
To this end,
it traverses the theory atoms over \lstinline{diff}$/1$ in Lines~\ref{prg:dl:propagator:init:loop:begin}--\ref{prg:dl:propagator:init:loop:end}.
Note that the loop simply ignores other theory atoms treated by other propagators.
In Lines~\ref{prg:dl:propagator:init:edge:begin}--\ref{prg:dl:propagator:init:edge:end},
we extract the edge from the theory atom.\footnote{
  For brevity, we omit the definition of the \lstinline{_eval} function,
  converting a theory term into a symbol.
  Furthermore, we assume that the user supplies valid theory atoms.
  A mature propagator checks validity and provides error messages.}
Each such atom is associated with a solver literal,
obtained in Line~\ref{prg:dl:propagator:init:map-literal}.
The mappings between solver literals and corresponding edges are then stored in the \lstinline{self._l2e} and \lstinline{self._e2l} dictionaries in
Lines~\ref{prg:dl:propagator:init:l2e} and~\ref{prg:dl:propagator:init:e2l}.\footnote{\python's \lstinline{setdefault} function is used to update the mappings.
  Depending on whether the given \lstinline{key} already appears in the dictionary,
  the function either retrieves the associated value or inserts and returns the second argument.}
In Line~\ref{prg:dl:propagator:init:watch} of the loop, a watch is added for each solver literal at hand,
so that the solver calls \lstinline{propagate} whenever the edge has to be added to the graph.
Up to here, we accommodated the non-strict semantics as we only consider the constraint occurring in the program and not its negation.
If the difference constraint atom occurs in the body, we impose the strict semantics,
meaning that, in case that the assigned literal is false,
we make sure that the negation of the difference constraint holds.
We check if the atom occurs in the body in Line~\ref{prg:dl:propagator:init:strict},
and if this is the case, we add an edge representing the negation of the difference constraint associated with the negated literal and watch the negated literal as well.

Function \lstinline{propagate}, given in Lines~\ref{prg:dl:propagator:propagate:begin}--\ref{prg:dl:propagator:propagate:end},
accesses \lstinline{control.thread_id} in Line~\ref{prg:dl:propagator:propagate:state}
to obtain the graph associated with the active thread.
The loops in Lines~\ref{prg:dl:propagator:propagate:loop:begin}--\ref{prg:dl:propagator:propagate:loop:end} then iterate over the list of changes and associated edges.
In Line~\ref{prg:dl:propagator:propagate:add-edge} each such edge is added to the graph.
If adding the edge produces a negative cycle,
a nogood is added in Line~\ref{prg:dl:propagator:propagate:add-nogood}.
Because an edge can be associated with multiple solver literals,
we use function \lstinline{_lit}
retrieving the first solver literal associated with an edge that is true,
to construct the nogood forbidding the cycle.
Given that the solver has to resolve the conflict and backjump,
the call to \lstinline{add_nogood} always yields false,
so that propagation is stopped without processing the remaining changes any further.\footnote{The optional arguments \lstinline{tag} and \lstinline{lock} of \lstinline{add_nogood} can be used to control the scope and lifetime of recorded nogoods.
  Furthermore, if a propagator adds nogoods that are not necessarily violated,
  function \lstinline{control.propagate} can be invoked to trigger unit propagation.}

Given that each edge added to the graph in Line~\ref{prg:dl:propagator:propagate:add-edge} is associated with the current decision level,
the implementation of function \lstinline{undo} is quite simple.
It calls function \lstinline{backtrack} on the solver's graph to remove all edges added on the current decision level.

\begin{remark}
  Here, we used a simplified \python\ version of the difference constraints propagator as a showcase.
  In practice, performance might fall short compared to solutions implemented in \C\ or \cpp.
  The \clingo\ package also offers the \lstinline{clingo.theory} module to load propagators implemented in other
  languages via the \emph{C Foreign Function Interface} for Python~\cite{cffi},
  thus combining convenient scripting with performance.
  The propagators of the extended ASP systems \clingoM{dl} and \clingcon\ can be loaded using this interface and
  can be used as a basis to implement customized ASP systems.
\end{remark}

\subsection{Solving flow shop problems}
\label{sec:flow:shop}

\begin{table}
\begin{figure}[H]
\centering
{\def\svgscale{.4}
\begingroup \makeatletter \providecommand\color[2][]{\errmessage{(Inkscape) Color is used for the text in Inkscape, but the package 'color.sty' is not loaded}\renewcommand\color[2][]{}}\providecommand\transparent[1]{\errmessage{(Inkscape) Transparency is used (non-zero) for the text in Inkscape, but the package 'transparent.sty' is not loaded}\renewcommand\transparent[1]{}}\providecommand\rotatebox[2]{#2}\ifx\svgwidth\undefined \setlength{\unitlength}{297.11999512bp}\ifx\svgscale\undefined \relax \else \setlength{\unitlength}{\svgscale\unitlength}\fi \else \setlength{\unitlength}{\svgwidth}\fi \global\let\svgwidth\undefined \global\let\svgscale\undefined \makeatother \begin{picture}(1,0.51534733)\put(0,0){\includegraphics[width=\unitlength,page=1]{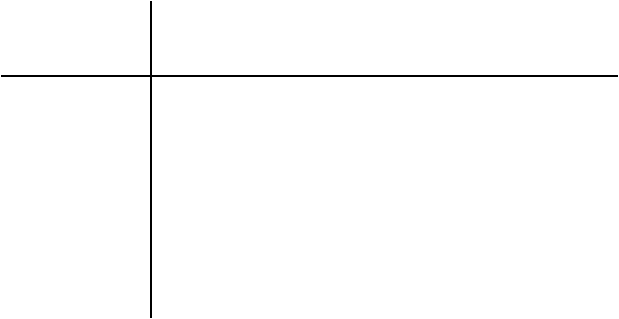}}\put(0.20382337,0.43268764){\color[rgb]{0,0,0}\makebox(0,0)[rb]{\smash{task}}}\put(0.28459882,0.43268764){\color[rgb]{0,0,0}\makebox(0,0)[lb]{\smash{duration on machine}}}\put(0.20382337,0.28459932){\color[rgb]{0,0,0}\makebox(0,0)[rb]{\smash{a}}}\put(0.20382337,0.16343615){\color[rgb]{0,0,0}\makebox(0,0)[rb]{\smash{b}}}\put(0.20382337,0.04227298){\color[rgb]{0,0,0}\makebox(0,0)[rb]{\smash{c}}}\put(0,0){\includegraphics[width=\unitlength,page=2]{figures/dl.pdf}}\end{picture}\endgroup  }
\caption{Flow shop instance with three tasks and two machines\label{fig:fs:ins}}
\end{figure}
\vspace{-1em} 
\begin{lstlisting}[label={prg:fs:ins},caption={Flow shop instance from Figure~\ref{fig:fs:ins} (\lstinline{fsI.lp})},language=clingos]
            machine(1).      machine(2).
task(a). duration(a,1,3). duration(a,2,4).
task(b). duration(b,1,1). duration(b,2,6).
task(c). duration(c,1,5). duration(c,2,5).
\end{lstlisting}
\begin{lstlisting}[label={prg:fs:enc},caption={Encoding of flow shop using difference constraints (\lstinline{fsE.lp})},linerange={1-16},language=clingos]
1 { cycle(T,U): task(U), U!=T } 1 :- task(T).%%#(\label{prg:fs:perm:begin}#)
1 { cycle(T,U): task(T), U!=T } 1 :- task(U).
reach(M) :- M = #min { T: task(T) }.
reach(U) :- reach(T), cycle(T,U).
:- task(T), not reach(T).

1 { start(T): task(T) } 1.
permutation(T,U) :- cycle(T,U), not start(U).%%#(\label{prg:fs:perm:end}#)

seq((T,M),(T,M+1),D) :- task(T), duration(T,M,D), machine(M+1).%%#(\label{prg:fs:diff:begin}\label{prg:fs:permutation:seq:machine}#)
seq((T1,M),(T2,M),D) :- permutation(T1,T2), duration(T1,M,D).%%#(\label{prg:fs:permutation:seq:task}#)

&diff { T1-T2 } <= -D :- seq(T1,T2,D).%%#(\label{prg:fs:permutation:seq}#)
&diff { 0-(T,M) } <= 0 :- duration(T,M,D).%%#(\label{prg:fs:diff:end}\label{prg:fs:null}#)

#show permutation/2.
\end{lstlisting}
\begin{figure}[H]
\centering
{\def\svgwidth{\linewidth}
\begingroup \makeatletter \providecommand\color[2][]{\errmessage{(Inkscape) Color is used for the text in Inkscape, but the package 'color.sty' is not loaded}\renewcommand\color[2][]{}}\providecommand\transparent[1]{\errmessage{(Inkscape) Transparency is used (non-zero) for the text in Inkscape, but the package 'transparent.sty' is not loaded}\renewcommand\transparent[1]{}}\providecommand\rotatebox[2]{#2}\ifx\svgwidth\undefined \setlength{\unitlength}{948.00009766bp}\ifx\svgscale\undefined \relax \else \setlength{\unitlength}{\svgscale\unitlength}\fi \else \setlength{\unitlength}{\svgwidth}\fi \global\let\svgwidth\undefined \global\let\svgscale\undefined \makeatother \begin{picture}(1,0.45569621)\put(0,0){\includegraphics[width=\unitlength,page=1]{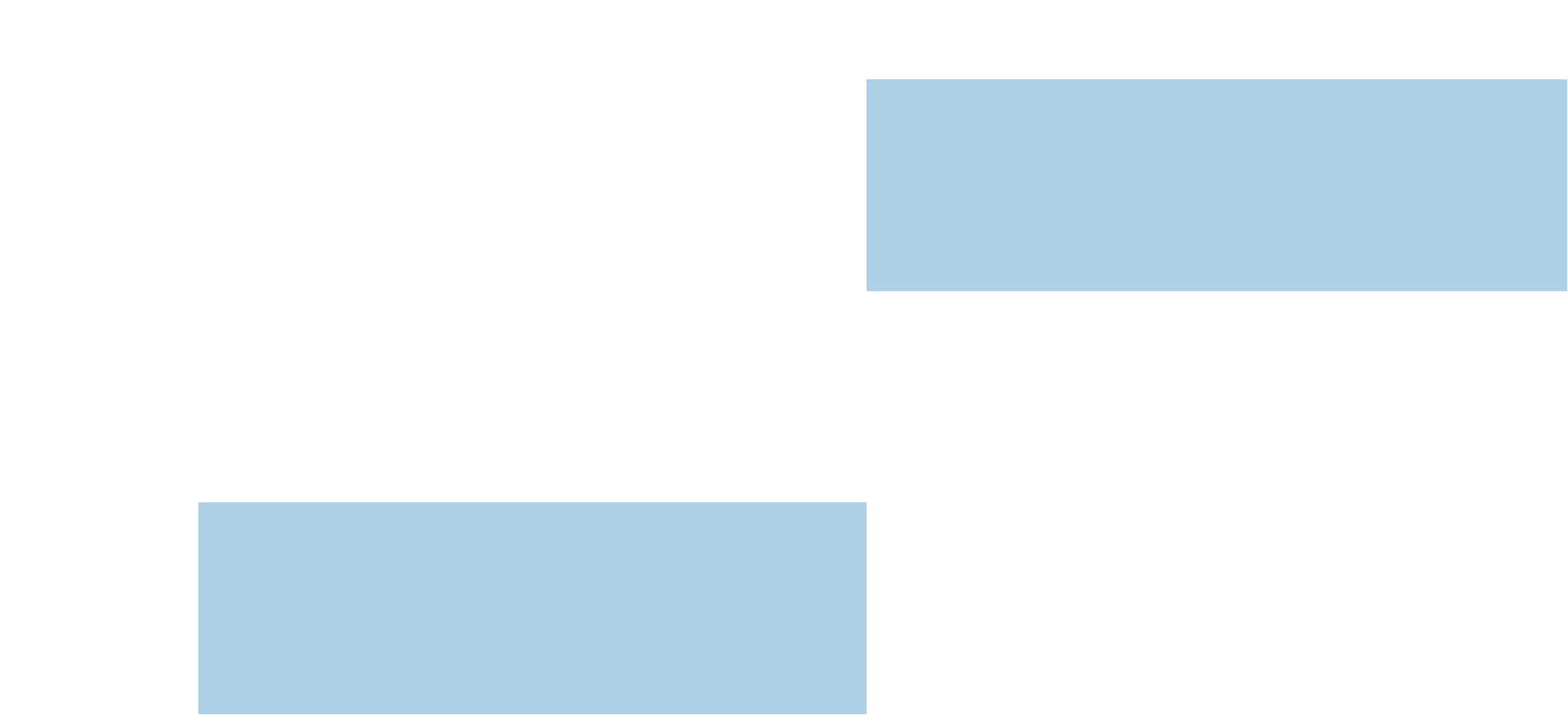}}\put(0.13924056,0.23206748){\color[rgb]{0,0,0}\makebox(0,0)[lb]{\smash{a < c < b}}}\put(0,0){\includegraphics[width=\unitlength,page=2]{figures/dl-sol.pdf}}\put(0.1139241,0.18565396){\color[rgb]{0,0,0}\makebox(0,0)[rb]{\smash{1}}}\put(0.1139241,0.14767931){\color[rgb]{0,0,0}\makebox(0,0)[rb]{\smash{2}}}\put(0,0){\includegraphics[width=\unitlength,page=3]{figures/dl-sol.pdf}}\put(0.1139241,0.41772158){\color[rgb]{0,0,0}\makebox(0,0)[rb]{\smash{machine}}}\put(0.13924056,0.36708866){\color[rgb]{0,0,0}\makebox(0,0)[lb]{\smash{a < b < c}}}\put(0,0){\includegraphics[width=\unitlength,page=4]{figures/dl-sol.pdf}}\put(0.1139241,0.32067507){\color[rgb]{0,0,0}\makebox(0,0)[rb]{\smash{1}}}\put(0.1139241,0.28270039){\color[rgb]{0,0,0}\makebox(0,0)[rb]{\smash{2}}}\put(0,0){\includegraphics[width=\unitlength,page=5]{figures/dl-sol.pdf}}\put(0.13924056,0.41772158){\color[rgb]{0,0,0}\makebox(0,0)[lb]{\smash{solution}}}\put(0,0){\includegraphics[width=\unitlength,page=6]{figures/dl-sol.pdf}}\put(0.13924056,0.0970464){\color[rgb]{0,0,0}\makebox(0,0)[lb]{\smash{b < a < c}}}\put(0,0){\includegraphics[width=\unitlength,page=7]{figures/dl-sol.pdf}}\put(0.1139241,0.0506329){\color[rgb]{0,0,0}\makebox(0,0)[rb]{\smash{1}}}\put(0.1139241,0.01265828){\color[rgb]{0,0,0}\makebox(0,0)[rb]{\smash{2}}}\put(0.5696203,0.23206748){\color[rgb]{0,0,0}\makebox(0,0)[lb]{\smash{c < a < b}}}\put(0,0){\includegraphics[width=\unitlength,page=8]{figures/dl-sol.pdf}}\put(0.5696203,0.36708846){\color[rgb]{0,0,0}\makebox(0,0)[lb]{\smash{b < c < a}}}\put(0,0){\includegraphics[width=\unitlength,page=9]{figures/dl-sol.pdf}}\put(0.5696203,0.0970463){\color[rgb]{0,0,0}\makebox(0,0)[lb]{\smash{c < b < a}}}\put(0,0){\includegraphics[width=\unitlength,page=10]{figures/dl-sol.pdf}}\put(0.54008433,0.36708862){\color[rgb]{0,0,0}\makebox(0,0)[rb]{\smash{18}}}\put(0.54008433,0.23206754){\color[rgb]{0,0,0}\makebox(0,0)[rb]{\smash{19}}}\put(0.54008433,0.09704645){\color[rgb]{0,0,0}\makebox(0,0)[rb]{\smash{16}}}\put(0.98734167,0.36708862){\color[rgb]{0,0,0}\makebox(0,0)[rb]{\smash{16}}}\put(0.98734167,0.23206754){\color[rgb]{0,0,0}\makebox(0,0)[rb]{\smash{20}}}\put(0.98734167,0.09704645){\color[rgb]{0,0,0}\makebox(0,0)[rb]{\smash{20}}}\end{picture}\endgroup  }
\vspace{-.2em}
\caption{Flow shop solutions for all possible permutations with the total execution length in the top right corner and optimal solutions with a blue background\label{fig:fs:sol}}
\end{figure}
\end{table}

To see our propagator in action, we consider the flow shop problem,
dealing with a set of tasks $T$ that have to be consecutively executed on $m$ machines.
Each task has to be processed on each machine from $1$ to $m$.
Different parts of one task are completed on each machine resulting in the completion of the task after execution on all machines is finished.
Before a task can be processed on machine $i$, it has to be finished on machine $i-1$.
The duration of different tasks on the same machine may vary.
A task can only be executed on one machine at a time and
a machine must not be occupied by more than one task at a time.
An (optimal) solution to the problem is a permutation of tasks so that all tasks are finished as early as possible.

Figure~\ref{fig:fs:ins} depicts a possible instance for the flow shop problem.
The three tasks \lstinline{a}, \lstinline{b}, and \lstinline{c} have to be scheduled on two machines.
The colored boxes indicate how long a task has to run on a machine.
Lighter shades of the same color are for the first and darker ones for the second machine.
For example, task \lstinline{a} needs to be processed for~$3$ time units on the first and~$4$ time units on the second machine.

Next, we encode this problem using ASP with difference constraints.
We give in Listing~\ref{prg:fs:ins} a straightforward encoding of the instance in Figure~\ref{fig:fs:ins}.
Listing~\ref{prg:fs:enc} provides the encoding of the flow shop problem.
Following the generate, define, and test methodology of ASP~\cite{lifschitz19a},
we first generate in Lines~\ref{prg:fs:perm:begin}--\ref{prg:fs:perm:end} all possible permutations of tasks,
where atoms of the form \lstinline{permutation(T,U)} encode that task~$T$ has to be executed before task~$U$.
Then, in the following Lines~\ref{prg:fs:diff:begin}--\ref{prg:fs:diff:end},
we use difference constraints to calculate the duration of the generated permutation.
The difference constraint in Line~\ref{prg:fs:permutation:seq} guarantees that the tasks are executed in the right order.
For example, $\text{\lstinline{(a,1)}} - \text{\lstinline{(a,2)}} \leq -d$ ensures that task~\lstinline{a} can only be executed on machine~\lstinline{2} if it has finished on machine~\lstinline{1}.
Hence, the variable \lstinline{(a,2)} has to be assigned so that it is greater or equal to $\text{\lstinline{(a,1)}}+d$ where $d$ is the duration of task \lstinline{a} on machine \lstinline{1}.
Similarly, $\text{\lstinline{(a,1)}} - \text{\lstinline{(b,1)}} \leq -d$ makes sure that task~\lstinline{b} can only be executed on machine~\lstinline{1} if task~\lstinline{a} has finished on machine~\lstinline{1}.
While the first constraint results in a set of facts (see Line~\ref{prg:fs:permutation:seq:machine}),
the latter is subject to the generated permutation of tasks (see Line~\ref{prg:fs:permutation:seq:task}).
The difference constraint in Line~\ref{prg:fs:null} ensures that all time points at which a task is started are greater than zero.
Note that this constraint is in principle redundant
but since sets of difference constraints always have infinitely many solutions
it is good practice to encode relative to a starting point.
Furthermore, note that~\lstinline{0} is actually a variable.
In fact, the \lstinline{Graph} class takes care of subtracting the value of variable~\lstinline{0} from all other variables when returning an assignment
to get easier interpretable solutions.

Running encoding and instance with the \lstinline{dl} propagator results in the following six solutions
corresponding to the solutions in Figure~\ref{fig:fs:sol}.\footnote{Note that in each solution all tasks are executed as early as possible.
  This is no coincidence and actually guaranteed by the algorithm implemented in the \lstinline{Graph} class.}
One for each possible permutation of tasks:
\begin{lstlisting}[language=shell]
UNIX> python dl-app.py fsE.lp fsI.lp 0
clingo-dl version 1.0
Reading from fsE.lp ...
Solving...
Answer: 1
permutation(b,a) permutation(a,c) dl((a,1),1) dl((a,2),7)  \
dl((b,1),0) dl((b,2),1) dl((c,1),4) dl((c,2),11)
Answer: 2
permutation(b,a) permutation(c,b) dl((a,1),6) dl((a,2),16) \
dl((b,1),5) dl((b,2),10) dl((c,1),0) dl((c,2),5)
Answer: 3
permutation(c,b) permutation(a,c) dl((a,1),0) dl((a,2),3)  \
dl((b,1),8) dl((b,2),13) dl((c,1),3) dl((c,2),8)
Answer: 4
permutation(b,c) permutation(c,a) dl((a,1),6) dl((a,2),12) \
dl((b,1),0) dl((b,2),1) dl((c,1),1) dl((c,2),7)
Answer: 5
permutation(b,c) permutation(a,b) dl((a,1),0) dl((a,2),3)  \
dl((b,1),3) dl((b,2),7) dl((c,1),4) dl((c,2),13)
Answer: 6
permutation(c,a) permutation(a,b) dl((a,1),5) dl((a,2),10) \
dl((b,1),8) dl((b,2),14) dl((c,1),0) dl((c,2),5)
SATISFIABLE
\end{lstlisting}

\subsection{Hybrid optimization with difference constraints}
\label{sec:dl:optimization}

\begin{lstlisting}[float=tp,label={prg:dl:theory:opt},caption={Application class \lstinline{DLOptApp} for difference constraints with optimization (\lstinline{dlO-app.py})},language=pythons]
import sys
from clingo.application import Application, clingo_main
from clingo.ast import ProgramBuilder, parse_files
from dl import DLPropagator, HeadBodyTransformer, THEORY

class DLOptApp(Application):
    program_name = "clingo-dl-opt"
    version = "1.0"

    def __init__(self):
        self._bound = None
        self._propagator = DLPropagator()

    def _on_model(self, model):
        self._propagator.on_model(model)
        for symbol in model.symbols(theory=True):
            if symbol.match("dl", 2):
                n, v = symbol.arguments
                if n.match("bound", 0):
                    self._bound = v.number#(\label{prg:dl:theory:opt:get-bound}#)
                    break

    def main(self, ctl, files):
        ctl.register_propagator(self._propagator)
        ctl.add("base", [], THEORY)
        ctl.add("bound", ["b"], "&diff(head) { bound-0 } <= b.")#(\label{prg:dl:theory:opt:bound}#)

        with ProgramBuilder(ctl) as bld:
            hbt = HeadBodyTransformer()
            parse_files(files, lambda stm: bld.add(hbt.visit(stm)))

        ctl.ground([("base", [])])
        while ctl.solve(on_model=self._on_model).satisfiable:
            print("Found new bound: {}".format(self._bound))
            ctl.ground([("bound", [self._bound - 1])])

        if self._bound is not None: print("Optimum found")

sys.exit(int(clingo_main(DLOptApp(), sys.argv[1:])))
\end{lstlisting}
Finally, to find optimal solutions,
we combine the algorithms in Listings~\ref{prg:opt:main} and~\ref{prg:dl:theory} to minimize the total execution time of the tasks.
The resulting algorithm is given in Listing~\ref{prg:dl:theory:opt}.
As with the algorithm in Listing~\ref{prg:dl:theory},
a propagator and theory language is registered before solving and the program is parsed to accommodate a uniform semantic treatment.
The control flow is similar to the branch-and-bound-based optimization algorithm in Listing~\ref{prg:opt:main}
except that we now minimize the variable \lstinline{bound}.
More precisely,
we minimize the difference between variable~\lstinline{0} and~\lstinline{bound}
by adding the difference constraint $\text{\lstinline{0}} - \text{\lstinline{bound}} \leq b$ to the program in Line~\ref{prg:dl:theory:opt:bound}
where $b$ is the best known execution time of the tasks as obtained from the assignment in Line~\ref{prg:dl:theory:opt:get-bound} minus $1$.
To bound the maximum execution time of the tasks,
we have to add one more line to the encoding in Listing~\ref{prg:fs:enc}:
\begin{lstlisting}[language=clingos,firstnumber=18]
  &diff { (T,M)-bound } <= -D :- duration(T,M,D).
\end{lstlisting}
This makes sure that each task ends within the given bound.
Running encoding and instance with the \lstinline{dl} propagator results in the optimum bound~$16$ where
the obtained solution corresponds to the lower left of the two optimal solutions indicated by a light blue background in Figure~\ref{fig:fs:sol}:
\begin{lstlisting}[language=shell]
UNIX> python dlO-app.py fsE.lp fsI.lp
clingo-dl-opt version 1.0
Reading from fsE.lp ...
Solving...
Answer: 1
permutation(b,a) permutation(a,c) dl(bound,16) dl((a,1),1)   \
dl((a,2),7) dl((b,1),0) dl((b,2),1) dl((c,1),4) dl((c,2),11)
Found new bound: 16
Solving...
Optimum found
UNSATISFIABLE
\end{lstlisting}
\hphantom{x}\\  \section{Guess-and-check programming reloaded}
\label{sec:gnt}

Finally, we present an implementation of guess-and-check programming that relies on a combination of two \clingo\
solvers.
In contrast to the approach taken in Section~\ref{sec:meta:gc},
where the logic programs comprising the guess and check parts are combined in a single disjunctive program and thus
solved by a single solver,
the idea is now to deal with both programs separately by means of two interacting solvers.
This last case-study nicely contrasts the efforts involved in meta-programming and the usage of solver APIs.
Also, it further illustrates features of \clingo's API, namely,
the manipulation of a program's abstract syntax tree (AST),
the interaction of (multi-threaded) \clingo\ instances via the propagator interface,
the usage of assumptions during solving, and
the addition of constraints to a program during runtime.

Unlike Section~\ref{sec:meta:gc},
we use \lstinline{#program} directives to declare rules belonging to the guess and check programs.
As above, guess atoms must not occur among the head atoms of the check program.
For example, the simple guess and check programs from Listings~\ref{prg:meta:guess} and~\ref{prg:meta:check}
can now be rolled into one, as shown in Listing~\ref{prg:gnt:example}.

\begin{lstlisting}[float,language=clingos,caption={Guess-and-check program (\lstinline{guess-check.lp})},label={prg:gnt:example}]
#program guess.
1 { a(1..2) }.

#program check.
:- not a(1).
\end{lstlisting}

Passing this to our guess-and-check application \lstinline{app.py} yields the same solution as with meta-programming:
\begin{lstlisting}[language=shell]
UNIX> python app.py guess-check.lp 0
guess-and-check version 1.0
Reading from guess-check.lp
Solving...
Answer: 1
a(2)
SATISFIABLE
\end{lstlisting}

In what follows,
we detail the inner working of the approach.
The idea is to have one solver guessing solution candidates, and another checking their compliance.
Their interaction is realized through \clingo's propagator interface and restricted to testing total candidates,
rather than partial ones as done in Section~\ref{sec:case}.
In this way, the checking solver acts as a propagator within the guessing one.

The overall design is partitioned in four classes.
Our description concentrates on these classes by following the overall workflow,
although the line numbers reflect positions in the source code.

As before, we start from a derivative of \clingo's \lstinline{Application} class
and implement its \lstinline{main} function as shown in Listing~\ref{prg:app:app}.
\begin{lstlisting}[float,caption={The \lstinline{GACApp} class for guess-and-check programming (\lstinline{app.py}, Lines {\ref{prg:app:begin}--\ref{prg:app:end}})},label={prg:app:app},linerange={98-111},firstnumber=98,language=pythons]
class GACApp(Application):#(\label{prg:app:begin}#)
    def __init__(self):
        self.program_name = "guess-and-check"
        self.version = "1.0"

    def main(self, ctl, files):#(\label{prg:app:main:begin}#)
        check = []#(\label{prg:app:main:check:ast:init}#)
        with ast.ProgramBuilder(ctl) as builder:
            trans = Transformer(builder, check)#(\label{prg:app:main:transformer}#)
            ast.parse_files(files, trans.add)#(\label{prg:app:main:parse}#)
        ctl.register_propagator(GACPropagator(check))#(\label{prg:app:main:propagator:registration}#)

        ctl.ground([("base", [])])
        ctl.solve()#(\label{prg:app:main:end}\label{prg:app:end}#)
\end{lstlisting}
As mentioned,
the primary solver object \lstinline{ctl} acts as the guesser,
while the checker is encapsulated as its propagator.
The \lstinline{main} function
starts by parsing the input programs,
registers the propagator,
grounds, and
solves.
The task of the \lstinline{Transformer} in Line~\ref{prg:app:main:transformer} is
to add rules from the guess part to the program in the primary solver \lstinline{ctl} via a \lstinline{ProgramBuilder} and
to collect rules from the check part in the list initialized in  Line~\ref{prg:app:main:check:ast:init}.
This is done during parsing in Line~\ref{prg:app:main:parse} by means of the \lstinline{add} function
defined in the \lstinline{Transformer} class.

\begin{lstlisting}[float,caption={The \lstinline{Transformer} class for classifying rules into the guess and check part (\lstinline{app.py}, Lines {\ref{prg:app:trans:begin}--\ref{prg:app:trans:end}})},label={prg:app:ast},firstnumber=8,linerange={8-27},language=pythons]
class Transformer:#(\label{prg:app:trans:begin}#)
    def __init__(self, builder, check):
        self._builder = builder
        self._state = "guess"
        self._check = check

    def add(self, stm):#(\label{prg:app:ast:add:begin}#)
        if stm.ast_type == ast.ASTType.Program:
            if stm.name == "check" and not stm.parameters:
                self._state = "check"#(\label{prg:app:ast:add:state:check}#)
            elif stm.name in ("base", "guess") and not stm.parameters:
                self._state = "guess"#(\label{prg:app:ast:add:state:guess}#)
            else:
                raise RuntimeError("unexpected program part")

        else:
            if self._state == "guess":
                self._builder.add(stm)#(\label{prg:app:ast:add:builder:add}#)
            else:
                self._check.append(stm)#(\label{prg:app:ast:add:append}\label{prg:app:ast:add:end}\label{prg:app:trans:end}#)
\end{lstlisting}
The \lstinline{add} function is given in Lines~\ref{prg:app:ast:add:begin} to~\ref{prg:app:ast:add:end} of Listing~\ref{prg:app:ast}
as the salient part of the \lstinline{Transformer} class.
It relies on variable \lstinline{_state} to distinguish whether a rule is read in the context
of a \lstinline{guess} (or \lstinline{base}) or \lstinline{check} program.
This variable is ``toggled'' in Lines~\ref{prg:app:ast:add:state:check} and~\ref{prg:app:ast:add:state:guess}
whenever a \lstinline{#program} directive is encountered.
Accordingly,
the rule's AST is either
added to the program builder of the guessing solver in Line~\ref{prg:app:ast:add:builder:add} or
appended to the \lstinline{check} list (in Line~\ref{prg:app:ast:add:append})
that is passed down from the \lstinline{main} function to gather the check program.

Once parsing is finished,
the filled list is used to initialize the propagator in Line~\ref{prg:app:main:propagator:registration}.
The corresponding \lstinline{GACPropagator} class is given in Listing~\ref{prg:app:pro}.
It administers one or several solver objects,
which are encapsulated by the \lstinline{Checker} class in Listing~\ref{prg:app:chk}.
\begin{lstlisting}[float,caption={The \lstinline{GACPropagator} class interfacing guessing and checking solver (\lstinline{app.py}, Lines {\ref{prg:app:propagator:begin}--\ref{prg:app:propagator:end}})},label={prg:app:pro},firstnumber=63,linerange={63-96},language=pythons]
class GACPropagator(Propagator):#(\label{prg:app:propagator:begin}#)
    def __init__(self, check):
        self._check = check
        self._checkers = []

    def init(self, init):#(\label{prg:app:pro:init:begin}#)
        for _ in range(init.number_of_threads):
            checker = Checker()#(\label{prg:app:pro:init:checker}#)
            self._checkers.append(checker)

            with checker.backend() as backend:
                for atom in init.symbolic_atoms:
                    guess_lit = init.solver_literal(atom.literal)
                    if init.assignment.is_false(guess_lit):#(\label{prg:app:pro:init:false}#)
                        continue

                    check_lit = backend.add_atom(atom.symbol)
                    if init.assignment.is_true(guess_lit):
                        backend.add_rule([check_lit], [])#(\label{prg:app:pro:init:true}#)
                    else:
                        backend.add_rule([check_lit], [], True)#(\label{prg:app:pro:init:choice}#)
                        checker.add(guess_lit, check_lit)#(\label{prg:app:pro:init:dictionary}#)

            checker.ground(self._check)#(\label{prg:app:pro:init:end}#)

    def check(self, control):
        assignment = control.assignment
        checker = self._checkers[control.thread_id]

        if not checker.check(control):#(\label{prg:app:pro:checker:call}#)
            conflict = []
            for level in range(1, assignment.decision_level+1):
                conflict.append(-assignment.decision(level))
            control.add_clause(conflict)#(\label{prg:app:propagator:end}#)
\end{lstlisting}
\begin{lstlisting}[float,caption={The \lstinline{Checker} class wrapping the checking solver (\lstinline{app.py}, Lines {\ref{prg:app:checker:begin}--\ref{prg:app:checker:end}})},label={prg:app:chk},firstnumber=29,linerange={29-61},language=pythons]
class Checker:#(\label{prg:app:checker:begin}#)
    def __init__(self):
        self._ctl = Control()
        self._map = []

    def backend(self):
        return self._ctl.backend()

    def add(self, guess_lit, check_lit):#(\label{prg:app:chk:add}#)
        self._map.append((guess_lit, check_lit))

    def ground(self, check):#(\label{prg:app:chk:ground}#)
        with ast.ProgramBuilder(self._ctl) as builder:
            for stm in check:
                builder.add(stm)

        self._ctl.ground([("base", [])])

    def check(self, control):
        assignment = control.assignment#(\label{prg:app:chk:assigment}#)

        assumptions = []
        for guess_lit, check_lit in self._map:
            if assignment.is_true(guess_lit):
                assumptions.append(check_lit)
            else:
                assumptions.append(-check_lit)

        ret = self._ctl.solve(assumptions)#(\label{prg:app:chk:solver:call}#)
        if ret.unsatisfiable is not None:
            return ret.unsatisfiable

        raise RuntimeError("search interrupted")#(\label{prg:app:checker:end}#)
\end{lstlisting}
Given that no partial checks are performed, the propagation class only implements function \lstinline{init}
and \lstinline{check} of \clingo's \lstinline{Propagator} interface from Figure~\ref{fig:interface}.

Let us first detail the initialization of the checkers in Lines~\ref{prg:app:pro:init:begin} to~\ref{prg:app:pro:init:end}.
In fact, the \lstinline{init} function may create several instances of the \lstinline{Checker} class,
depending on the number of threads of the primary solver.
Each such checker (cf.\ Line~\ref{prg:app:pro:init:checker}) is initialized by looping over the atoms of the primary solver
that provide the respective guess.
While atoms are dropped in Line~\ref{prg:app:pro:init:false}
that have been found to be false after grounding (and pre-processing) the guess program,
either a fact or a choice rule is added to the checker in Lines~\ref{prg:app:pro:init:true} and~\ref{prg:app:pro:init:choice}
depending on whether the atom was found to be true or unknown, respectively.
Clearly, unknown atoms of the guesser are most relevant to the checking solver,
since their truth value is still subject to change.
To this end, each checker comprises a dictionary mapping (unknown) guess literals to check literals;
it is filled in Line~\ref{prg:app:pro:init:dictionary}.\footnote{In Section~\ref{sec:meta}, this was done via predicate \lstinline{guess/1}.}
Once all facts and choice rules are added to the checker,
the check program gathered during parsing is grounded in Line~\ref{prg:app:pro:init:end} and added as well.
The corresponding \lstinline{add} and \lstinline{ground} functions are defined in the \lstinline{Checker} class
and implemented in a straightforward way in the lines following Lines~\ref{prg:app:chk:add} and~\ref{prg:app:chk:ground}, respectively.

Just the same way as during initialization,
both \lstinline{GACPropagator} and \lstinline{Checker} work hand in hand in their respective \lstinline{check} functions.
The propagator's \lstinline{check} function is called once the guessing solver has found a stable model;
it immediately calls the \lstinline{check} function of the associated checker
in Line~\ref{prg:app:pro:checker:call}.
In doing so, it passes along the
\lstinline{Control} object providing (limited) access to the underlying solver.
This includes access to the assignment of the solver which of course corresponds to a model.
The latter is at once extracted upon entering the checker's \lstinline{check} function in Line~\ref{prg:app:chk:assigment}
and analyzed afterwards.
To this end, the function loops over the dictionary associating (originally unknown) guess and check atoms
to transfer the guessed literals into a list of checker literals that are then used as assumptions
in the subsequent call of the checker in Line~\ref{prg:app:chk:solver:call}.
Technically, assumptions are added to the solver's assignment and
amount semantically to the addition of integrity constraints (unlike externals;\footnote{Also, assumptions only affect the current solve call.
  Opposed to this,
  assignments to externals persist over solve call (as long as the externals are not released, reassigned, or defined).}
cf.\ Section~\ref{sec:glance})
In this way, the checker is forced to search for stable models comprising all guessed literals.
If this fails, the checker's \lstinline{check} succeeds,
as does the propagator's \lstinline{check}.
Otherwise,
the propagator extracts from the guesser's stable model all underlying decision literals and adds them as an integrity
constraint, thus eliminating the combination of literals from the search space.

\section{Discussion}\label{sec:discussion}

This tutorial aims at enabling ASP users to become ASP engineers.

Our role model has been the landmark paper by \citeANP{eensor03a}~\citeNN{eensor03a}
that aimed at
{\em``give{\em[ing]} sufficient details about implementation to enable the reader to construct his or her own solver in a very short time.
This will allow users of SAT-solvers to make domain specific extensions or adaptions of current state-of-the-art SAT-techniques,
to meet the needs of a particular application area.''}
Their presentation of the \cpp\ source code of the SAT solver \minisat\ significantly boosted research in SAT by
{\em``bridge{\em[ing]} the gap between existing descriptions of SAT-techniques
and their actual implementation''}~\cite{eensor03a}.

We hope to achieve a similar effect with the tutorial at hand.
However,
unlike following suit in easing a white box approach to ASP solving,
dealing with system modifications,
we rather advocate the gray and black box approach put forward in the introduction,
and make a case for application interface and meta programming, respectively.
This is motivated by the much more elaborate model-ground-solve workflow of ASP systems
that must often be addressed in its entirety to provide a certain functionality.

To this end, we describe several essential techniques for
extending the ASP system \clingo\ or implementing customized special-purpose systems.
We have started with the lighter approach of meta programming in ASP
and continued with application interface programming in \python\
(although several alternatives are available).
Central to this is the new \lstinline{Application} class of \clingo\ that
permits to draw on \clingo's infrastructure by starting processes similar to the one in \clingo.
This allows us to build customized ASP-based systems by overriding \clingo's \lstinline{main} function,
as illustrated by various examples throughout the tutorial.
In particular, we have seen how derivatives of the \lstinline{Application} class can be used to
engage manipulations to programs' abstract syntax trees,
control various forms of multi-shot solving, and
set up theory propagators for foreign inferences.
Multi-shot solving provides us with fine-grained control of ASP reasoning processes, while
theory solving allows us to refine basic ASP solving by incorporating foreign types of constraints.
Because of ASP's model-ground-solve methodology both techniques pervade its whole workflow,
starting with extensions to the input language, over means for incremental and theory-enhanced grounding,
to stateful and theory-enhanced solving.
Multi-shot solving even adds a fourth dimension to \emph{control} ASP reasoning processes.

Although meta programming has been around in ASP since its beginning
(cf.\ Section~\ref{sec:meta} for a brief discussion),
we hope that the reification feature of \clingo\ makes it more attractive
as a lightweight alternative to extend ASP systems.
The idea of implementing ASP systems by pipelining was first advocated by Tomi Janhunen and
used in his normalization toolbox~\cite{jannie11a,bogeja14a,bogeja16a}.
In both cases, an intermediate ASP format is used to pass data from one solver to the next.
While we used a fact-based representation of \aspif,
the normalization tools rely on the machine-oriented \smodels\ format.
Interestingly, \lctocasp~\cite{cakaossc16a} implements a system for non-monotonic constraint solving by
translating one \aspif\ specification into another.
That is, it takes the output of \gringo, compiles out non-monotonicity, and feeds the result into \clingcon,
an extension of \clingo\ with monotonic linear constraints over integers.

As mentioned in the introduction,
\dlv\ and \clingo\ constitute nowadays the only genuine ASP systems in use.
Accordingly,
they are the only possible providers of native APIs for ASP.
As detailed by~\citeANP{alcadofuleperiveza17a}~\citeNN{alcadofuleperiveza17a},
the latest versions of \dlv\ combine the \idlv\ grounder~\cite{cafupeza17a} with the ASP solver \wasp~\cite{aldoleri15a}.
As with \gringo,
the input language of \idlv\ covers the second ASP language standard~\cite{cafageiakakrlemarisc19a}.
Furthermore, it offers the integration of computable functions, similar to the mechanism (using terms preceded by `\lstinline{@}')
sketched at the beginning of Section~\ref{sec:app}.
Unlike this, a full-fledged \python\ API is offered by the ASP solver \wasp~\cite{dodric20a}.
This has interesting applications to heuristic control and propagators for integrity
constraints~\cite{dogalemurisc16a,cudorisc20a}.
In fact, former versions of \dlv\ offer powerful \java\ integration~\cite{felegrri12a},
compliant with the Object-Relational Mapping standard (ORM),
and implemented by wrapping \dlv\ at its core.
A \python\ library providing an ORM interface to \clingo\ is also available~\cite{clorm}.
A framework for developing applications embedding ASP on mobile devices is proposed
by~\citeANP{fugezaancape16a}~\citeNN{fugezaancape16a}.

Prior to the availability of APIs, various systems extending ASP have been built.
For example,
\dlvhex~\cite{redl16a,eigeiakarescwe18a} provides higher-order logic programs,
whose higher-order atoms are implemented externally in \cpp\ or \python;
it is build upon \clingo's infrastructure.
At the time, such a white box approach was only feasible thanks to a close collaboration between both
groups at Vienna and Potsdam, not to mention that this fostered the development of \clingo's API quite a bit.
Interestingly, also \clingcon~\cite{bakaossc16a}, an extension of \clingo\ with linear constraints over integers,
started out as a white box approach and has just recently been transformed into a gray box approach,
since otherwise its maintenance had been infeasible.
Similarly yet much earlier, \adsolver~\cite{megezh08a} extended \smodels~\cite{niesim97a}
with linear constraints over integers.
Another category of ASP systems,
such as \ezsmt~\cite{liesus16a}, \dingo~\cite{jalini11a}, and \aspmt~\cite{barlee14b}
translate ASP with constraints to SMT and use appropriate backends.
Similarly, \mingo~\cite{lijani12a} translates to Mixed Integer Linear Programming.
Interestingly,
the semantics of such hybrid ASP systems can be given in a theory-independent way by using
denotational semantics~\cite{cakaossc16a,cafascwa20b}.

Both meta and application interface programming greatly facilitate the development of ASP-based systems
and therefore ease the transposition of ideas into practice.
The lighter approach of meta programming is well suited for rapid prototyping and moreover enjoys
elaboration tolerance.
However, once more control is needed, API programming is indispensable.
Although it lacks full elaboration tolerance,
it has nonetheless the great advantage to offer a high level of abstraction.
This makes any project much easier to handle and to maintain than modifying the source code of an ASP system.

Last but not least,
ASP has come a long way to turn into a mature and quite sophisticated approach to declarative problem solving.
However, this sophistication should not become an obstacle to further technological advances.
We hope that this tutorial contributes to coping with this challenge.
After all, we, the ASP community, have the hard job of making our users' lives easy.

\subsubsection*{Acknowledgments}

This work was partially funded by DFG grants SCHA 550/11 and~15.

We are grateful to the anonymous reviewers and in particular to Mirosław Truszczy{\'n}ski for his relentless efforts to improve the presentation of our paper ---
thank you so much!

 \appendix

\section{Saturation-based meta encoding}
\label{sec:meta:D}

The saturation-based meta encoding in Listing~\ref{prg:meta:D} relies on a partition of the
atoms of the input program induced by the strongly connect components of its positive dependency graph.
Each atom and each loop of the program is contained in some part.
The idea is to mimic the consecutive application of the immediate consequence operator to each component of the partition.

In a nutshell, the encoding in Listing~\ref{prg:meta:D} combines the following parts~\cite{gekasc11b}:
\begin{enumerate}
\item guessing an interpretation (in Lines~\ref{metaD:interpretation:begin} to~\ref{metaD:interpretation:end}),
\item deriving the unsatisfiability-indicating atom \lstinline{bot} if the interpretation is not a supported model
  (where each true atom occurs positively in the head of some rule whose body holds;
  cf.\ Lines~\ref{metaD:interpretation:support:begin} and~\ref{metaD:interpretation:support:end}),
\item deriving \lstinline{bot} if the true atoms of some non-trivial strongly connected component
  are not acyclicly derivable
  (checked via determining the complement of a fixpoint of the immediate consequence operator;
  cf.\ Lines~\ref{metaD:interpretation:acyclicity:begin} to~\ref{metaD:interpretation:acyclicity:end}), and
\item saturating interpretations that do not correspond to stable models
  by deriving all truth assignments (for atoms) from \lstinline{bot}
  (in Lines~\ref{metaD:saturation:true} and~\ref{metaD:saturation:fail}).
\end{enumerate}
\begin{lstlisting}[float=htp,caption={A disjunctive meta encoding implementing saturation (\texttt{metaD.lp})},label={prg:meta:D},language=clingo,basicstyle=\scriptsize\ttfamily]
sum(B,G,T) :- rule(_,sum(B,G)), T = #sum { W,L: weighted_literal_tuple(B,L,W) }.

supp(A,B) :- rule(     choice(H),B), atom_tuple(H,A).
supp(A,B) :- rule(disjunction(H),B), atom_tuple(H,A).

supp(A) :- supp(A,_).

atom(|L|) :- weighted_literal_tuple(_,L,_).
atom(|L|) :- literal_tuple(_,L).
atom( A ) :- atom_tuple(_,A).

fact(A) :- rule(disjunction(H),normal(B)), atom_tuple(H,A), not literal_tuple(B,_).

true(atom(A))                :- fact(A).%%#(\label{metaD:interpretation:begin}#)
true(atom(A)); fail(atom(A)) :- supp(A), not fact(A).
               fail(atom(A)) :- atom(A), not supp(A).

true(normal(B)) :- literal_tuple(B),
  true(atom(L)): literal_tuple(B, L), L>0;
  fail(atom(L)): literal_tuple(B,-L), L>0.
fail(normal(B)) :- literal_tuple(B, L), fail(atom(L)), L>0.
fail(normal(B)) :- literal_tuple(B,-L), true(atom(L)), L>0.

true(sum(B,G)) :- sum(B,G,T),
  #sum {
    W,L: true(atom(L)), weighted_literal_tuple(B, L,W), L>0;
    W,L: fail(atom(L)), weighted_literal_tuple(B,-L,W), L>0
  } >= G.
fail(sum(B,G)) :- sum(B,G,T),
  #sum {
    W,L: fail(atom(L)), weighted_literal_tuple(B, L,W), L>0;
    W,L: true(atom(L)), weighted_literal_tuple(B,-L,W), L>0
  } >= T-G+1.%%#(\label{metaD:interpretation:end}#)

bot :- rule(disjunction(H),B), true(B), fail(atom(A)): atom_tuple(H,A).%%#(\label{metaD:interpretation:support:begin}#)
bot :- true(atom(A)), fail(B): supp(A,B).%%#(\label{metaD:interpretation:support:end}#)

internal(C,normal(B)) :- scc(C,A), supp(A,normal(B)), scc(C,A'),%%#(\label{metaD:interpretation:acyclicity:begin}#)
                         literal_tuple(B,A').
internal(C,sum(B,G))  :- scc(C,A), supp(A,sum(B,G)),  scc(C,A'),
                         weighted_literal_tuple(B,A',W).

external(C,normal(B)) :- scc(C,A), supp(A,normal(B)), not internal(C,normal(B)).
external(C,sum(B,G))  :- scc(C,A), supp(A,sum(B,G)),  not internal(C,sum(B,G)).

steps(C,Z-1) :- scc(C,_), Z = { scc(C,A): not fact(A) }.

wait(C,atom(A),0)   :- scc(C,A), fail(B): external(C,B), supp(A,B).
wait(C,normal(B),I) :- internal(C,normal(B)), steps(C,Z), I=0..Z-1,
                       fail(normal(B)).
wait(C,normal(B),I) :- internal(C,normal(B)), steps(C,Z), I<Z,
                       literal_tuple(B,A), wait(C,atom(A),I).

wait(C,sum(B,G),I)  :- internal(C,sum(B,G)), steps(C,Z), I=0..Z-1, sum(B,G,T),
  #sum {
    W,L:   fail(atom(L)),   weighted_literal_tuple(B, L,W), L>0, not scc(C,L);
    W,L: wait(C,atom(L),I), weighted_literal_tuple(B, L,W), L>0,     scc(C,L);
    W,L:   true(atom(L)),   weighted_literal_tuple(B,-L,W), L>0
  } >= T-G+1.
wait(C,atom(A),I)   :- wait(C,atom(A),0), steps(C,Z), I=1..Z,
                       wait(C,B,I-1): supp(A,B), internal(C,B).

bot :- scc(C,A), true(atom(A)), wait(C,atom(A),Z), steps(C,Z).%%#(\label{metaD:interpretation:acyclicity:end}#)

true(atom(A)) :- supp(A), not fact(A), bot.%%#(\label{metaD:saturation:true}#)
fail(atom(A)) :- supp(A), not fact(A), bot.%%#(\label{metaD:saturation:fail}#)
\end{lstlisting}
\begin{lstlisting}[float=ht,caption={Auxiliary \texttt{\#show} statements for Listing~\ref{prg:meta:D} (\texttt{show.lp})},label={prg:meta:show},language=clingos]
#show.
#show X: output(X,B),     literal_tuple(B,A), true(atom(A)).
#show X: output(X,B), not literal_tuple(B,_).
\end{lstlisting}

As an example,
consider the simple logic program \lstinline{a.lp}:
\begin{lstlisting}[language=clingos,numbers=none]
1 { a(1..2) }.
\end{lstlisting}
Computing its stable models with the meta encoding in Listing~\ref{prg:meta:D}
(along with the auxiliary \lstinline{#show} statements from Listing~\ref{prg:meta:show})
yields the three expected models:
\begin{lstlisting}[language=shells]
UNIX> clingo --output=reify --reify-sccs a.lp | \
      clingo - metaD.lp show.lp  0
clingo version 5.5.0
Reading from - ...
Solving...
Answer: 1
a(2)
Answer: 2
a(1)
Answer: 3
a(1) a(2)
SATISFIABLE
\end{lstlisting}
Now, the addition of an empty integrity constraint, namely ``\lstinline{:-.}'',
makes the program unsatisfiable.
This is reflected by a single answer set containing all atoms of the program.
This should not to be confused with the third model obtained above:
\begin{lstlisting}[language=shells]
UNIX> clingo --output=reify --reify-sccs a.lp <(echo ":-.") | \
      clingo - metaD.lp show.lp  0
clingo version 5.5.0
Reading from - ...
Solving...
Answer: 1
a(1) a(2)
SATISFIABLE
\end{lstlisting}
In fact,
additional show statements would reveal that the actual stable model also contains the artificial atom
\lstinline{bot} from which all atoms occurring in the original program are derivable
(cf.\ Lines~\ref{metaD:saturation:true} and~\ref{metaD:saturation:fail} in Listing~\ref{prg:meta:D}).
In other words, this special atom expresses the non-existence of stable models,
and by saturating the model with all atoms
it can only exist if no true stable models exist.
This is because the semantics of disjunctive logic programs is based on subset minimization.
Saturation makes sure that \lstinline{bot} is derived only if it is inevitable, that is,
if it is impossible to construct any other models.\footnote{In fact,
  without the two saturating rules in Lines~\ref{metaD:saturation:true} and~\ref{metaD:saturation:fail},
  Listing~\ref{prg:meta:D} would produce a stable model for each interpretation of the original program.
  The ones without \lstinline{bot} represent stable models, while the ones with \lstinline{bot} are mere
  interpretations.
  By saturation, all these interpretations are mapped to the set of all atoms.
  Given that the latter is a superset of all conceivable stable models, it can only exist if no stable models exist.
}

\section{Intermediate language}\label{sec:aspif}
\newcommand\Space{\text{\textvisiblespace}}

To accommodate the rich input language, a general grounder-solver interface is needed.
Although this could be left internal to \clingo,
it is good practice in ASP and neighboring fields to explicate such interfaces via an intermediate language.
This also allows for using alternative downstream solvers or transformations.

Unlike the block-oriented \smodels\ format, the \aspif\ format is line-based.
Notably, it abolishes the need of using symbol tables in \smodels' format~\cite{lparseManual}
for passing along meta-expressions and allows \gringo\ to output information as soon as it is grounded.
An \aspif\ file starts with a header, beginning with the keyword \lstinline{asp}
along with version information and optional tags, viz.\
\[
  \texttt{asp} \Space v_m \Space v_n \Space v_r \Space t_1 \Space \dots \Space t_k
\]
where $v_m,v_n,v_r$ are non-negative integers representing the version in terms of \textit{major}, \textit{minor}, and \textit{revision} numbers,
and each $t_i$ is a tag for $k\geq 0$.
Currently, the only tag is \lstinline{incremental}, meant to set up the underlying solver for multi-shot solving.
An example header is given in the first lines of Listings~\ref{prg:ezy:aspif} and~\ref{aspif:diff} below.
The rest of the file comprises one or more logic programs.
Each logic program is a sequence of lines of \aspif\ statements followed by a \lstinline{0}, one statement or \lstinline{0} per line, respectively.
Positive and negative integers are used to represent positive or negative literals, respectively.
Hence, \lstinline{0} is not a valid literal.

Let us now briefly describe the format of \aspif\ statements and illustrate them with the
simple logic program in Listing~\ref{prg:ezy} as well as the result of grounding a subset of Listing~\ref{prg:lc} only pertaining to difference constraints
in Listing~\ref{aspif:diff}.

\begin{lstlisting}[float=ht,label={prg:ezy:aspif},caption={Representing the logic program from Listing~\ref{prg:ezy} in \aspif\ format},escapeinside={|}{|}]
asp 1 0 0|\label{aspif:ezy:header}|
1 1 1 1 0 0|\label{aspif:ezy:rule:one}|
1 0 1 2 0 1 1|\label{aspif:ezy:rule:two}|
1 0 1 3 0 1 -1|\label{aspif:ezy:rule:tri}|
4 1 a 1 1|\label{aspif:ezy:output:one}|
4 1 b 1 2|\label{aspif:ezy:output:two}|
4 1 c 1 3|\label{aspif:ezy:output:tri}|
0
\end{lstlisting}\newcounter{DUID}\newcommand{\myparagraph}[1]{\par\emph{#1}}
\myparagraph{Rule statements} have form
\addtocounter{DUID}{1}
\[\texttt{\theDUID} \Space H \Space B\]
in which head $H$ has form
\[h \Space m \Space a_1 \Space \dots \Space a_m\]
where
$h \in \{\texttt{0},\texttt{1}\}$ determines whether the head is a disjunction or a choice,
$m \geq 0$ is the number of head elements, and
each $a_i$ is an atom.

Body $B$ has one of two forms:
\begin{itemize}
\item Normal bodies have form
  \[\texttt{0} \Space n \Space l_{1} \Space \dots \Space l_n\]
  where
  $n \geq 0$ is the length of the rule body, and
  each $l_i$ is a literal.
\item Weight bodies have form
  \[\texttt{1} \Space l \Space n \Space l_1 \Space w_1  \Space \dots \Space l_n \Space w_n\]
  where
  $l$ is a positive integer to denote the lower bound,
  $n \geq 0$ is the number of literals in the rule body, and
  each $l_i$ and $w_i$ are a literal and a positive integer.
\end{itemize}
All types of ASP rules are included in the above rule format.
Heads are disjunctions or choices, including the special case of one-element disjunctions for representing normal rules.
As in the \smodels\ format,
aggregate rules are restricted to one-element bodies,
just that in \aspif\ cardinality constraints are taken as special weight constraints.
Otherwise, a body is simply a conjunction of literals.

The three rules in Listing~\ref{prg:ezy} are represented by the statements in Lines~\ref{aspif:ezy:rule:one}--\ref{aspif:ezy:rule:tri} of Listing~\ref{prg:ezy:aspif}.
For instance, the four occurrences of \lstinline{1} in Line~\ref{aspif:ezy:rule:one} capture a rule with a choice in the head, having one element, identified by \lstinline{1}.
The two remaining zeros capture a normal body with no element.
For another example,
Lines~\ref{aspif:d:rule:one}--\ref{aspif:d:rule:six} of Listing~\ref{aspif:diff} represent 6 of the facts in Listing~\ref{prg:grd:diff},
the four regular atoms in Lines~\ref{grd:diff:fact:one}--\ref{grd:diff:fact:for}
along two comprising theory atoms in Lines~\ref{grd:diff:fact:fiv} and \ref{grd:diff:fact:six}.

\myparagraph{Minimize statements} have form
\addtocounter{DUID}{1}
\[\texttt{\theDUID} \Space p \Space n \Space l_1 \Space w_1 \Space \dots \Space l_n \Space w_n\]
where
$p$ is an integer priority,
$n \geq 0$ is the number of weighted literals,
each $l_i$ is a literal, and
each $w_i$ is an integer weight.
Each of the above expressions gathers weighted literals sharing the same priority $p$
from all \lstinline{#minimize} directives and weak constraints in a logic program.
As before, maximize statements are translated into minimize statements.

\myparagraph{Projection statements} result from \lstinline{#project} directives and have form
\addtocounter{DUID}{1}
\[\texttt{\theDUID} \Space n  \Space a_1 \Space \dots \Space a_n\]
where
$n \geq 0$ is the number of atoms, and
each $a_i$ is an atom.

\myparagraph{Output statements} result from \lstinline{#show} directives and have form
\addtocounter{DUID}{1}
\[\texttt{\theDUID} \Space m \Space s \Space n  \Space l_1 \Space \dots \Space l_n\]
where
$n \geq 0$ is the length of the condition,
each $l_i$ is a literal, and
$m\geq0$ is an integer indicating the length in bytes of string $s$
(where $s$ excludes byte `\textbackslash0' and newline).
The output statements in Lines~\ref{aspif:ezy:output:one}--\ref{aspif:ezy:output:tri} of Listing~\ref{prg:ezy:aspif} print the symbolic representation of atom
\lstinline{a}, \lstinline{b}, or \lstinline{c}, whenever the corresponding atom is true.
For instance, the string `\lstinline{a}' is printed  if atom `\lstinline{1}' holds.
Unlike this,
the statements in Lines~\ref{aspif:d:output:one}--\ref{aspif:d:output:for} of Listing~\ref{aspif:diff} unconditionally print the symbolic representation
of the atoms stemming from the four facts in Lines~\ref{grd:diff:fact:one}--\ref{grd:diff:fact:for} of Listing~\ref{prg:grd:diff}.

\myparagraph{External statements} result from \lstinline{#external} directives and have form
\addtocounter{DUID}{1}
\[\texttt{\theDUID} \Space a \Space v\]
where
$a$ is an atom, and
$v \in \{0,1,2,3\}$ indicates free, true, false, and release.

\myparagraph{Assumption statements} have form
\addtocounter{DUID}{1}
\[\texttt{\theDUID} \Space n \Space l_1 \Space \dots \Space l_n\]
where
$n\geq 0$ is the number of literals, and
each $l_i$ is a literal.
Assumptions instruct a solver to compute stable models containing $l_1,\dots,l_n$.
They are only valid for a single solver call.

\myparagraph{Heuristic statements} result from \lstinline{#heuristic} directives and have form
\addtocounter{DUID}{1}
\[\texttt{\theDUID} \Space m \Space a \Space k \Space p  \Space n  \Space l_1 \Space \dots \Space l_n\]
where
$m\in\{0,\dots,5\}$ stands for the ($m$+1)th heuristic modifier among \lstinline{level}, \lstinline{sign}, \lstinline{factor}, \lstinline{init}, \lstinline{true}, and \lstinline{false}, $a$ is an atom,
$k$ is an integer,
$p$ is a non-negative integer priority,
$n \geq 0$ is the number of literals in the condition, and
the literals $l_i$ are the condition under which the heuristic modification should be applied.

\myparagraph{Edge statements} result from \lstinline{#edge} directives and have form
\addtocounter{DUID}{1}
\[\texttt{\theDUID} \Space u \Space v \Space n  \Space l_1 \Space \dots \Space l_n\]
where
$u$ and $v$ are integers representing an edge from node $u$ to node $v$,
$n \geq 0$ is the length of the condition, and
the literals $l_i$ are the condition for the edge to be present.

Let us now turn to the theory-specific part of \aspif.
Once a theory expression is grounded,
\gringo\ outputs a serial representation of its syntax tree.
To illustrate this,
we give in Listing~\ref{aspif:diff} the (sorted) result of grounding all lines of Listing~\ref{prg:lc} related to difference constraints,
viz. Lines~\ref{prg:lc:begin-theory}--\ref{prg:lc:rule-duration} and Line~\ref{prg:lc:rule-diff}.
\begin{lstlisting}[label={aspif:diff},caption={\aspif\ format (excerpt of result)},escapeinside={|}{|}]
asp 1 0 0|\label{aspif:d:header}|
1 0 1 1 0 0|\label{aspif:d:rule:one}|
1 0 1 2 0 0|\label{aspif:d:rule:two}|
1 0 1 3 0 0|\label{aspif:d:rule:tri}|
1 0 1 4 0 0|\label{aspif:d:rule:for}|
1 0 1 5 0 0|\label{aspif:d:rule:fiv}|
1 0 1 6 0 0|\label{aspif:d:rule:six}|
4 7 task(1) 0|\label{aspif:d:output:one}|
4 7 task(2) 0|\label{aspif:d:output:two}|
4 15 duration(1,200) 0|\label{aspif:d:output:tri}|
4 15 duration(2,400) 0|\label{aspif:d:output:for}|
9 0 1 200|\label{aspif:d:theory:term:one}|
9 0 3 400
9 0 6 1
9 0 11 2|\label{aspif:d:theory:term:for}|
9 1 0 4 diff
9 1 2 2 <=
9 1 4 1 -
9 1 5 3 end
9 1 8 5 start|\label{aspif:d:theory:term:nin}|
9 2 7 5 1 6|\label{aspif:d:theory:term:end:one}|
9 2 9 8 1 6|\label{aspif:d:theory:term:start:one}|
9 2 10 4 2 7 9|\label{aspif:d:theory:term:end:start:one}|
9 2 12 5 1 11
9 2 13 8 1 11
9 2 14 4 2 12 13|\label{aspif:d:theory:term:fit}|
9 4 0 1 10 0|\label{aspif:d:theory:atom:one}|
9 4 1 1 14 0
9 6 5 0 1 0 2 1|\label{aspif:d:theory:atom:tri}|
9 6 6 0 1 1 2 3
0
\end{lstlisting}

\myparagraph{Theory terms} are represented using the following statements:
\addtocounter{DUID}{1}
\begin{align}
\texttt{\theDUID} \Space \texttt{0} & \Space u \Space w \label{eq:number}\\
\texttt{\theDUID} \Space \texttt{1} & \Space u \Space n \Space s \label{eq:symbols}\\
\texttt{\theDUID} \Space \texttt{2} & \Space u \Space t \Space n \Space u_1 \Space \dots \Space u_n \label{eq:compound-term}
\end{align}
where
$n \geq 0$ is a length,
index $u$ is a non-negative integer,
integer $w$ represents a numeric term,
string $s$ of length $n$ represents a symbolic term (including functions) or an operator,
integer $t$ is either \texttt{-1}, \texttt{-2}, or \texttt{-3} for tuple terms in parentheses, braces, or brackets, respectively, or an index of a symbolic term or operator, and
each $u_i$ is an integer for a theory term.
Statements (\ref{eq:number}), (\ref{eq:symbols}), and (\ref{eq:compound-term})
capture
numeric terms,
symbolic terms,
as well as
compound terms (tuples, sets, lists, and terms over theory operators).

Fifteen theory terms are given in Lines~\ref{aspif:d:theory:term:one}--\ref{aspif:d:theory:term:fit} of Listing~\ref{aspif:diff}.
Each of them is identified by a unique index in the third spot of each statement.
While Lines~\ref{aspif:d:theory:term:one}--\ref{aspif:d:theory:term:nin} stand for primitive entities of type (\ref{eq:number}) or (\ref{eq:symbols}),
the ones beginning with '\lstinline[mathescape=t]{9$\Space$2}' represent compound terms.
For instance, Lines~\ref{aspif:d:theory:term:end:one} and~\ref{aspif:d:theory:term:start:one} represent \lstinline{end(1)} or  \lstinline{start(1)}, respectively,
and Line~\ref{aspif:d:theory:term:end:start:one} corresponds to \lstinline{end(1)-start(1)}.

\myparagraph{Theory atoms} are represented using the following statements:
\begin{align}
\texttt{\theDUID} \Space \texttt{4} & \Space v \Space n \Space u_1 \Space \dots \Space u_n \Space m \Space l_1 \Space \dots \Space l_m \label{eq:theory-element}\\
\texttt{\theDUID} \Space \texttt{5} & \Space a \Space p \Space n \Space v_1 \Space \dots \Space v_n \label{eq:theory-atom}\\
\texttt{\theDUID} \Space \texttt{6} & \Space a \Space p \Space n \Space v_1 \Space \dots \Space v_n \Space g \Space u_1 \label{eq:theory-atom-bounded}
\end{align}
where
$n \geq 0$ and $m \geq 0$ are lengths,
index $v$ is a non-negative integer,
$a$ is an atom or \texttt{0} for directives,
each $u_i$ is an integer for a theory term,
each $l_i$ is an integer for a literal,
integer~$p$ refers to a symbolic term,
each $v_i$ is an integer for a theory atom element, and
integer~$g$ refers to a theory operator.
Statement (\ref{eq:theory-element}) captures elements of theory atoms and directives, and
statements (\ref{eq:theory-atom}) and (\ref{eq:theory-atom-bounded}) refer to the latter.

For instance,
Line~\ref{aspif:d:theory:atom:one} captures the (single) theory element in `\lstinline+{ end(1)-start(1) }+',
and
Line~\ref{aspif:d:theory:atom:tri} represents the theory atom
`\lstinline[morekeywords={&diff},alsoletter={\&}]+&diff { end(1)-start(1) } <= 200+'.

\myparagraph{Comments} have form
\addtocounter{DUID}{1}
\[\texttt{\theDUID} \Space s\]
where $s$ is a string not containing a newline.

The \aspif\ format constitutes the default output of \gringo~5.
With \clasp~3.2,
ground logic programs can be read in both \smodels\ and \aspif\ format.
The tool \lpconvert\ can be used to convert between both formats~\cite{lpconvert}.

 \bibliographystyle{acmtrans}

 \end{document}